\def\year{2021}\relax
\documentclass[letterpaper]{article} 
\usepackage{aaai21}  
\usepackage{times}  
\usepackage{helvet} 
\usepackage{courier}  
\usepackage[hyphens]{url}  
\usepackage{graphicx} 
\usepackage{natbib}  
\usepackage{caption} 
\usepackage{float}

\usepackage{xcolor}
\usepackage{amsmath,amssymb}
\usepackage{calrsfs}
\DeclareMathAlphabet{\pazocal}{OMS}{zplm}{m}{n}

\usepackage{multirow}
\usepackage{makeidx}
\usepackage{graphicx}
\usepackage{amsmath,amssymb} 
\usepackage{color}

\usepackage{graphicx}
\usepackage{float}
\usepackage{subcaption}
\usepackage{color}

\usepackage{graphicx}
\usepackage{booktabs}

\title{Spectral Distribution Aware Image Generation}
\author{
    Steffen Jung, \textsuperscript{\rm 1}
    Margret Keuper \textsuperscript{\rm 2}
    \\
}
\affiliations {
    \textsuperscript{\rm 1} Max Planck Institute for Informatics, Saarland Informatics Campus \\
    \textsuperscript{\rm 2} Data and Web Science Group, University of Mannheim \\
    steffen.jung@mpi-inf.mpg.de, keuper@uni-mannheim.de
}

\begin{document}


%
%
%
%
%
%
\maketitle              
\begin{abstract}
Recent advances in deep generative models for photo-realistic images have led to high quality visual results. Such models learn to generate data from a given training distribution such that generated images can not be easily distinguished from real images by the human eye. Yet, recent work on the detection of such fake images pointed out that they are actually easily distinguishable by artifacts in their frequency spectra. In this paper, we propose to generate images according to the frequency distribution of the real data by employing a spectral discriminator. The proposed discriminator is lightweight, modular and works stably with different commonly used GAN losses. We show that the resulting models can better generate images with realistic frequency spectra, which are thus harder to detect by this cue.

\end{abstract}
\section{Introduction}
Image generation using generative adversarial networks has made huge progress in recent years. Especially the generation of photo-realistic images at high resolution has arrived at a level where it becomes hard for humans to distinguish between real and generated images. While the training data distribution appears to be well learned in the model's latent space, it is surprising how reliably real and generated images can be distinguished even when various cloaking techniques such as blurring or compression are applied~\cite{cispa2965}. In recent works on the detection of generated images \cite{frank2020leveraging,margret2020upconvolution}, it was argued that this effect, at least partially, is due to artifacts introduced during the generation process itself. Commonly used up-sampling schemes in fact seem to generate these artifacts which are mostly in the high frequency domain and can not be corrected by the network itself~\cite{frank2020leveraging,margret2020upconvolution,bai2020fake}. 
Such artifacts are undesired, not only because generated images can easily be identified as such, but also because they might be perceivable for example as grid patterns on the generated images.

\citet{margret2020upconvolution} propose a GAN regularization approach as a remedy. They argue that a generator network, given a sufficient amount of convolutional layers, can generate images with realistic frequency spectra if they are penalized for deviating from the average frequency spectrum of the real images during training. In fact, the proposed regularization not only allowed to produce more realistic frequency spectra but also had a stabilizing effect on the training.
However, the resulting spectra were still not able to match the distribution of the real data. 

In this paper, we address this problem in a different way. Instead of introducing a regularization term, we propose to use a second discriminator which directly acts on the power spectra of real and generated images. This way, the generator network is not forced to produce images with average power spectra as in \cite{margret2020upconvolution} but is enabled to learn the distribution of frequency spectra from training data.
Since the training of GANs is computationally expensive we argue that an additional discriminator should be lightweight so that diverse architectures can easily adopt it. We therefore base our discriminator on one-dimensional (1D) projections of the frequency spectra instead of acting on the full two-dimensional data. We show that the resulting model can be trained with different commonly used GAN losses and evaluate its ability to fit the real images' frequency spectra in terms of a proposed cloaking score as well as in terms of the performance of frequency based generated image detection \cite{margret2020upconvolution}.

\section{Related Work}
Generative convolutional neural networks have recently been successful in a wide range of applications, such as the generation of photo-realistic images at high resolution \cite{karras2017ProGAN,brock2018large,karras2019StyleGAN,karras2019StyleGAN2} to style transfer~\cite{isola2017image,zhu2017unpaired,zhu2017toward,huang2018multimodal}, or more generally image-to-image translation~\cite{pathak2016context,iizuka2017globally,zhu2017toward,choi2018stargan,mo2018instanceaware,karras2019style} and text-to-image translation \cite{reed2016generative,dai2017towards,zhang2017stackgan,zhang2018stackgan++}. 
Generative adversarial networks (GANs)~\cite{gan14nips} play a crucial role in this context. They aim to approximate a latent-space model of the underlying data distributions from training images in a zero-sum game between a generator and a discriminator network. From this latent data distribution model, new samples can be generated (drawn) by sampling. Recent works towards improving GANs proposed different loss functions, regularizations or latent space constraints~\cite{gulrajani2017wgan,mao2017least,gulrajani2017improved,miyato2018spectral_norm,mirza2014conditional,donahue2016adversarial,gurumurthy2017deligan,margret2020upconvolution,brock2018BigGAN,kodali2017convergence} to improve training stability and aim at high image resolutions  \cite{karras2017ProGAN,karras2019StyleGAN,karras2019StyleGAN2}.  

Since, with this progress, generated images are hard to distinguish from real images by the human eye, recent work has gained interest in the detection of generated images. On the one end, automatically detecting generated images helps to protect content authenticity in the context of deep fakes. On the other hand, it can help to improve the generation process itself as it allows to find systematic mistakes currently made by image generation networks. One such systematic mistake seems to be especially apparent in the frequency spectra of generated images~\cite{durall2019unmasking,Wang_2020_CVPR,frank2020leveraging,bai2020fake}. By feeding the Fourier transform or the discrete cosine transform of generated images into a deep network~\cite{Wang_2020_CVPR} or simpler learning models such as support vector machines~\cite{durall2019unmasking} or ridge regression~\cite{frank2020leveraging}, surprisingly high detection rates can be achieved. As analyzed for example in \cite{durall2019unmasking,frank2020leveraging}, these systematic artifacts in the frequency domain are an effect of the generation process itself, more precisely in the up-convolutions.
%
%
%
%
In~\cite{margret2020upconvolution}, a regularization approach acting on the frequency spectra of generated images has been proposed, which supports the training process by penalizing whenever a generated image's spectrum deviates from the average spectrum of the real data.
Our approach is related to~\cite{margret2020upconvolution}. However, we argue that a pointwise regularization of all generated spectra w.r.t. the average spectrum of real images is suboptimal since it does not properly allow to learn the data distribution. Instead, we propose to use a discriminator on the power spectra in order to learn the generation of images according to both, the distribution of the real data in spatial as well as in frequency domain.

\paragraph{Contributions.} We make the following contributions:
\begin{itemize}
    \item We propose to learn to generate images with a higher fidelity to the real images' frequency distribution by employing a discriminator that acts on the frequency spectra.
    \item The proposed discriminator is efficient and modular and can be trained stably with different GAN losses.
    \item We propose a measure for the spectral distribution fidelity which allows to assess how well generated images can be distinguished from real ones by their frequency spectra.
    \item We show in various experiments that the proposed approach enables to generate images with highly realistic frequency spectra and therein outperforms the recent method from~\cite{margret2020upconvolution} without sacrificing image quality in terms of FID.
\end{itemize}

\section{Spectral Properties of Image Generation}
Generative neural network architectures such as GANs generate high-dimensional outputs (i.e.~high-resolution images) from low dimensional latent space samples. Therefore, they rely on stepwise up-scaling mechanisms which successively increase the output resolution, followed by convolutional layers. Such up-sampling can be done for example using "bed of nails", nearest neighbor or bilinear interpolation, all of which have different effects on the properties of the resulting up-sampled feature map or image.
The spectral properties of an image $I$ can be analyzed by its discrete Fourier transform
\begin{align}
&{\hat{I}}(k,\ell)=\sum_{m=0}^{M-1}\sum_{n=0}^{N-1}e^{-2\pi i\cdot\frac{m\cdot k}{M}}e^{-2\pi i\cdot\frac{n\cdot \ell}{N}}\cdot I(m,n), \\
&\mathrm{for }\quad k=0,\dots,M-1,\quad \ell=0,\dots,N-1,\nonumber
\end{align}
\begin{figure}[t]
\centering
\begin{minipage}[t][][b]{0.9\columnwidth}
\begin{tabular}{@{}c@{}c@{}}
\hspace{0.9cm}\includegraphics[height=1.6cm]{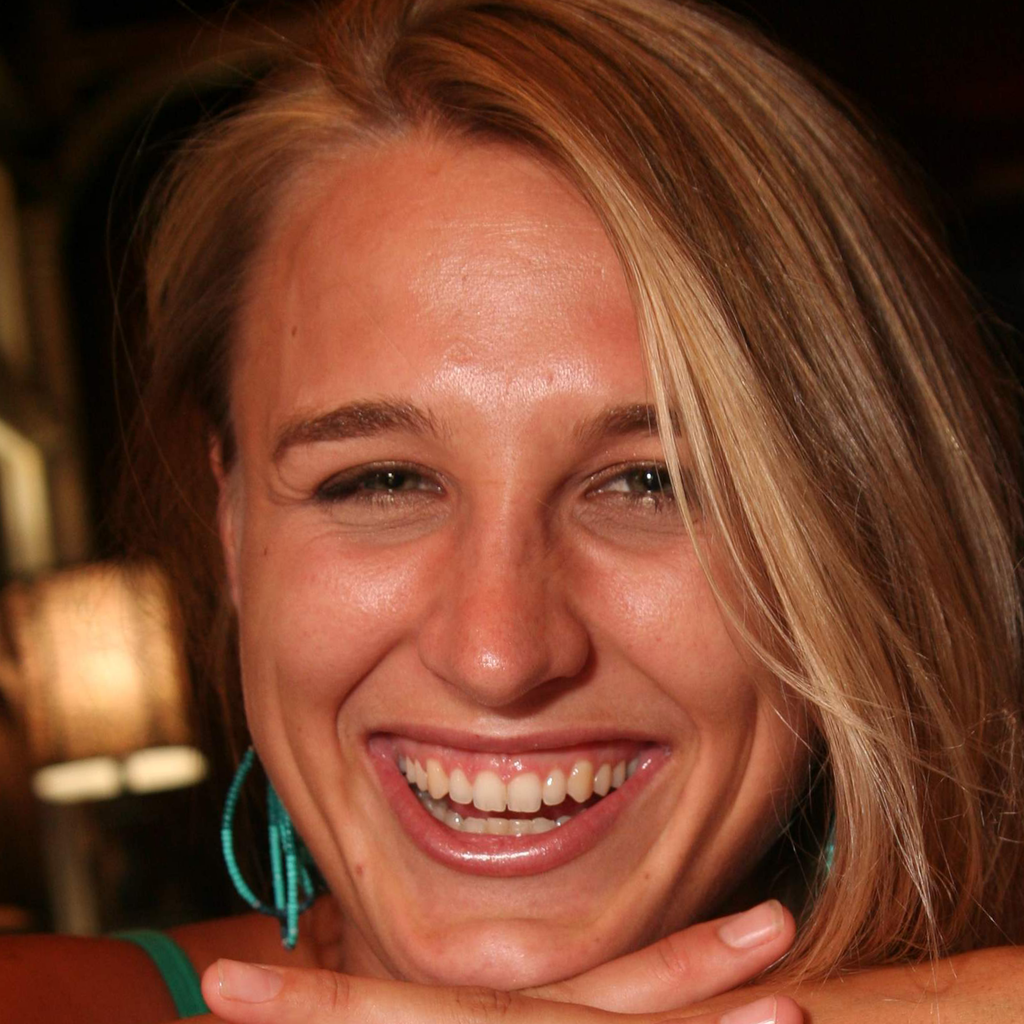}&
\hspace{1.3cm}\includegraphics[height=1.8cm]{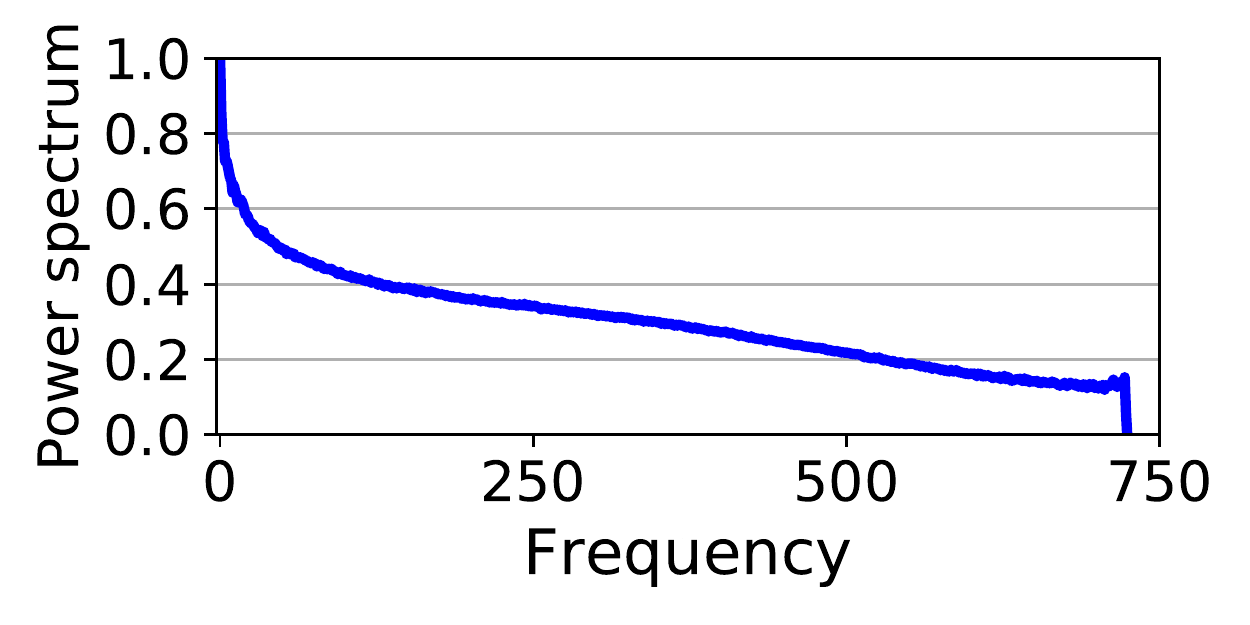}\\
\hspace{0.9cm}\includegraphics[height=1.6cm]{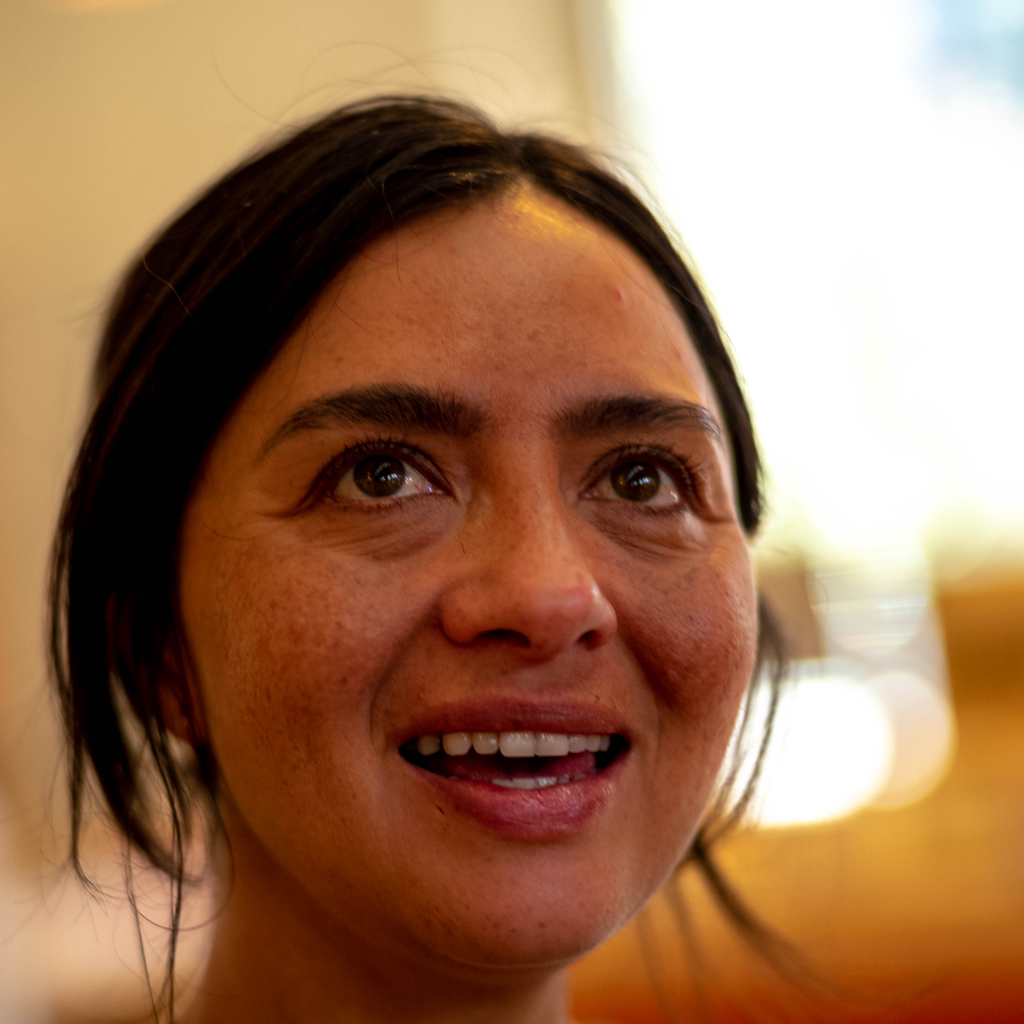}&
\hspace{1.3cm}\includegraphics[height=1.8cm]{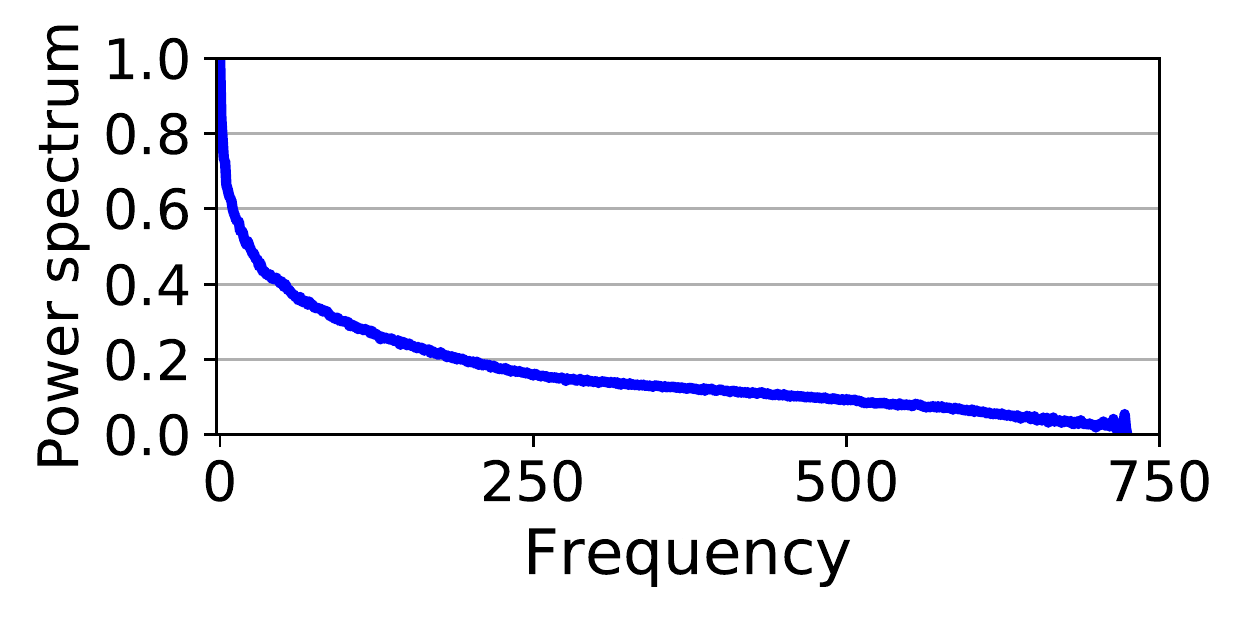}\\
\hspace{0.9cm}\includegraphics[height=1.6cm]{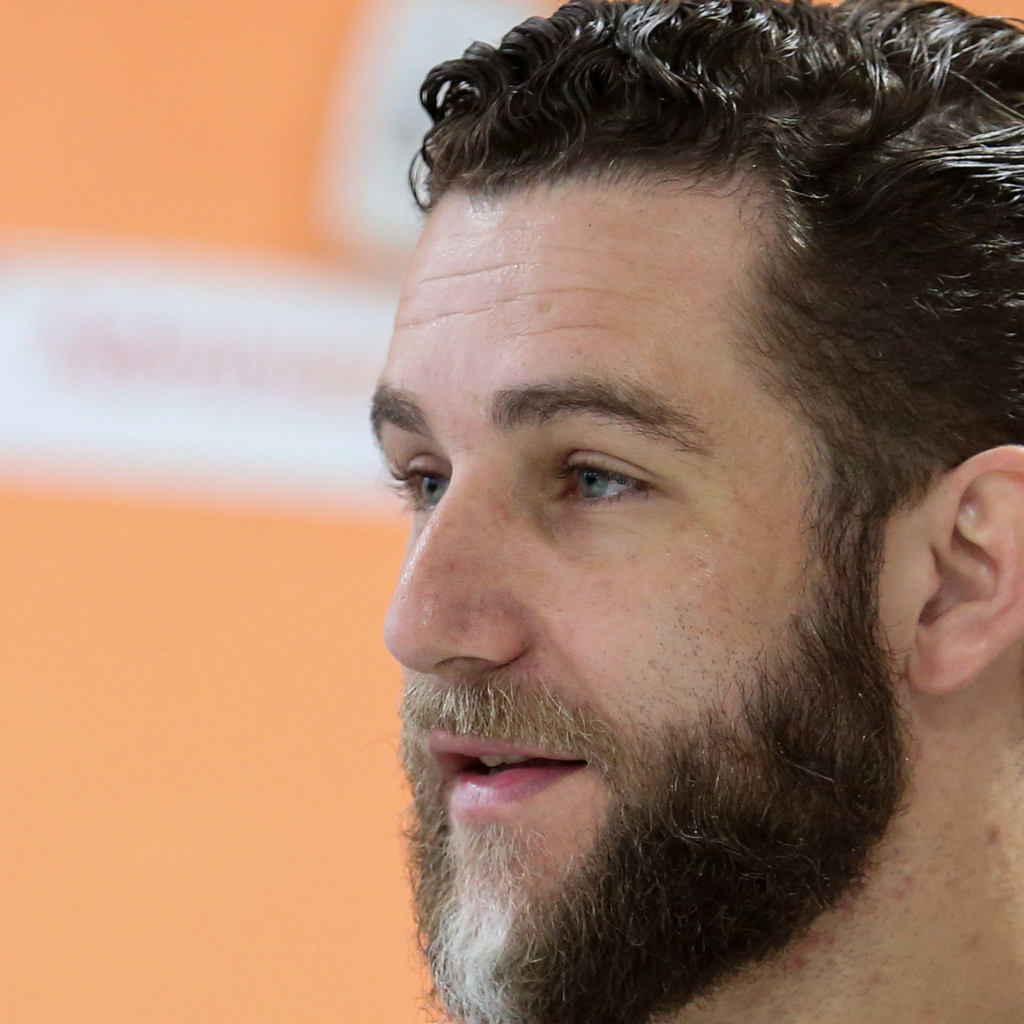}&
\hspace{1.3cm}\includegraphics[height=1.8cm]{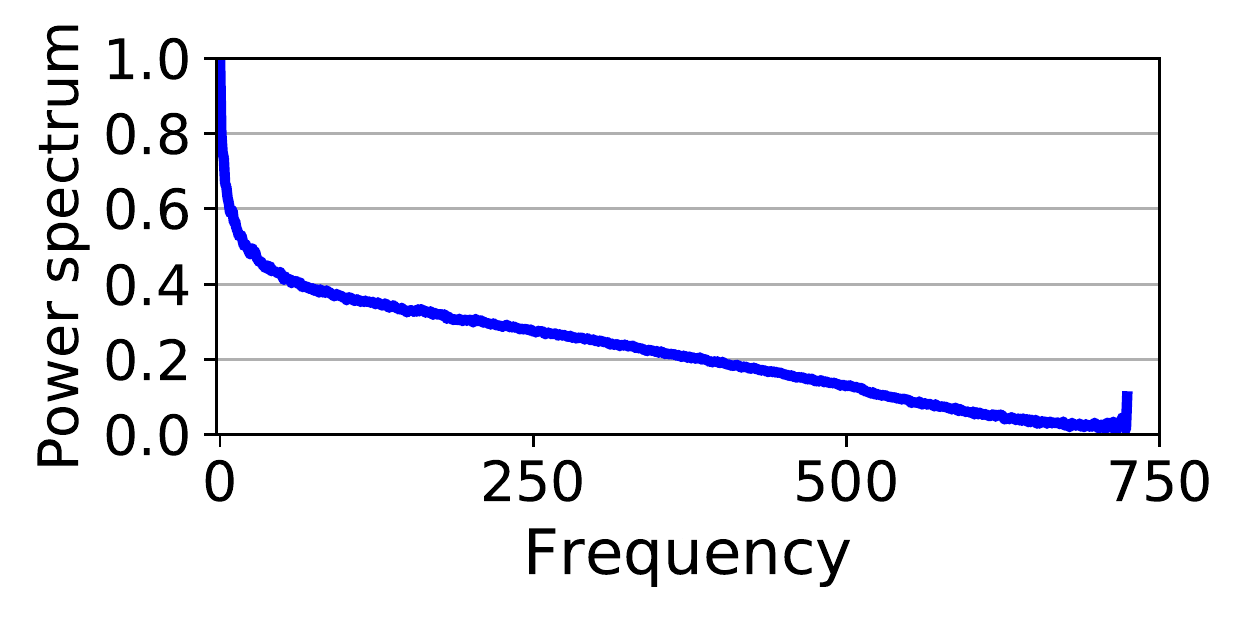}\\
\end{tabular}
\end{minipage}
\hspace{0.5cm}
\begin{minipage}[t][][c]{1\columnwidth}
\centering
\includegraphics[width=1\textwidth]{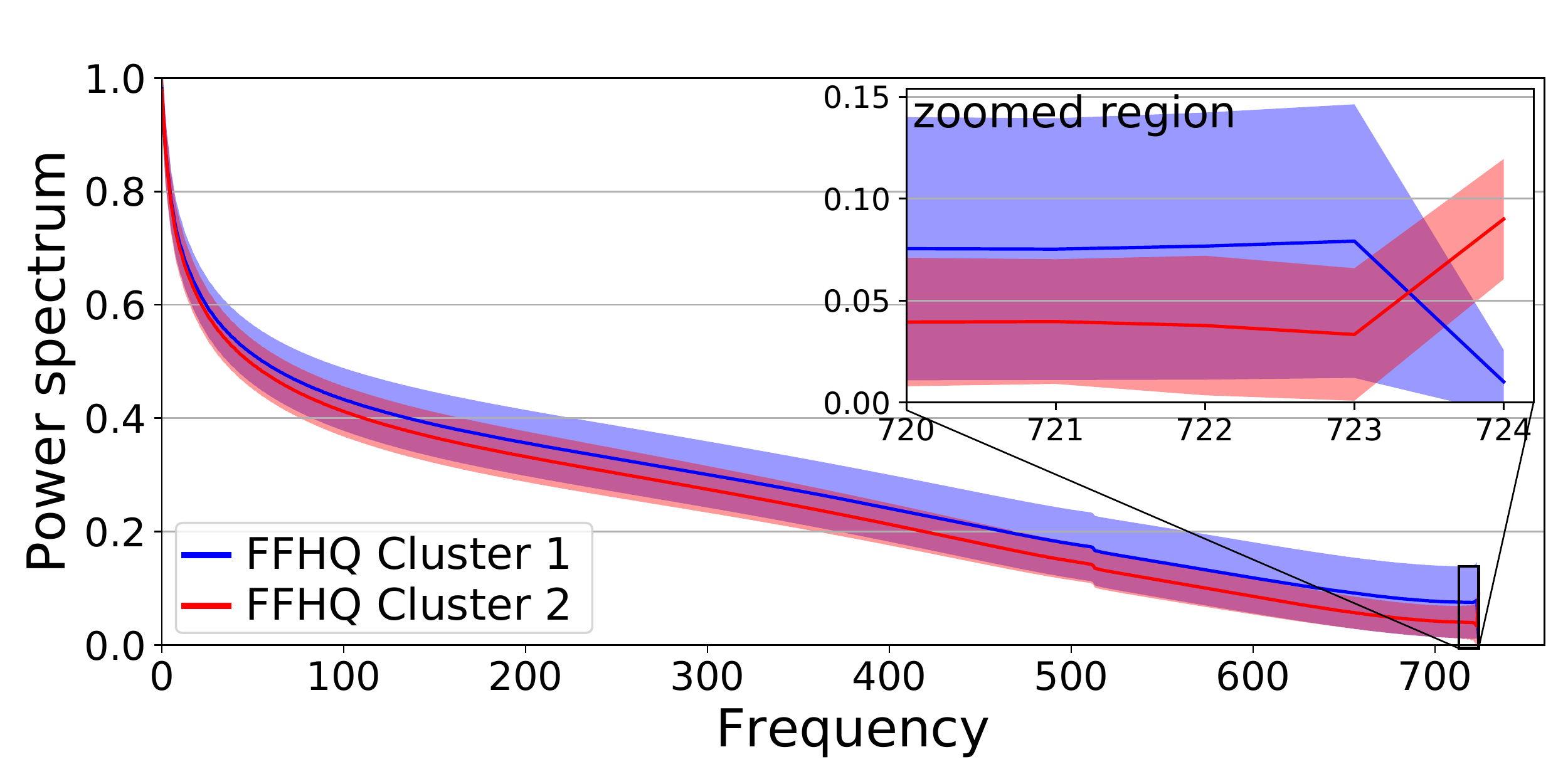}
\end{minipage}
    \caption{(top) The frequency profiles from the FFHQ images are diverse, suggesting that their distribution is not uni-modal. (bottom) Average spectral profiles of the real data (FFHQ) after clustering with k-means (k=$2$). In the highest frequencies, the profiles can be well separated in two clusters.
    } 
    \label{fig1}
\end{figure}
which transforms a 2D image into a 2D array of its spatial frequencies. During up-sampling, an images frequency spectrum is altered depending on the up-sampling method. While bilinear interpolation results in smooth images with few high frequency components, the more commonly used "bed of nails" interpolation up-sampling, which fills the missing values with zeros, initializes the up-sampled image (or feature map) with many high frequency components. In \cite{margret2020upconvolution}, a theoretic analysis of the effect of upsampling using bed of nails interpolation is provided, which we summarize below.
\subsection{Spectral Effects of Up-Sampling.}
For simplicity,  \cite{margret2020upconvolution} consider in their analysis the case of one-dimensional signals, which generalizes to higher dimensions. For a signal $a$ the discrete Fourier Transform $\hat{a}$ is given by
\begin{equation}
\hat{a}_k=\sum_{j=0}^{N-1}e^{-2\pi i\cdot\frac{jk}{N}}\cdot a_j, \quad \mathrm{for }\quad k=0,\dots,N-1.
\end{equation}
Increasing $a$'s spatial resolution by factor $2$ results in 
\begin{align}
\hat{a}^{up}_{\bar{k}}&=\sum_{j=0}^{2\cdot N-1}e^{-2\pi i\cdot\frac{j\bar{k}}{2\cdot N}}\cdot a^{up}_j
\nonumber\\
&=\sum_{j=0}^{N-1}e^{-2\pi i\cdot\frac{2\cdot j\bar{k}}{2\cdot N}}\cdot a_j
+ \sum_{j=0}^{N-1}e^{-2\pi i\cdot\frac{2\cdot (j+1)\bar{k}}{2\cdot N}}\cdot b_j,
\label{eq:theory2}
\end{align}
for $\bar{k}=0,\dots,2N-1$, where $b_j=0$ for "bed of nails" interpolation (and $b_j=a_j$ for nearest neighbor interpolation).
The first term in Eq. \eqref{eq:theory2} is similar to the original Fourier Transform while the second term is zero for $b_j=0$.  
It can be seen that if the spatial resolution is increased by a factor of $2$ the frequency axes are scaled by a factor of $\frac{1}{2}$. From sampling theoretic considerations, it is \cite{margret2020upconvolution}
\begin{align}
\eqref{eq:theory2}
&=\sum_{j=0}^{2\cdot N-1}e^{-2\pi i\cdot\frac{j\bar{k}}{2\cdot N}}\cdot \sum_{t=-\infty}^{\infty}a^{up}_j\cdot\delta(j-2t).\label{eq:theory3}
\end{align}

The point-wise multiplication with the Dirac impulse comb removes exactly the values for which $a^{up}=0$. In order to apply the convolution theorem~\cite{conv-theorem}, one has to assume $a$ being a periodic signal. Then, it is 
\begin{align}\hat{a}^{up}_{\bar{k}} &
=\frac{1}{2}\cdot  \sum_{t=-\infty}^{\infty} \left(\sum_{j=-\infty}^{\infty}
e^{-2\pi i\cdot\frac{j\bar{k}}{2\cdot N}}a^{up}_j\right)\left(\bar{k}-\frac{t}{2}\right)\, \nonumber\\
&\overset{\eqref{eq:theory2}}{=}\frac{1}{2}\cdot  \sum_{t=-\infty}^{\infty} \left(\sum_{j=-\infty}^{\infty}
e^{-2\pi i\cdot\frac{j\bar{k}}{N}}\cdot a_j\right)\left(\bar{k}-\frac{t}{2}\right).
\end{align}
Thus, the frequency spectrum $\hat{a}^{up}$ will contain replica of the frequency spectrum of $a$. More precisely, all frequencies beyond $\frac{N}{2}$ are up-sampling artifacts which can only be removed if the up-sampled signal is smoothed appropriately.

In addition to these theoretic considerations \cite{margret2020upconvolution} also show practically that the correction of the resulting spectra is not possible with the commonly used $3\times 3$ convolutional filters.



\subsection{Analysis of Real Data Distribution}
\label{sec:distribution}
In order to analyze the distributions of the real as well as the generated images' frequency spectra, we consider an aggregate representation as done in \cite{margret2020upconvolution}. We compute the magnitude of the 2D spectral frequencies and integrate over the resulting 2D array for every radius to get a one-dimensional profile of the power spectrum, called \emph{azimuthal integral}
\begin{align}\label{eq:AI}
&AI(\omega_k)=  \int_0^{2\pi} \|\hat{I}\left(\omega_k\cdot \mathrm{cos}(\phi),\omega_k\cdot \mathrm{sin}(\phi)\right)\|^2 \mathrm{d}\phi\, \nonumber\\
&\mathrm{for }\quad k=0,\dots,M/2-1\, ,
\end{align}
assuming square images $I$. As pointed out in~\cite{margret2020upconvolution} this notation is obviously abusive for $\hat{I}$ being discrete. Practically, the integral is implemented as a sum over interpolated values. 
See Fig. \ref{fig1}(top) for some examples of real images from the Flickr-Faces-HQ dataset (FFHQ) \cite{karras2019StyleGAN} and their corresponding frequency profiles computed using Eq.~\eqref{eq:AI}. 

From these examples, one can see that the frequency profiles of the images are diverse and vary significantly, especially in the high frequency regime. Fig.~\ref{fig1}(bottom) shows the average frequency profiles after clustering the images of FFHQ by the magnitude of their highest frequency. The profiles can be well clustered into two groups just by looking at the highest frequency, which indicates that the frequency distribution of the real data can not be well approximated by a uni-modal Gaussian distribution as done in~\cite{margret2020upconvolution}. Since the true distribution is unknown, we argue that a suitable way of generating images with a higher spectral fidelity is to use an adversarial approach. Our model therefore contains an additional discriminator, taking as input the frequency profile of real and generated images.

\subsection{GAN Evaluation in the Frequency Domain}
\label{sec:cloaking}
\begin{figure}[t]
        \centering
        \includegraphics[width=\linewidth]{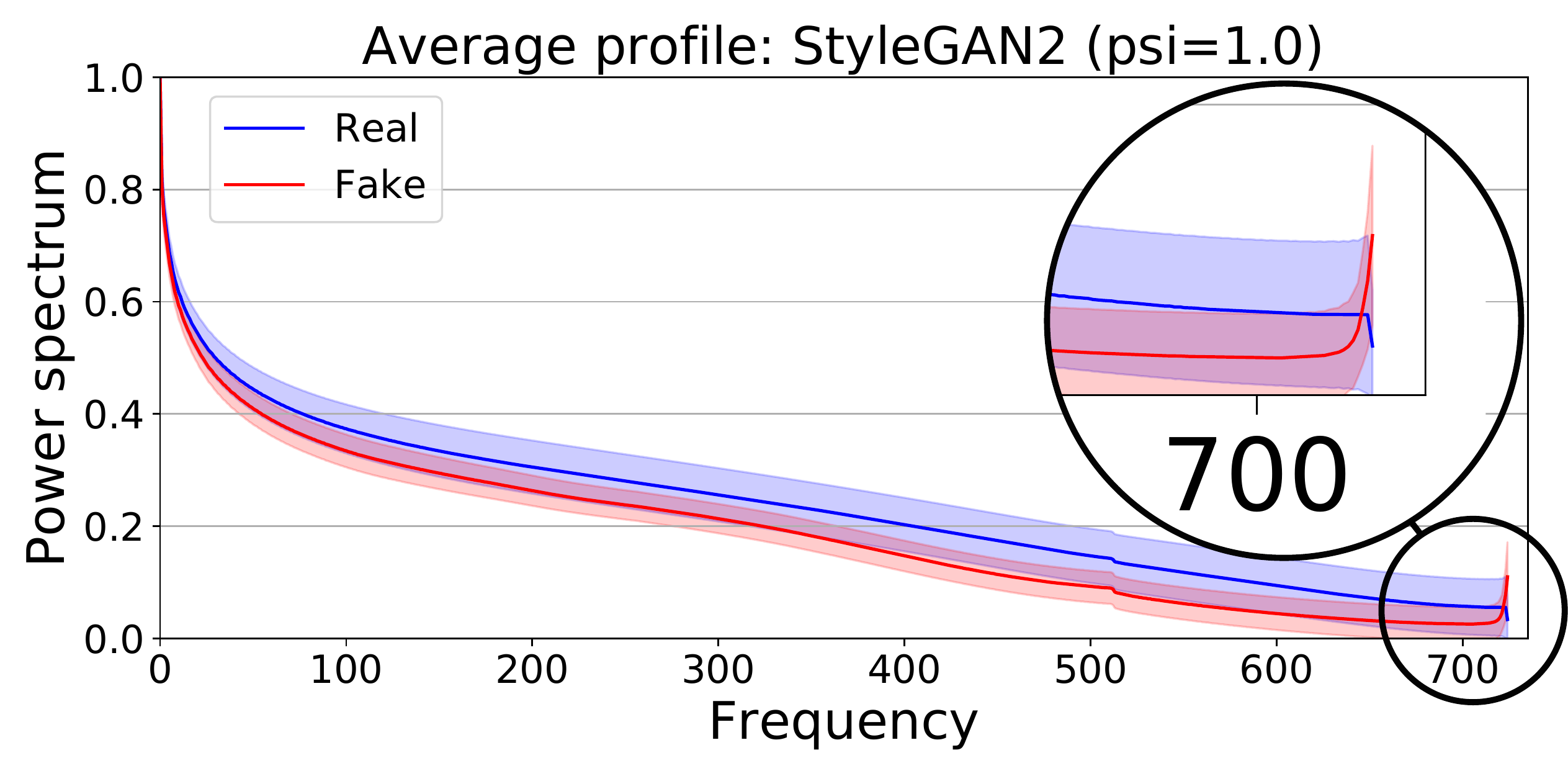}
        \caption{
            Average spectral profiles of real data (FFHQ) and data generated by StyleGAN2~\cite{karras2019StyleGAN2}.
            High frequency components in the generated images indicate grid artifacts or noise, which was not removed by the discriminator. 
        }
        \label{fig:cprofiles}
    \end{figure}
In general, the evaluation of the quality of generated images by GANs is highly subjective.
Therefore, \cite{heusel2017fid} introduced the \textit{Fréchet Inception Distance} (FID) to provide a method of comparing different image generating models.
Since its proposal, this quality measure is widely adopted as one of the key indicators to proof the qualitative performance of GANs.
At the time of writing, StyleGAN2 \cite{karras2019StyleGAN2} is the best performing model according to FID, and trained on the FFHQ dataset.
Subjectively, generated images by StyleGAN2 are hard to distinguish from the real images.
This is reflected by a low FID score of $2.84 \pm 0.03$.
However, recent advances in DeepFake detection \cite{durall2019unmasking,margret2020upconvolution, frank2020leveraging} showed that it is possible to recognize such images using frequency domain representations.
Figure \ref{fig:cprofiles} shows the averaged profiles of a real data distribution, represented by FFHQ, and the corresponding images drawn from the learned distribution by StyleGAN2.
It is visible that the power spectrum is misaligned throughout most frequencies.
Images produced by StyleGAN2 contain high frequency components, which can be observed by a rapid increase in the power spectrum of the last frequencies. Such behavior indicates the presence of grid artifacts and high frequency noise. 
Thus, 
we argue that generated images should not only be assessed by their FID score but also by their spectral distribution fidelity and propose a method to evaluate the distribution alignment in the frequency domain, as a supplementary metric. 
Our method transforms images into spectral profile vectors by azimuthal integration (Eq.~\eqref{eq:AI}).
Then, we train a Logistic Regression on the spectral profiles given by all images taken from the real distribution, and a corresponding number of generated images given by a model. 
We then translate the training accuracy into a score according to 
\begin{equation}
    \text{Cloaking Score} = 1 - 2 \cdot \lvert \text{Accuracy} - 0.5 \rvert.
    \label{eq:cloaking}
\end{equation}

The \textit{Cloaking Score} (CS) ranges in $[0,1]$, where a score of $1.0$ indicates perfect spectral alignment, and a score of $0.0$ indicates that generated images can be linearly separated from real images in the spectral domain. For the StyleGAN2~\cite{karras2019StyleGAN2} example in Figure~\ref{fig:cprofiles}, the CS is 0.042 (acc. $0.979$) after $1\,000$ epochs, 0.022 (acc. $0.989$) after $10\,000$ epochs, and 0.018 (acc. $0.991$) after $40\,000$ training epochs.
%
%
Since the CS evaluation method gives rise to a trade-off between preciseness and runtime, we settled for scores after $1\,000$ epochs.
Runtime for $140k$ $64^2$-sized images is around $3$min, when all images are read from disk.

\section{Learning to Generate Spectral Distributions}
\label{sec:model}
\begin{figure*}
\centering
    \includegraphics[width=0.75\textwidth]{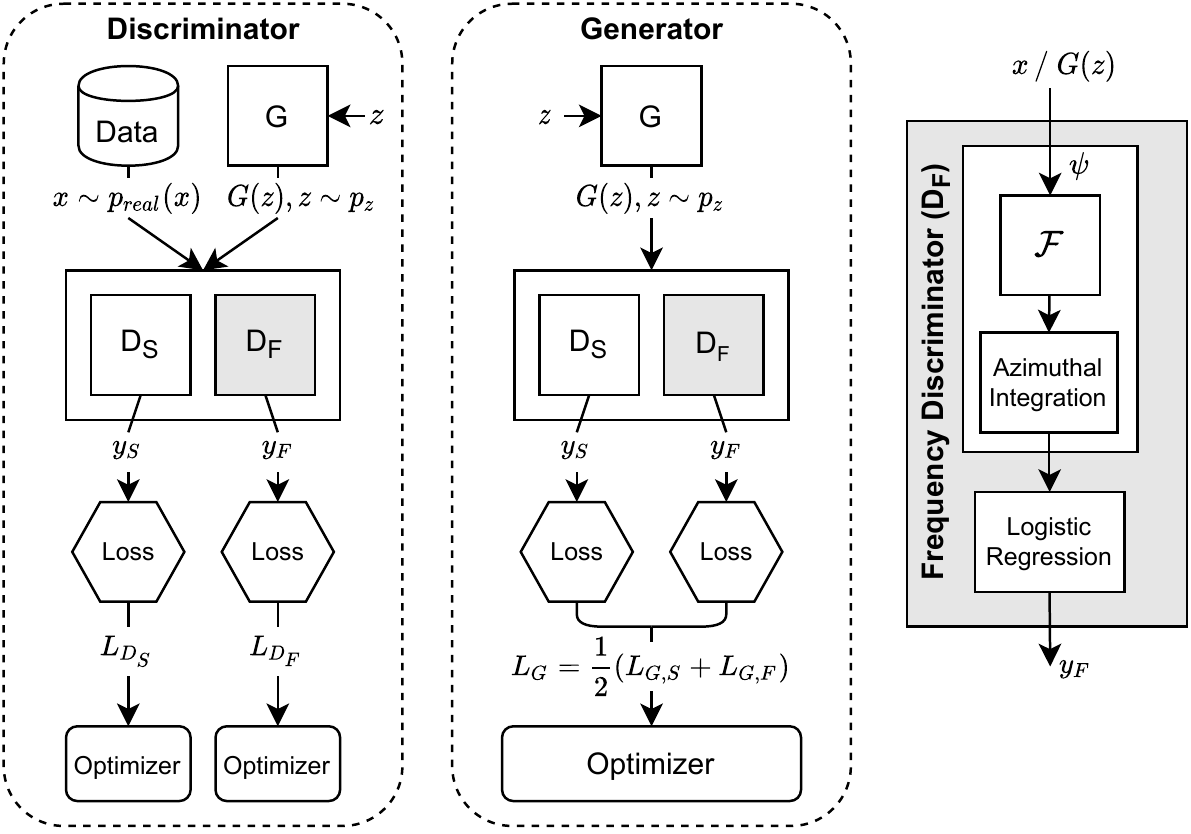}
    \caption{
    Training process of the proposed model.
    Losses are computed based on predictions representing the realness score of the given image in the spatial domain ($y_S$) as well as the frequency domain ($y_F$).
    Both spatial ($D_S$) and frequency ($D_F$) discriminators are trained separately by individual optimizer. 
    When training generator $G$, both resulting losses are averaged.} 
    \label{fig:training}
\end{figure*}
Generative adversarial networks are trained in a minimax game between generator and discriminator networks, 
where the discriminator wants to recognize generated images and the generator wants the discriminator to perform poorly, i.e.~to generate images which the discriminator can not tell apart from real ones, leading to objectives such as \cite{gan14nips}
\begin{align}
\min_G \max_D V(D,G)&=&\mathbb{E}_{{\bf x}\sim p_{\mathrm{data}}({\bf x})}[\mathrm{log}(D({\bf x}))]\,+\,\quad\nonumber\\
&&\mathbb{E}_{{\bf z}\sim p_{{\bf z}}({\bf z})}
[\mathrm{log}\left(1-D(G({\bf z}))\right)].\,
\end{align}
\citet{margret2020upconvolution} propose to add a penalty term, a \textit{Spectral Regularization}, to the generator to reduce discrepancies between the real and generated spectral distributions.
This regularization is implemented as a cross entropy loss on the $AI$ \eqref{eq:AI} profiles of the generator output, to minimize the difference of the generated images' spectra to the mean spectrum of real images.
Hence, the generator tries to minimize this penalty by forcing each image to reflect the same, average profile.
In our experiments, this leads to an unstable training progress (see supplementary material). Out of $5$ runs, only $3$ models were able to reach $500$ epochs without degenerating due to mode collapse.
As we have shown above, the real spectral distribution is not a uni-modal Gaussian distribution around the average profile.
Instead, there are certain characteristics that the generator needs to be able to learn. The regularizer by \cite{margret2020upconvolution} seems suboptimal in this respect. Therefore, we argue that instead of learning one average profile, the generator should be taught to generate images according to the real data distribution in spatial as well as in spectral domain.

\begin{table}
    \caption{Investigated Discriminator Losses.}
    \label{tab0}
    \centering
    \begin{tabular}{ll}
    \toprule
    $\mathcal{L}_{\mathrm{DCGAN}}$& $- \mathbb{E}_{{\bf x}
    }
    [\mathrm{log}(D({\bf x}))]\,-\,
\mathbb{E}_{{\bf \hat{x}}
}
[\mathrm{log}\left(1-D({\bf \hat{x}})\right)]$\\
$\mathcal{L}_{\mathrm{LSGAN}}$& $- \mathbb{E}_{{\bf x}
}[(D({\bf x})-1)^2]\,+\,
\mathbb{E}_{{\bf \hat{x}}
}
[D({\bf \hat{x}})^2]$\\
$\mathcal{L}_{\mathrm{WGAN}}$& $- \mathbb{E}_{{\bf x}
}[D({\bf x})]\,+\,
\mathbb{E}_{{\bf \hat{x}}
}
[D({\bf \hat{x}})]$\\
$\mathcal{L}_{\mathrm{WGAN-GP}}$& $\mathcal{L}_{\mathrm{WGAN}}+$\\
&$\lambda \mathbb{E}_{{\bf \hat{x}} 
}[(\|\nabla D(\alpha {\bf x}+(1-\alpha {\bf \hat{x}}))\|_2-1)^2] $\\

    \bottomrule
    
    \end{tabular}
    \end{table}

We propose to use a second discriminator network ($D_F$) for this purpose (see Figure \ref{fig:training}).
While one could obviously consider to use full 2D power spectra and a convolutional architecture as in the original discriminator, we argue that an additional discriminator should be as lightweight and modular as possible.
The proposed frequency spectrum discriminator $D_F$ takes as input real or generated images which then go through a spectral transformation layer $\psi$ that computes the magnitude of their Fourier transform.
Afterwards, the azimuthal integral (Eq.~\eqref{eq:AI}) is computed, projecting the 2D spectrum onto a 1D vector. Then, we add a fully connected layer with Sigmoid activation function, which is trained to discriminate between real and generated images. 

Such a simple discriminator can in principle be trained with any of the commonly used GAN loss functions. To investigate the dependency of the discriminator performance on the employed loss, we consider models with the commonly used loss functions in Tab. \ref{tab0} and focus on simple architectures (i.e. DCGAN \cite{radford2015dcgan}, LSGAN \cite{mao2017least}, and Wasserstein GAN (WGAN)~\cite{gulrajani2017wgan,gulrajani2017improved}).

We train both the spatial ($D_S$) and the frequency ($D_F$) discriminators by separate losses, applying the same loss function (e.g.~BCE in case of DCGAN). For training the generator $G$, both resulting losses are combined by averaging (see Fig.~\ref{fig:training}). 
%
Additionally, we keep the proposed architectural suggestions by \cite{margret2020upconvolution}, namely increasing the last upconvolutional layer to $8 \times 8$, and adding an additional block of three $5 \times 5$ to the end of the generator network, which, as they argue empirically, is necessary to provide the network with the capacity to "repair" the spectral artifacts.  
\paragraph{Implementation Details.}
The memory comsumption of the proposed discriminator depends on the image width and height $n$ and adds only  $\left\lceil n/\sqrt{2} \right\rceil + 1$ parameters. We evaluate it with different architectures.
DCGAN, LSGAN and WGAN-GP were trained using Adam optimizer, for WGAN, we used RMSprop, with a learning rate of $0.0002$ and a batch size of $128$ for $500$ epochs. In all cases, we apply the same loss function on both discriminators.


\section{Experiments}
\label{sec:experiments}
\begin{table}
    \caption{Evaluation of the architectural change by \citet{margret2020upconvolution} on the image generation quality for DCGAN and LSGAN at $64^2$ pixel resolution. In both cases, the changed up-sampling and additional convolutional layers lead to an improved FID and spectral difference, while the cloaking score is low.}
    \centering
    \begin{tabular}{l c c c}
        \toprule
         Model & FID $\downarrow$ & SD $\downarrow$ & CS $\uparrow$ \\
        \midrule
DCGAN$_{orig}$ & 17.749 & 1.510 & {\bf 0.35} \\
DCGAN &{\bf 15.257} &{\bf 1.293} & 0.16\\
LSGAN$_{orig}$&18.423&1.602&{\bf 0.28}\\
LSGAN& {\bf 15.518} & {\bf 0.468} & 0.04\\
\bottomrule
    \end{tabular}
    \label{tab:taborig}
\end{table}

\begin{table*}
    \caption{Evaluation of the proposed discriminator in terms of GAN quality measures and generated image detection scores. 
    }
    \label{tab1}
    \small
    \centering
    \begin{tabular}{l|l|ccc|ccccc}
        \toprule
        &&\multicolumn{3}{c}{GAN Quality}&\multicolumn{5}{c}{Detection Accuracy}\\
         &&&&&\multicolumn{2}{c}{Durall et al.$^*$}
        &\multicolumn{2}{c}{Durall et al.}&Wang et al.\\
        Size & Model & FID $\downarrow$ & SD $\downarrow$ & CS $\uparrow$ & SVM $\downarrow$ & LR $\downarrow$ & SVM $\downarrow$ & LR $\downarrow$ & ACC $\downarrow$ \\
        \midrule
\multicolumn{1}{c|}{\multirow{9}{*}{$64^2$}} 
& DCGAN & {\bf 15.257} & 1.293 & 0.16& 0.89&0.89& 0.89&0.89  &0.81\\
& DCGAN (Durall et al.)& 29.875 & 0.311 & 0.25 & 0.51&{\bf 0.49}& 0.79&0.82&0.67 \\
 & DCGAN, Spectral & 15.591 & {\bf 0.042} & {\bf 0.84}  &{\bf 0.50} &0.50 &{\bf 0.59}&{\bf 0.58}&{\bf 0.66}\\ 
 & LSGAN & 15.518 & 0.468 & 0.04 & 0.94 & 0.94& 0.94 & 0.94&0.67\\
 & LSGAN, Spectral & {\bf 15.515} & {\bf 0.041} & {\bf 0.86}  & {\bf 0.51} & {\bf 0.50} & {\bf 0.55}&{\bf 0.56}&{\bf 0.64}\\ 
 & WGAN & {\bf 47.704} & 1.291 & 0.01 & 0.99&0.99&0.99&0.99&0.79\\
 & WGAN, Spectral & 47.948 & {\bf 0.029} & {\bf 0.85} & {\bf 0.50}&{\bf 0.50} & {\bf 0.62}&{\bf 0.65}&0.79\\ 
 & WGAN-GP & {\bf 39.404} & 0.575 & 0.18 & 0.92&0.94&0.92&0.94&0.76\\
 & WGAN-GP, Spectral & 39.441 & {\bf 0.252} & {\bf 0.54} &{\bf 0.51}&{\bf 0.51}&{\bf 0.75}&{\bf 0.76}&{\bf 0.65}\\ \midrule
\multicolumn{1}{c|}{\multirow{9}{*}{$128^2$}} & DCGAN & 19.908 & 1.579 & 0.00  & 0.99 & 0.99 &0.99&0.99 &0.82\\
& DCGAN (Durall et al.) & 41.896 & 0.449 & 0.08 & \bf{0.51} & 0.51 & 0.98 & 0.98 & 0.87\\
 & DCGAN, Spectral & {\bf 19.898} & {\bf 0.087} & {\bf 0.72} & {\bf 0.51}&{\bf 0.50}&{\bf 0.66}&{\bf 0.69}&{\bf 0.81} \\ 
 & LSGAN & 20.850 & 3.786 & 0.01  & 0.99 &0.99&0.99&0.99&0.81\\
 & LSGAN, Spectral & {\bf 19.043} & {\bf 0.074} & {\bf 0.76} &  {\bf 0.50}& {\bf 0.50} & {\bf 0.82}&{\bf 0.83}&{\bf 0.80}\\ 
 & WGAN & {\bf 20.757} & 1.996 & 0.01& 0.99 & 0.99 & 0.99&0.99 &0.83 \\
 & WGAN, Spectral & 24.568 & {\bf 0.136} & {\bf 0.66} & {\bf 0.52}& {\bf 0.53} &{\bf 0.61} &{\bf 0.60}&{\bf 0.81}\\ 
 & WGAN-GP & 47.630 & 2.349 & 0.02 &  0.99& 0.99 &   0.99& 0.99&0.96\\ 
 & WGAN-GP, Spectral & {\bf 42.740} & {\bf 0.260} & {\bf 0.38}  &  {\bf 0.50}&{\bf 0.50}& {\bf 0.82} & {\bf 0.83}&{\bf 0.62} \\\midrule
\multicolumn{1}{c|}{\multirow{5}{*}{$256^2$}} & DCGAN & {\bf 20.132} & 9.826 & 0.00 & 1.00 & 1.00 & 1.00 & 1.00 & 0.96\\
 & DCGAN (Durall et al.) & 29.813 & 0.786 & 0.01 & 0.50 & 0.50 & 1.00 & 0.99 & 0.99 \\
 & DCGAN, Spectral & 20.863 & {\bf 0.229} & {\bf 0.50}&{\bf 0.50}& {\bf 0.50} &  {\bf 0.89}&{\bf 0.93}&{\bf 0.85} \\ 
 & LSGAN & 19.893 & 5.955 & 0.00 & 1.00 & 1.00 & 1.00 & 1.00 & 0.85\\
 & LSGAN, Spectral & {\bf 19.884} & {\bf 0.216} & {\bf 0.46}  &{\bf 0.50} &{\bf 0.50}  & {\bf 0.77}&{\bf 0.82}&{\bf 0.72}\\
\bottomrule


    \end{tabular}
\end{table*}

\begin{table}
    \caption{Evaluation of the finetuned StyleGAN2.}
    \label{tab1-stylegan}
    \small
    \centering
    \begin{tabular}{l|l|ccc}
        \toprule
        &&\multicolumn{3}{c}{GAN Quality}\\
        Size & Model & FID $\downarrow$ & SD $\downarrow$ & CS $\uparrow$ \\
        \midrule
            \multicolumn{1}{c|}{\multirow{2}{*}{$1024^2$}} & StyleGAN2 & {\bf 2.733} & 32.250 & 0.09 \\
            & StyleGAN2, Spectral & 3.326 & {\bf 5.978} & {\bf 0.24} \\
        \bottomrule
    \end{tabular}
        
\end{table}

Experiments are conducted on the FFHQ dataset, which contains $70$~$000$ high-quality face images at $1024 \times 1024$ resolution showing large variations in terms of age, ethnicity, and backgrounds.
From this dataset, we downsample three versions in resolution $64\times64$, $128\times128$ and $256\times256$.
We evaluate all experiments on 10k examples in terms of FID and the proposed cloaking score (CS). If the distribution of the real data is matched in both spatial and frequency domain, the resulting FID should be low while the cloaking score should be high. Additionally, we report the sum of absolute differences between the average profile of 10k generated images with the average profile of all real images, denoted as \textit{spectral difference} (SD).
This measure is similar to the optimization target in~\cite{margret2020upconvolution}.

First, we validate the impact of adding additional convolutional layers and increasing the filter size to $5\times5$ as suggested by \citet{margret2020upconvolution} on DCGAN and LSGAN (see Tab.~\ref{tab:taborig}).
In both cases, the modification leads to an improved FID and a lower spectral difference. Yet, the cloaking score is even lower than the one of the baseline approach, indicating that the generated distributions are still linearly separable from their integrated frequency spectra ($AI$, Eq.~\eqref{eq:AI}). This leads to two observations: (1) Increasing the amount of convolutions at the highest resolution can increase image quality of simple GAN methods, and (2) the conventional spatial discriminator does not provide the necessary signal to bring the spectral distributions closer together.
From these observations, we continue with the modified networks to investigate whether the proposed spectral discriminator can provide the necessary training signal, and is able to use the network capacity, to assimilate the generated images' frequency spectra to the real distribution.

To evaluate the proposed spectral discriminator $D_F$ in the context of diverse losses, we train variants of \mbox{DCGAN} \cite{radford2015dcgan}, LSGAN~\cite{mao2017least}, WGAN~\cite{arjovsky2017wasserstein} and WGAN~GP~\cite{gulrajani2017improved} with and without $D_F$.
 While the modified DCGAN and LSGAN models can generate images at up to $256^2$ pixel resolution, the training behavior of the Wasserstein GANs becomes easily unstable. Thus, we only consider them until $128^2$ pixels resolution.
Additionally, to evaluate scalability to high resolutions, we finetune a pretrained state-of-the-art model, StyleGAN2, after adding $D_F$ to its training process.

\begin{figure}
    \begin{subfigure}[t]{.49\columnwidth}
        \centering
        \includegraphics[width=\linewidth]{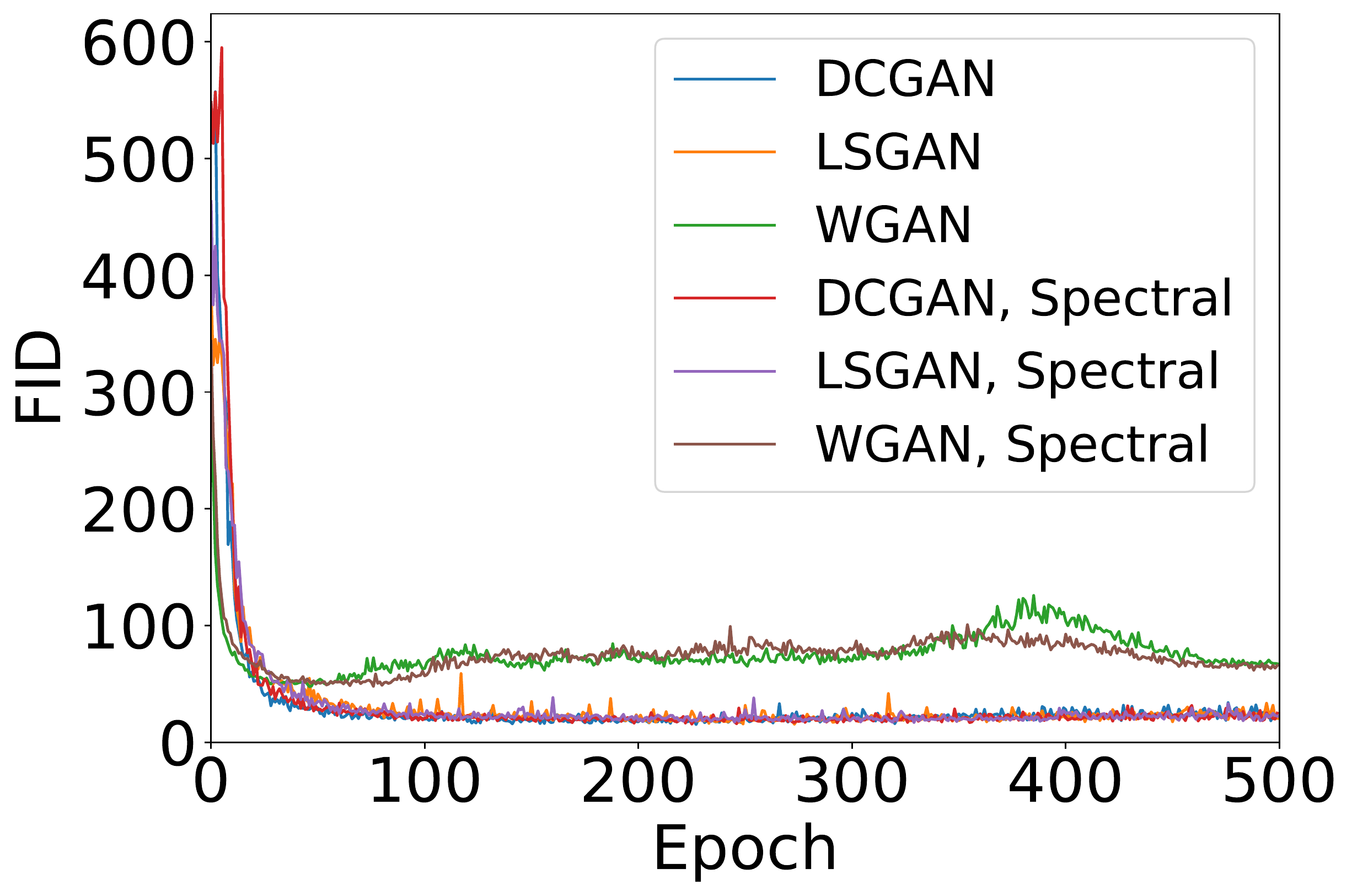}
        \caption{
            The resulting FID scores are similar, and the training is stable.
        }
        \label{fig:profiles}
    \end{subfigure}
    \hfill
    \begin{subfigure}[t]{.49\columnwidth}
        \centering
        \includegraphics[width=\linewidth]{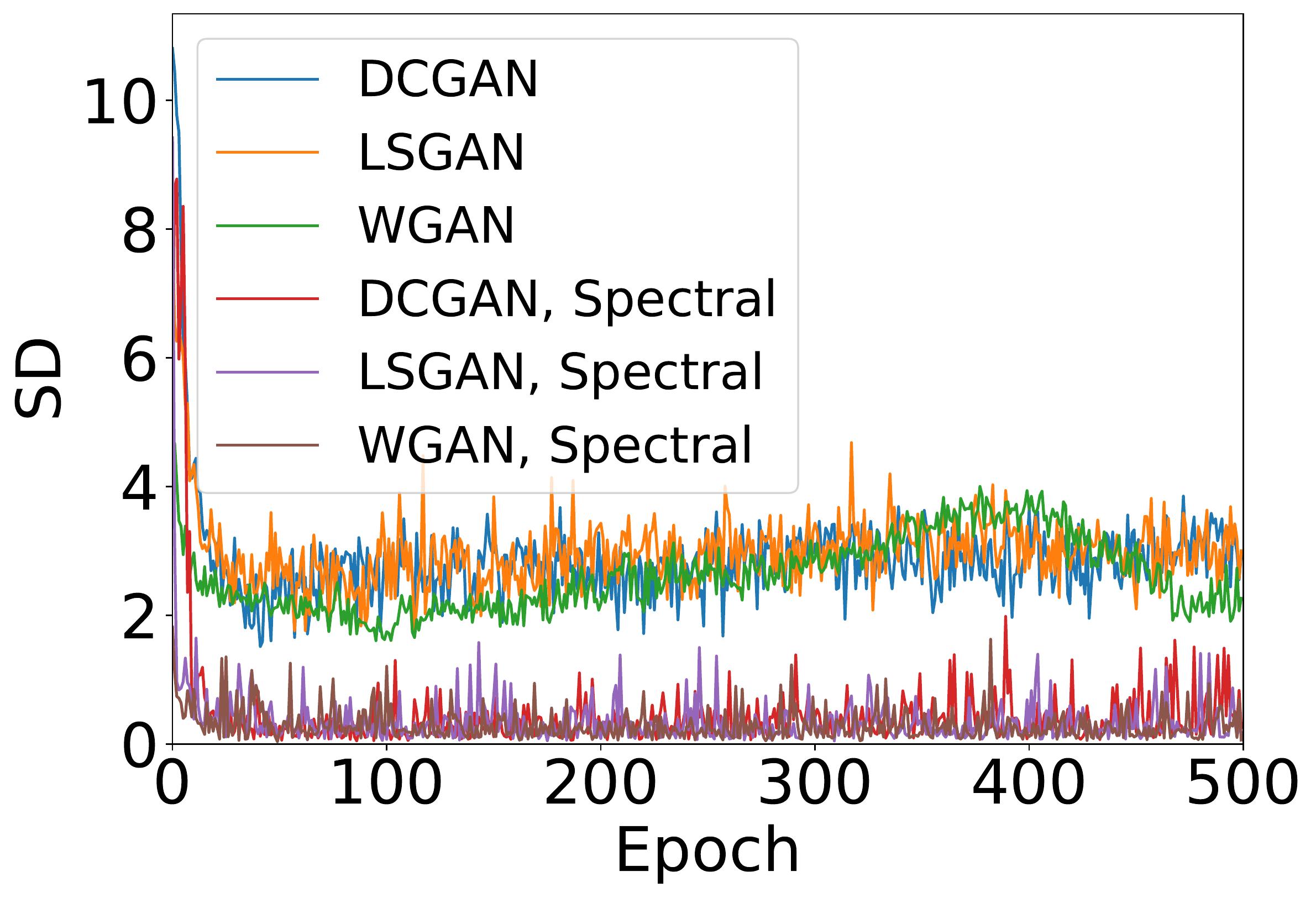}
        \caption{
           With the proposed discriminator $D_F$, the spectral difference is considerably reduced.
        }
        \label{fig:detection}
    \end{subfigure}
    \caption{
        FID and SD with our models trained on FFHQ64.
        \emph{Spectral} indicates that $D_F$ was used, and omitted otherwise.
    }
    \label{fig:cloaking}
\end{figure}
\begin{figure*}[ht]
\tiny
\begin{tabular}{@{\hspace{0cm}}c@{\hspace{0cm}}c@{}c@{}c@{}c@{}c@{}c@{}c@{}}
\includegraphics[width=0.125\linewidth]{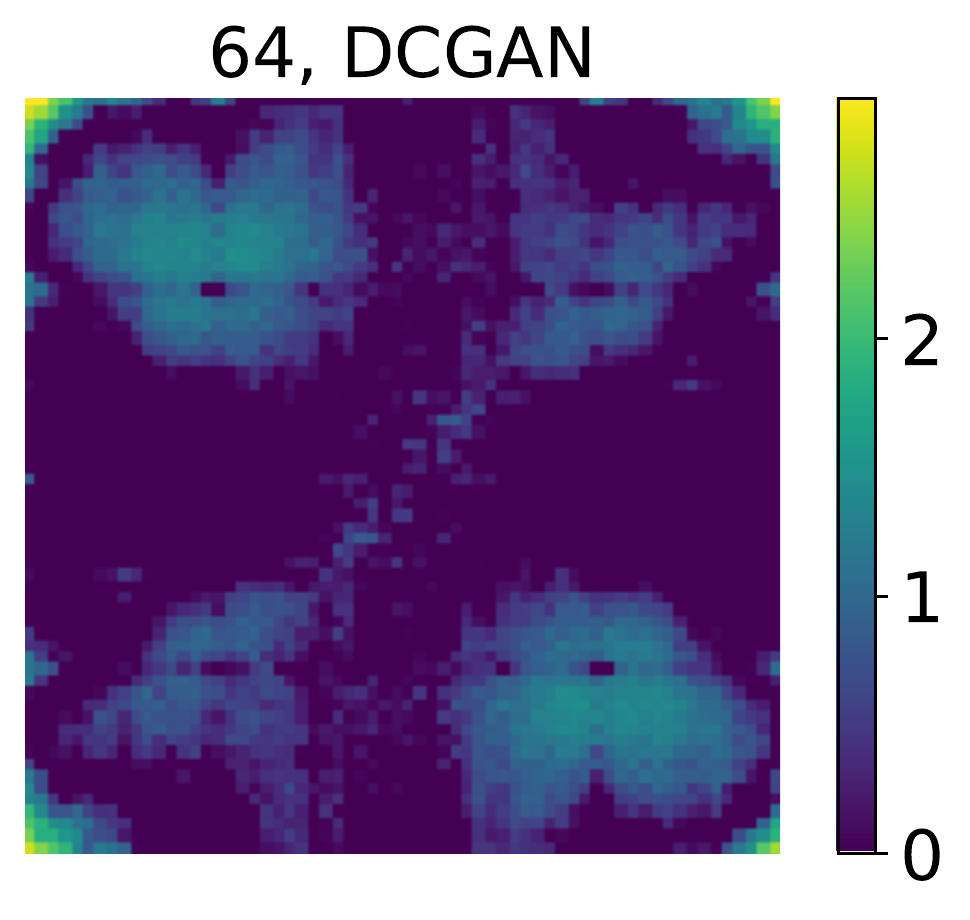}&
\includegraphics[width=0.125\linewidth]{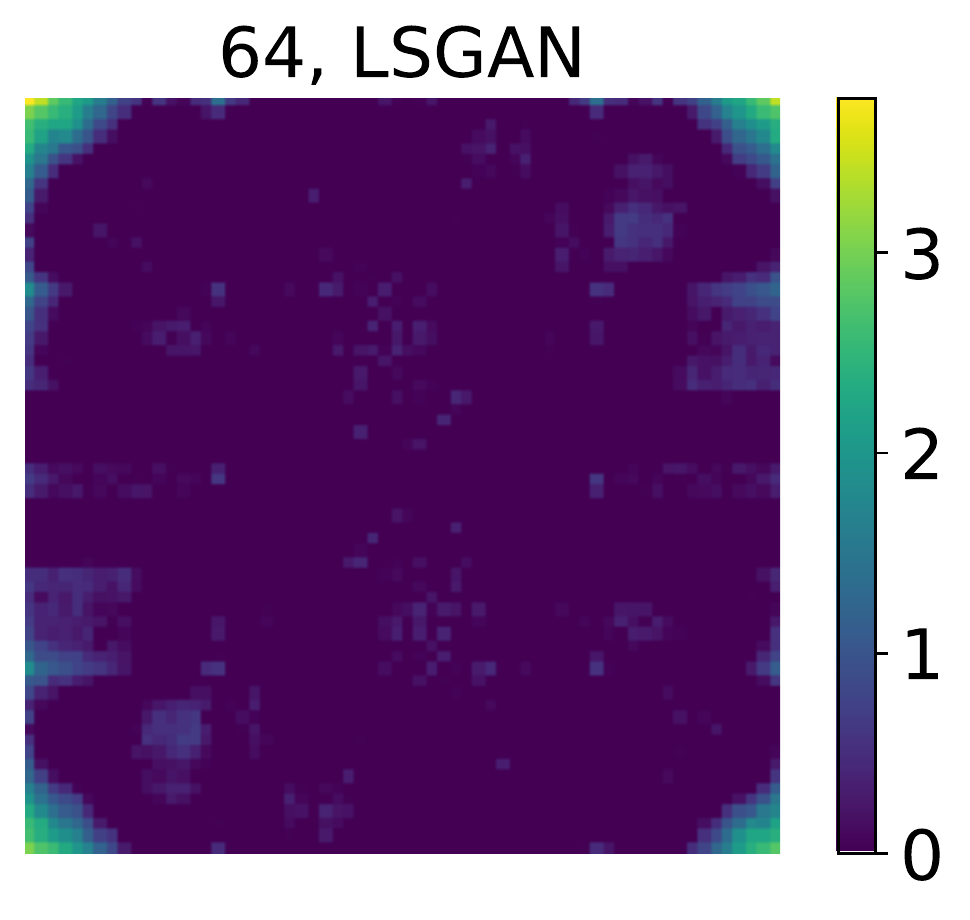}&
\includegraphics[width=0.125\linewidth]{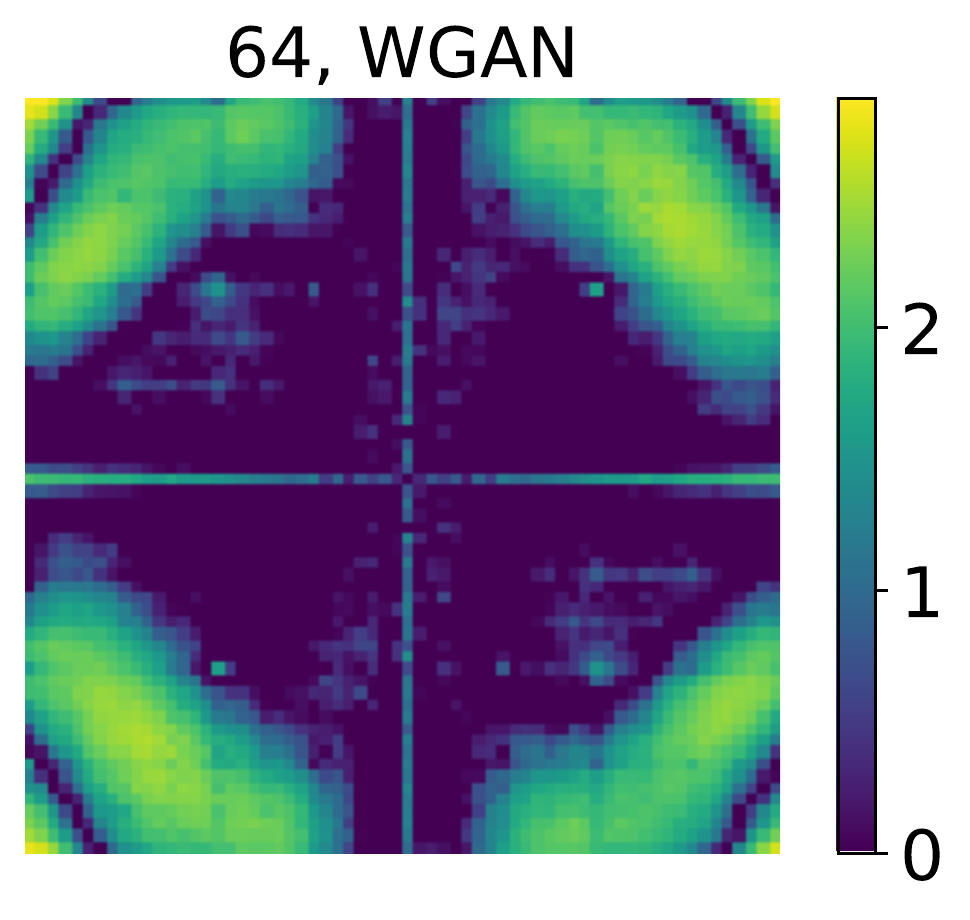}&
\includegraphics[width=0.125\linewidth]{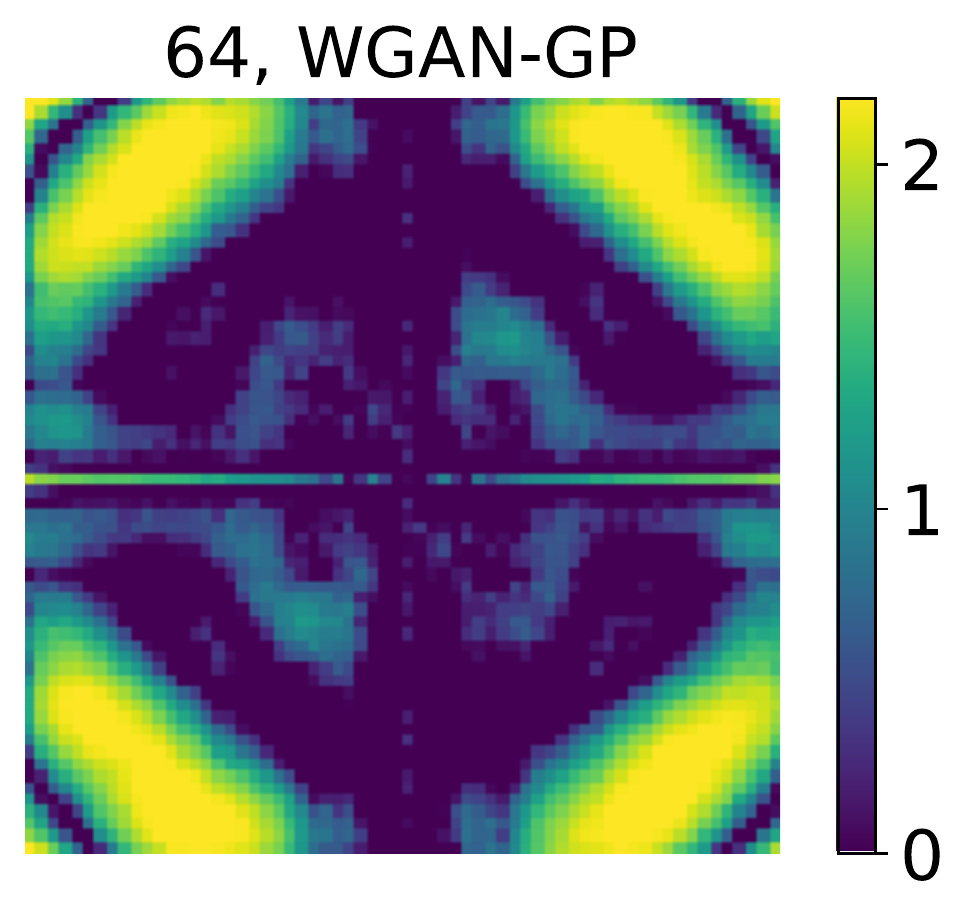}&
\includegraphics[width=0.125\linewidth]{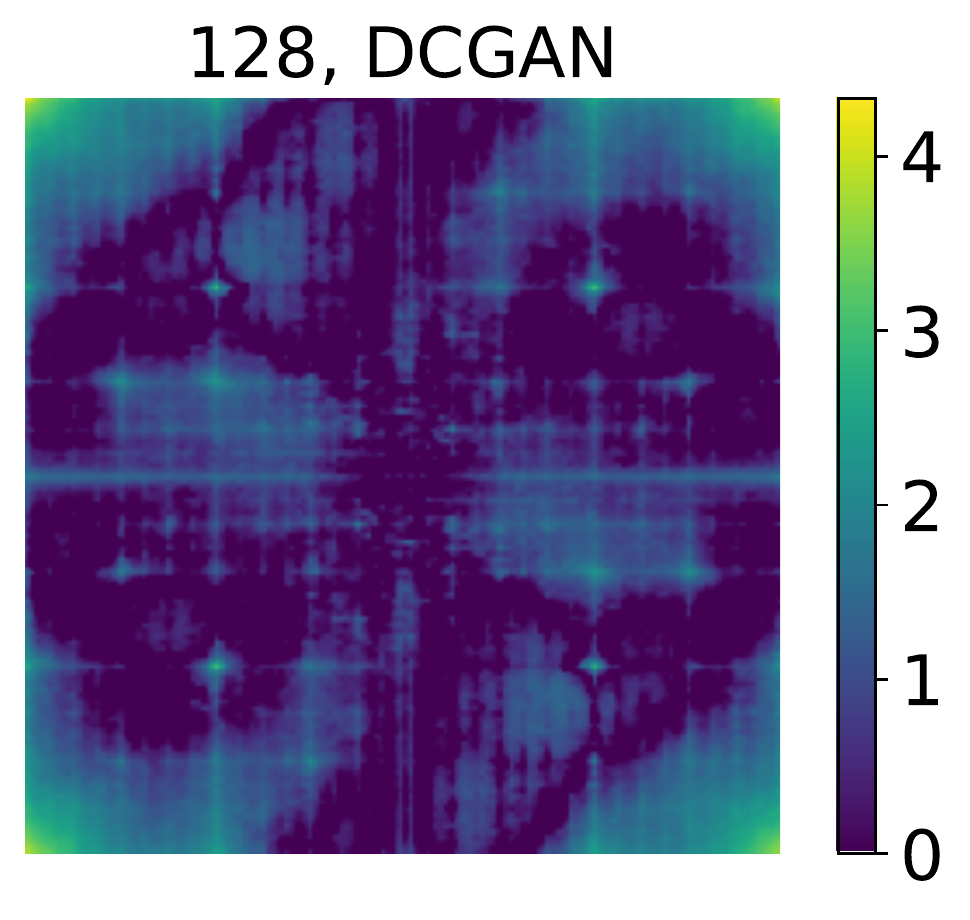}&
\includegraphics[width=0.125\linewidth]{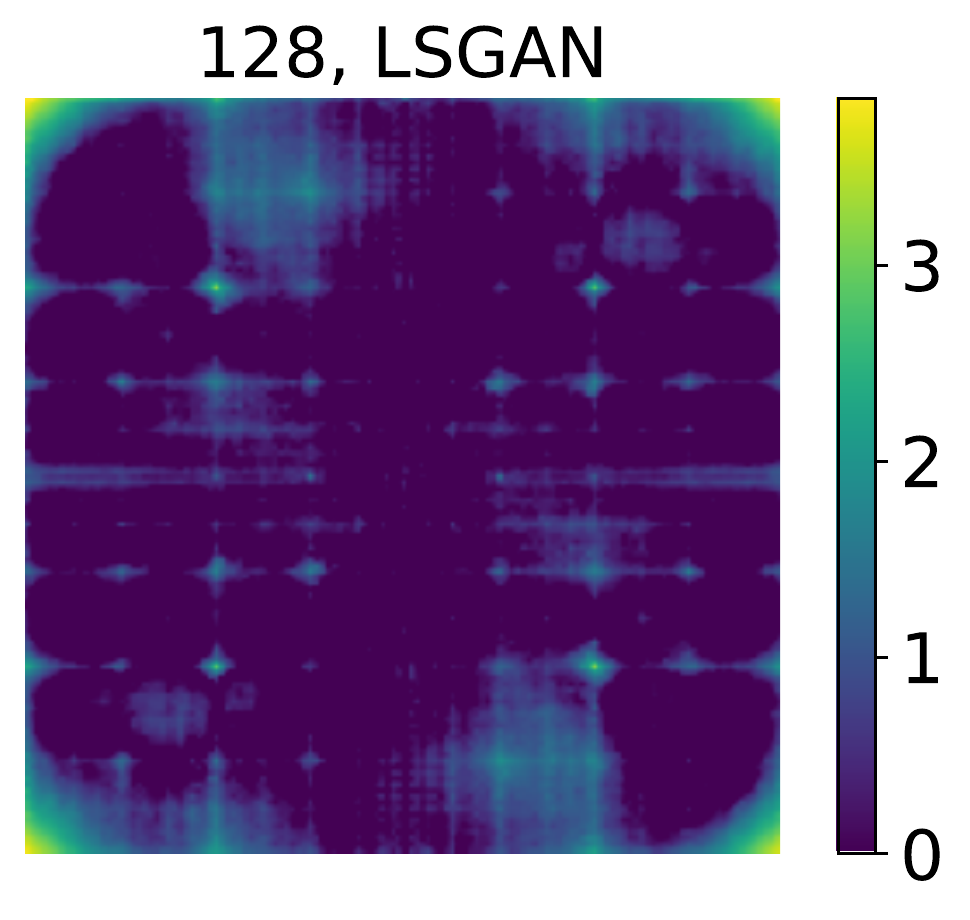}&
\includegraphics[width=0.125\linewidth]{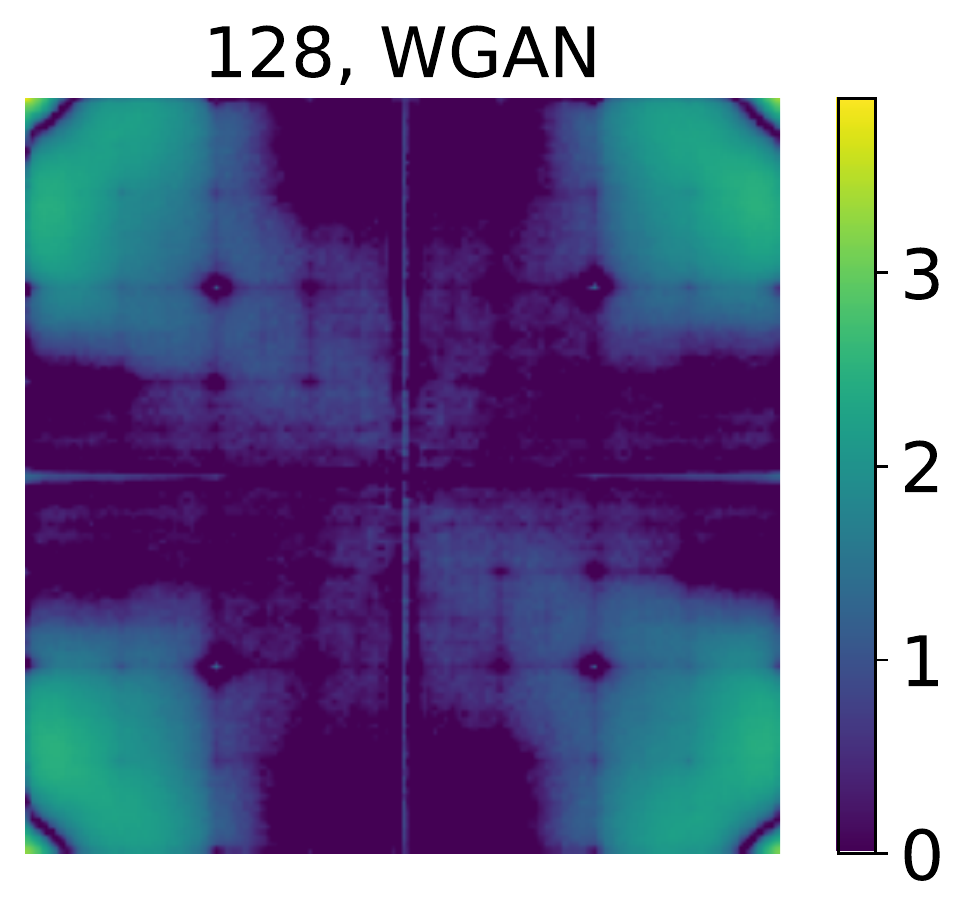}&
\includegraphics[width=0.125\linewidth]{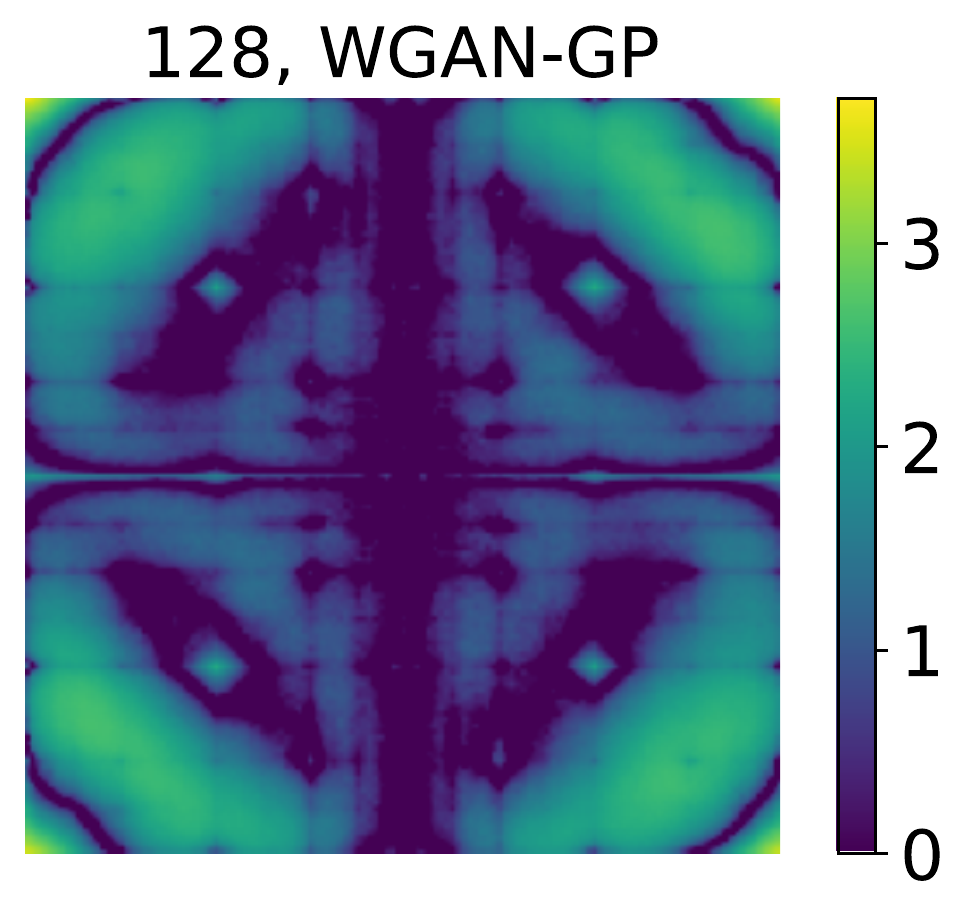}\\
\includegraphics[width=0.125\linewidth]{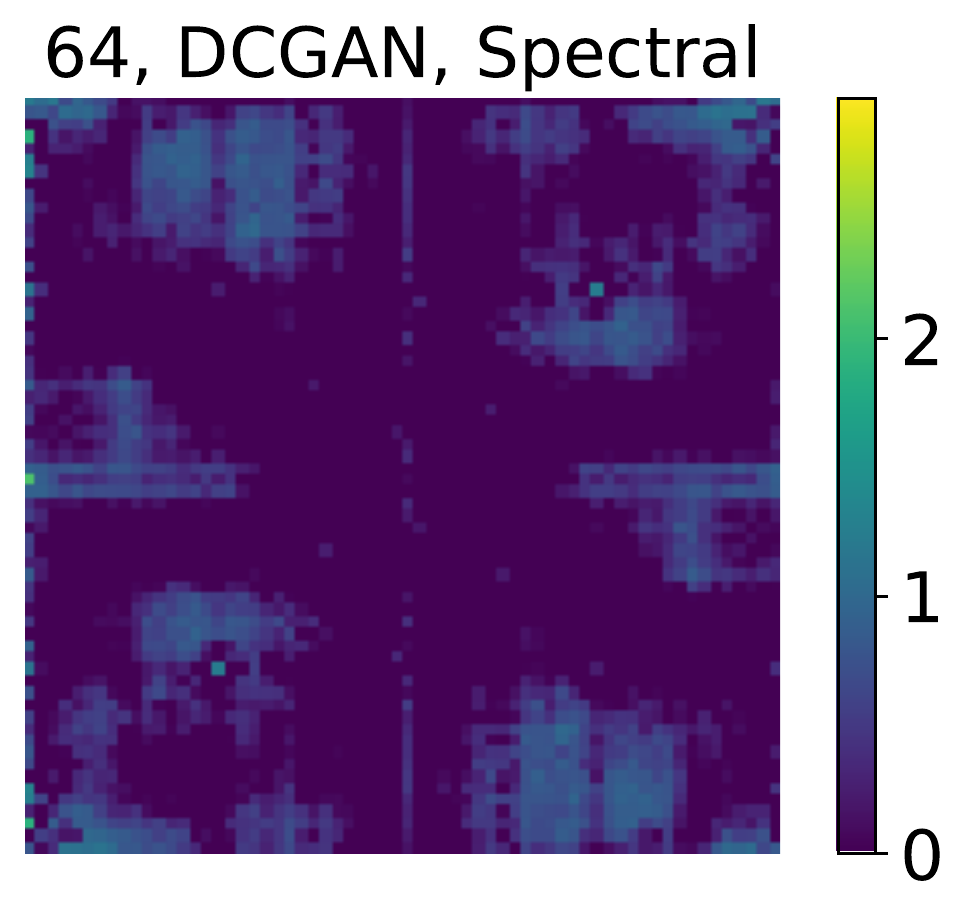}&
\includegraphics[width=0.125\linewidth]{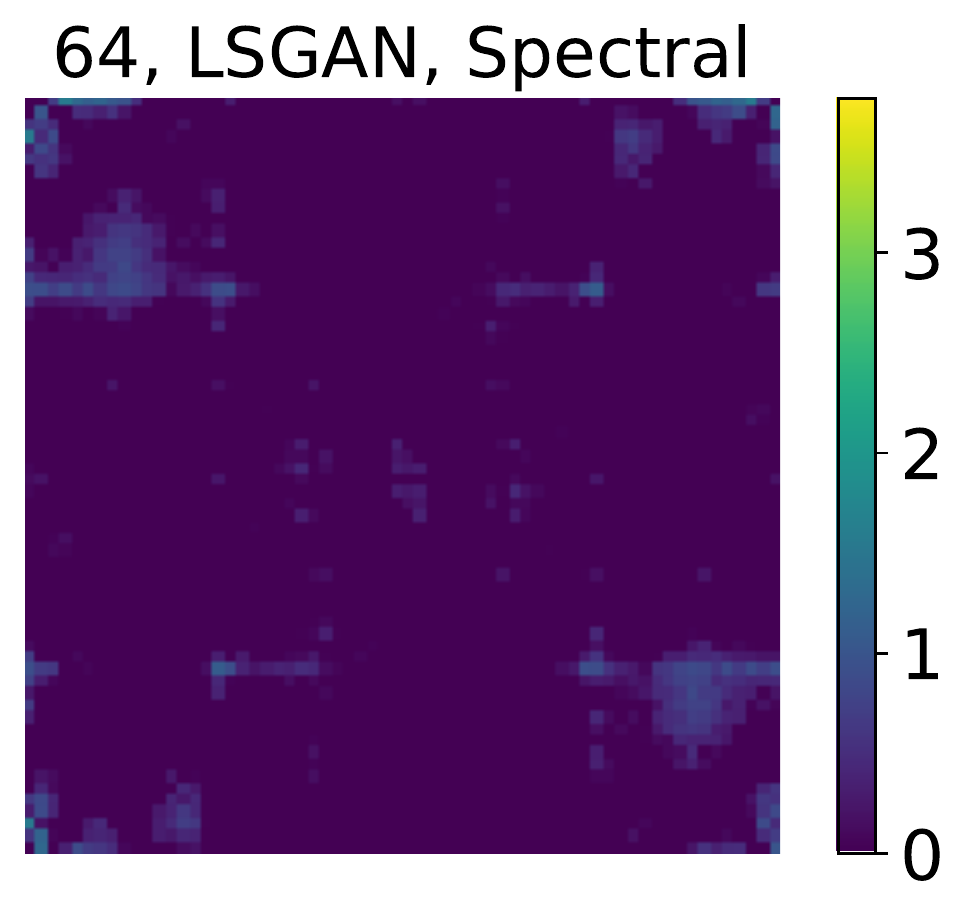}&
\includegraphics[width=0.125\linewidth]{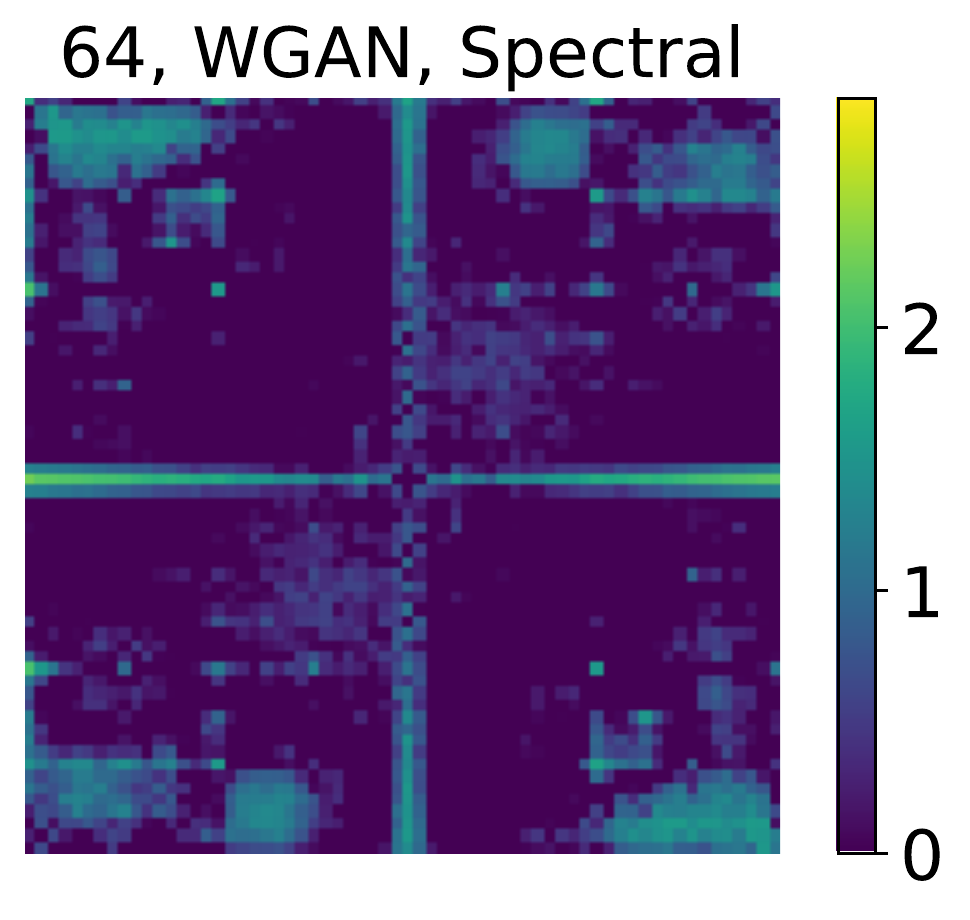}&
\includegraphics[width=0.125\linewidth]{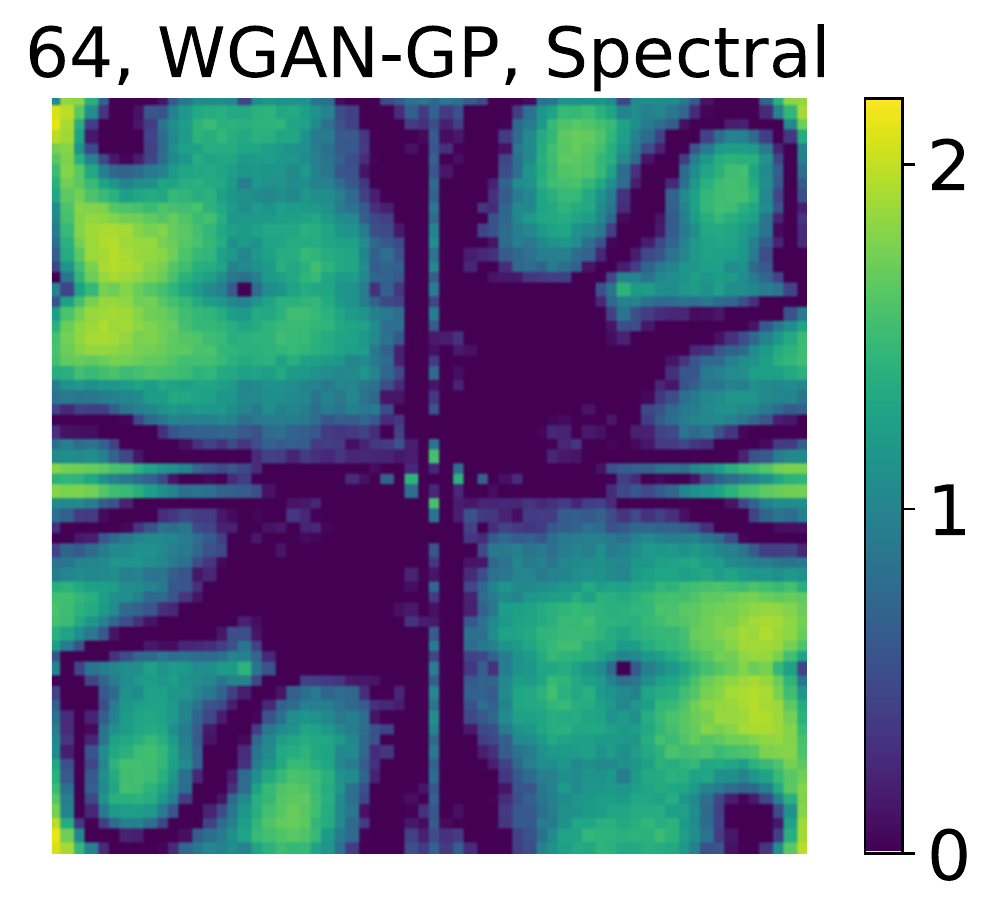}&
\includegraphics[width=0.125\linewidth]{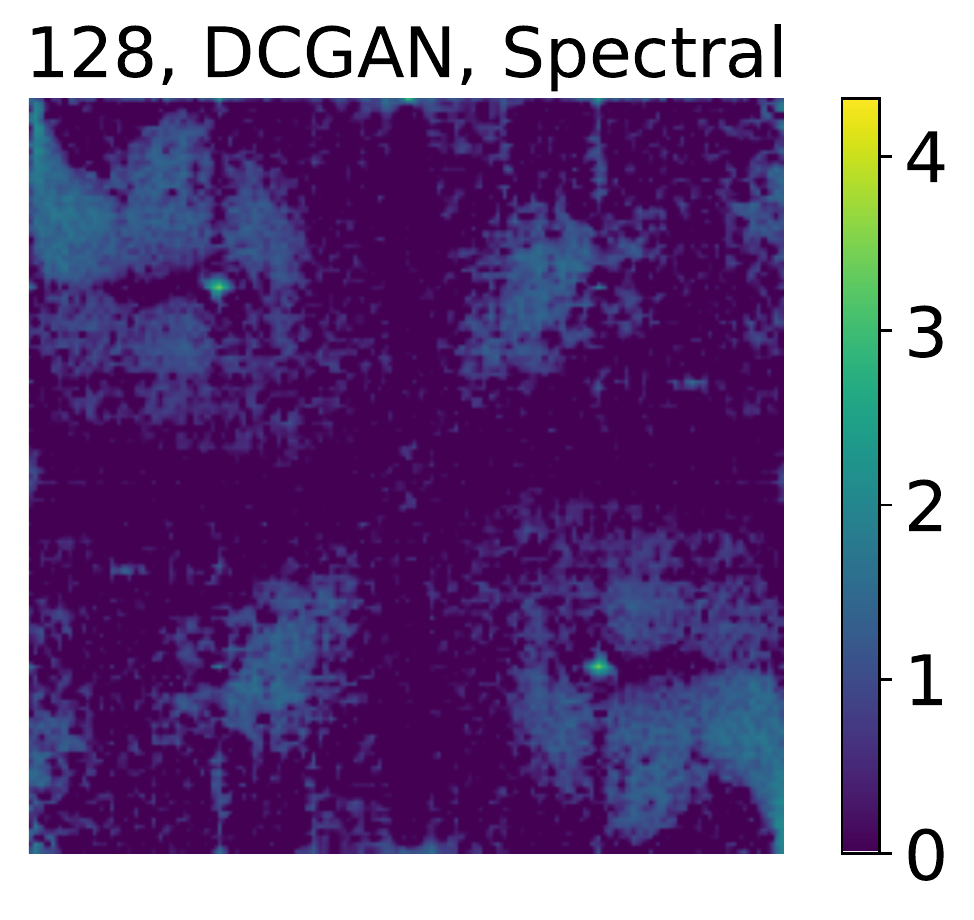}&
\includegraphics[width=0.125\linewidth]{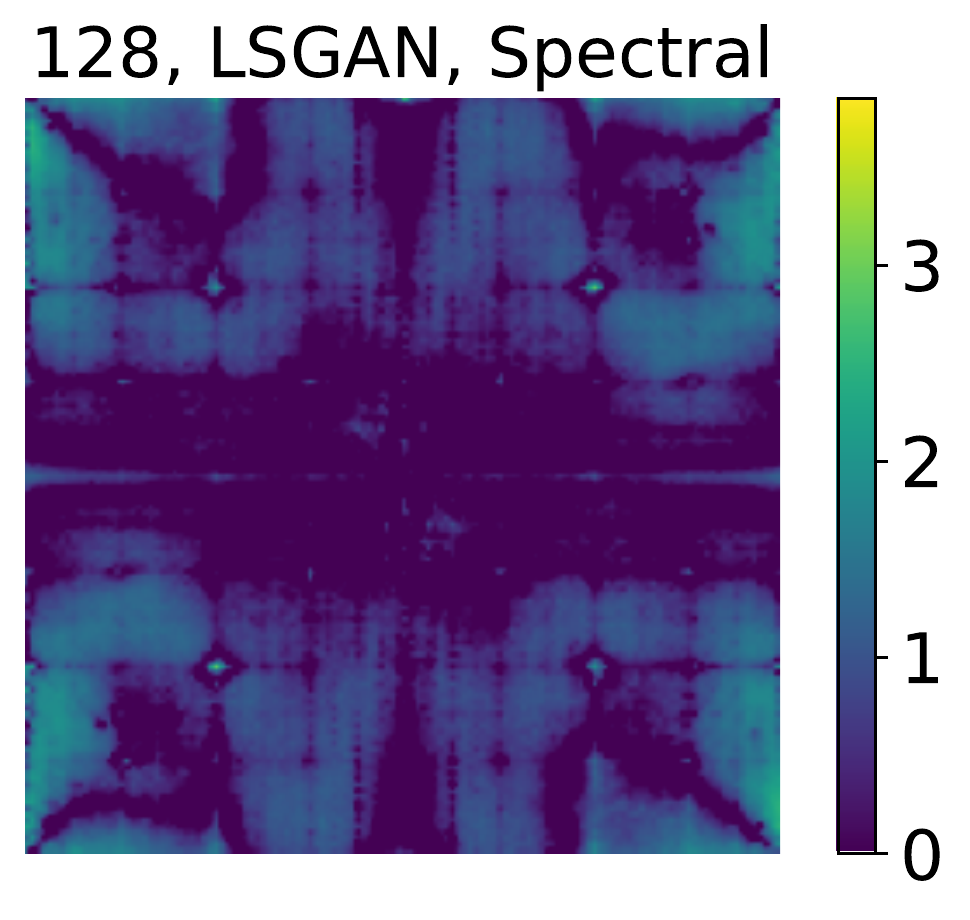}&
\includegraphics[width=0.125\linewidth]{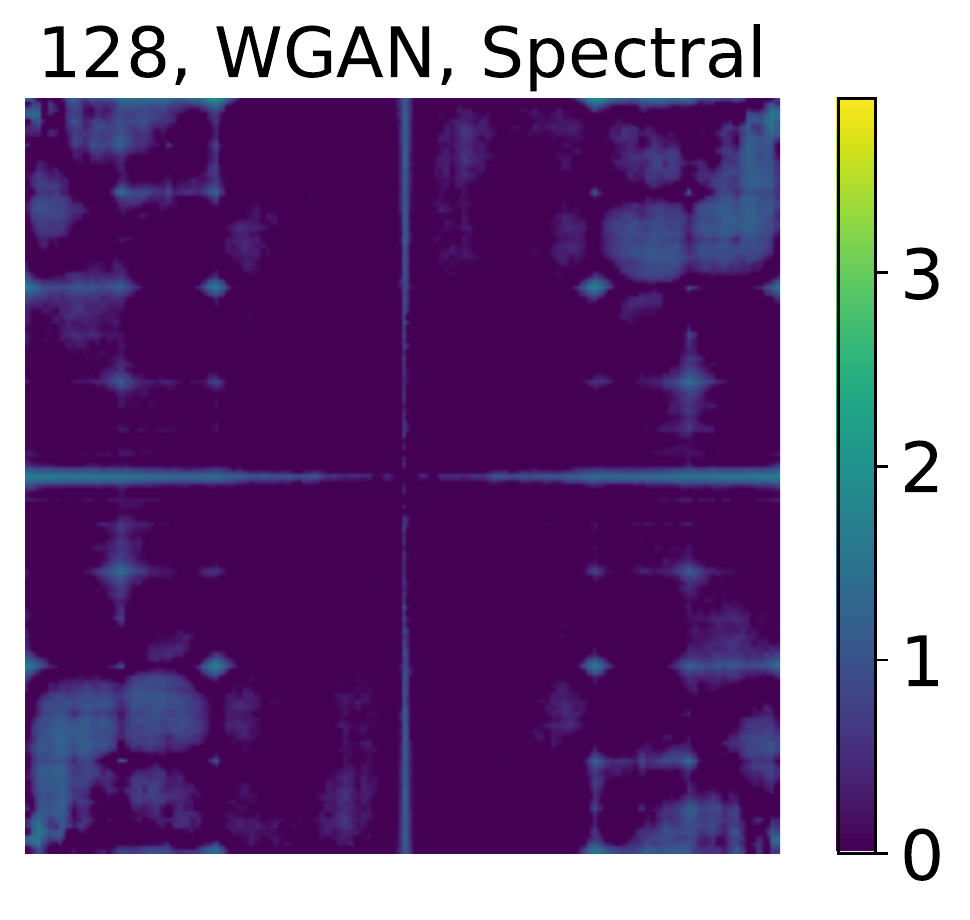}&
\includegraphics[width=0.125\linewidth]{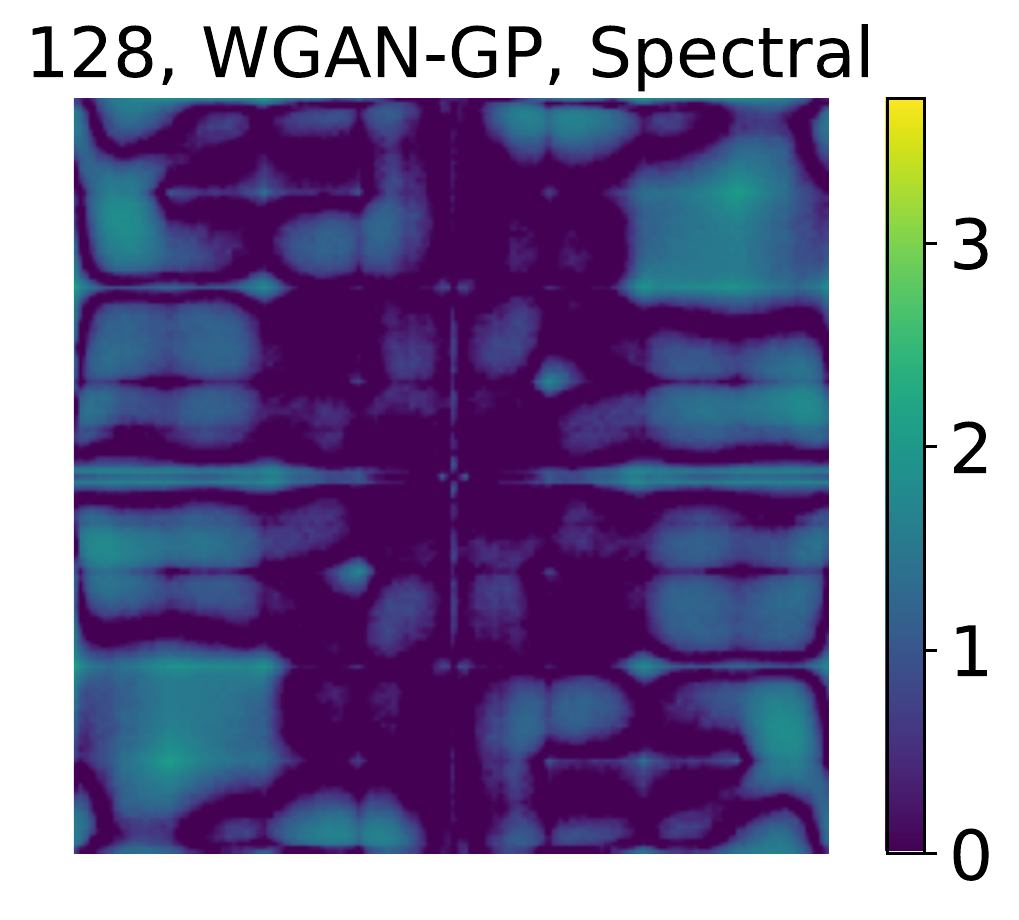}\\
\end{tabular}
\caption{
        Mean absolute differences of the 2D power spectra of real and generated images. (Top) Differences for the original models. (Bottom) Differences for the models with spectral discriminator $D_F$. While the differences of the original models follow a characteristic pattern, our modified models show less specific differences and lower magnitudes.
    }
    \label{fig:spectra}
\end{figure*}
\begin{figure}[!ht]
\scriptsize
\begin{tabular}{@{\hspace{0cm}}c@{\hspace{0cm}}c@{\hspace{0cm}}}
        \includegraphics[width=0.49\linewidth]{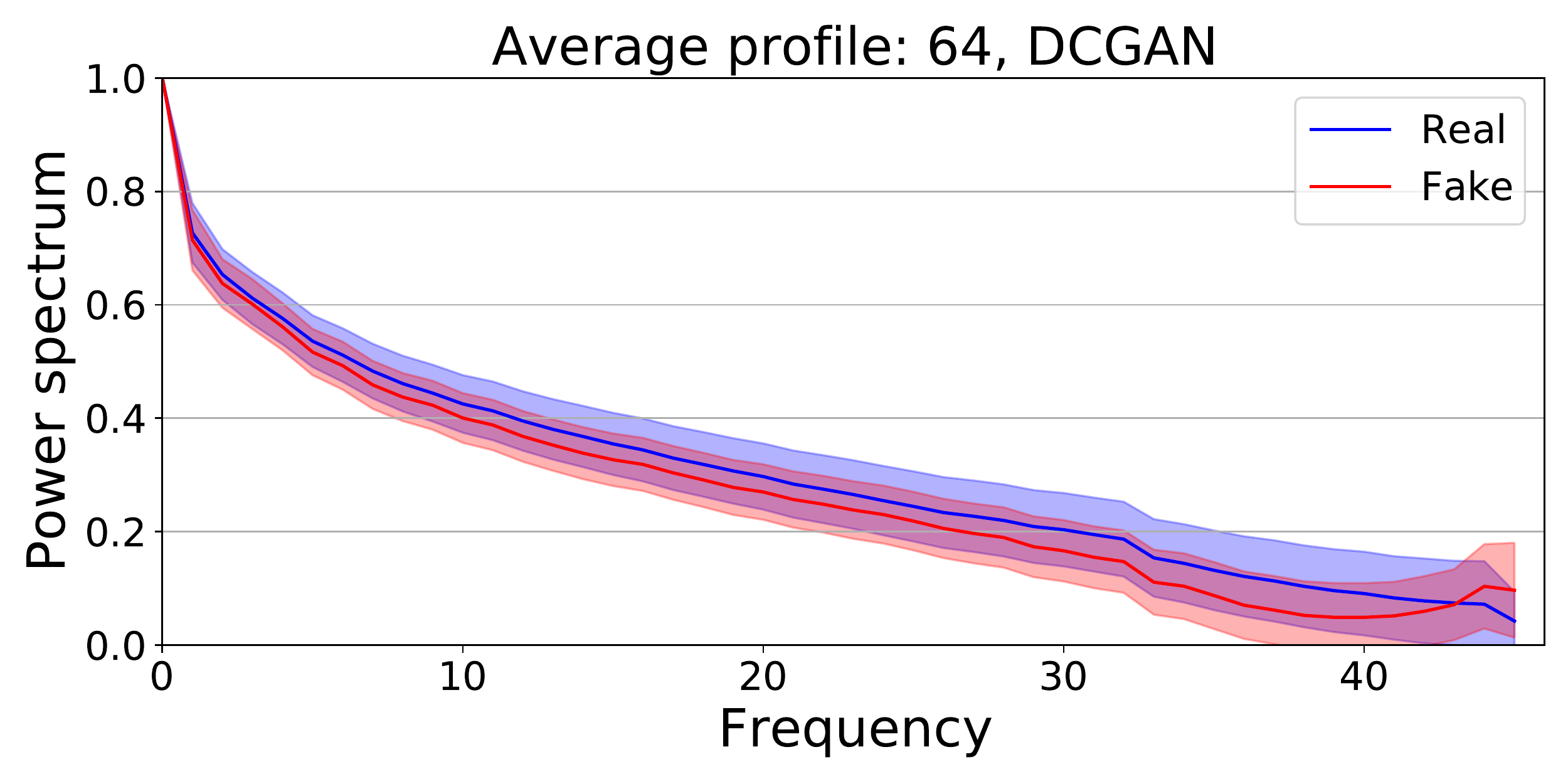}&
        \includegraphics[width=0.49\linewidth]{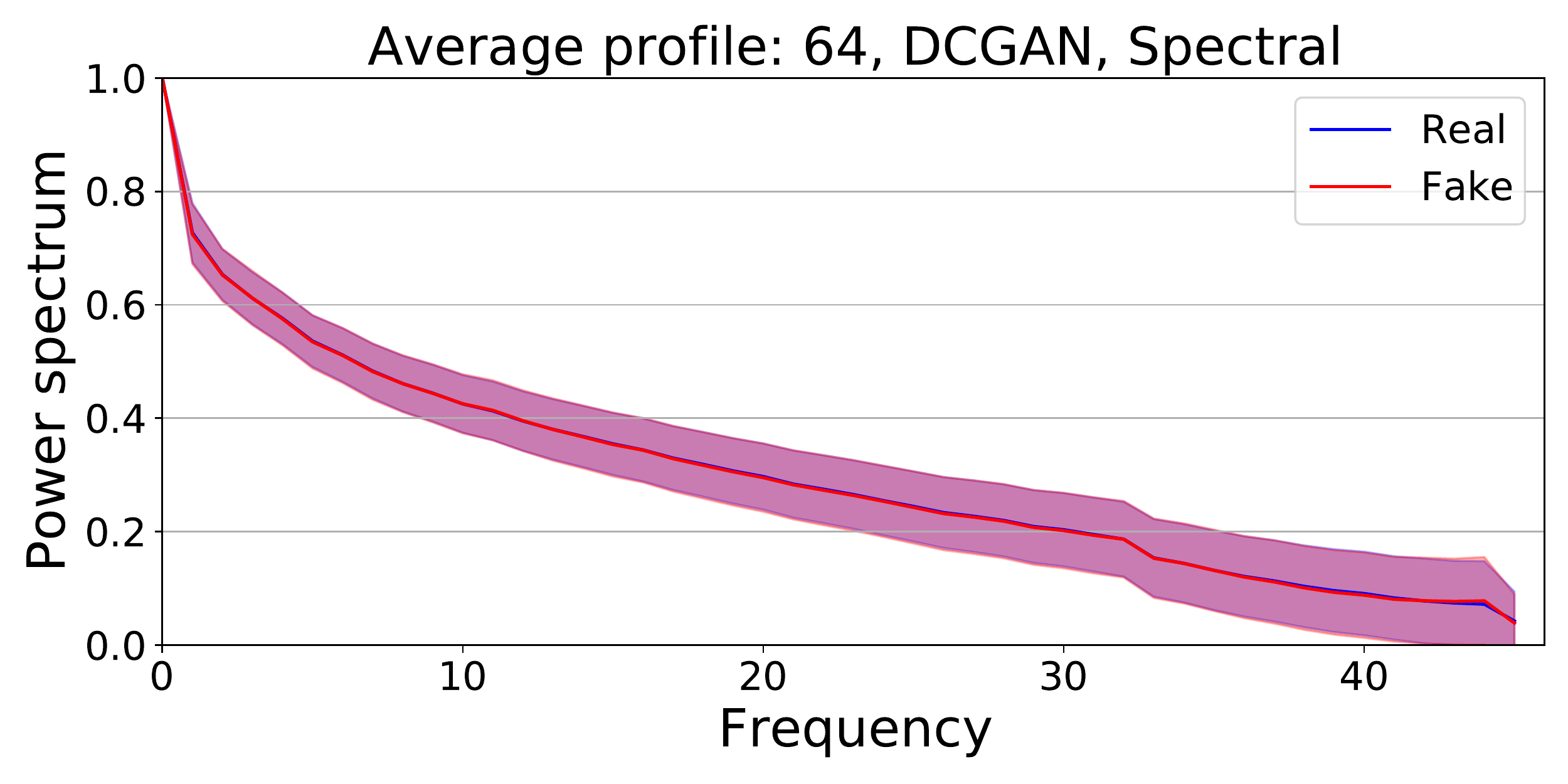}\\
        DCGAN& DCGAN Spectral\\
        \includegraphics[width=0.49\linewidth]{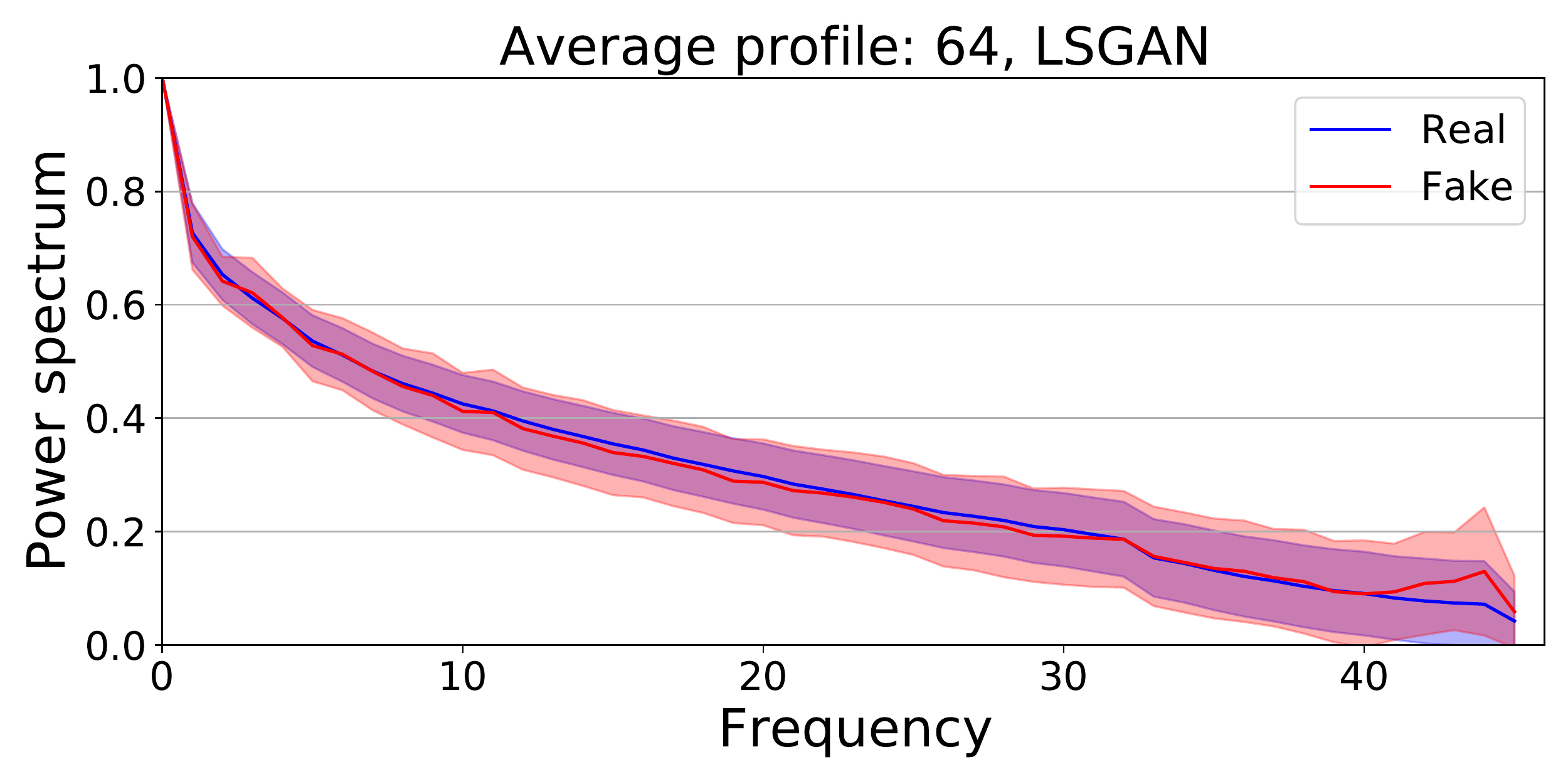}&
        \includegraphics[width=0.49\linewidth]{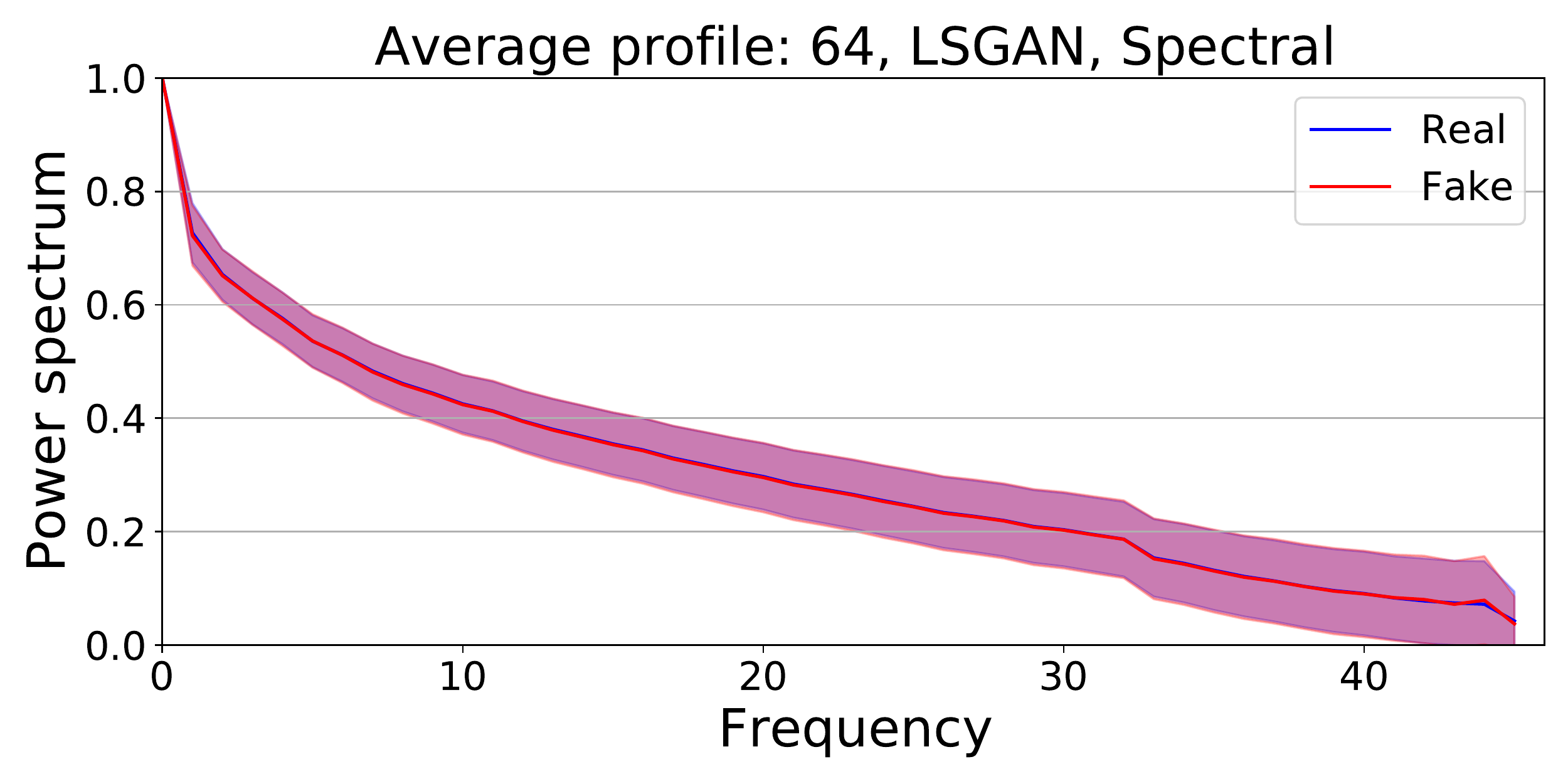}\\
        LSGAN& LSGAN Spectral\\
                \includegraphics[width=0.49\linewidth]{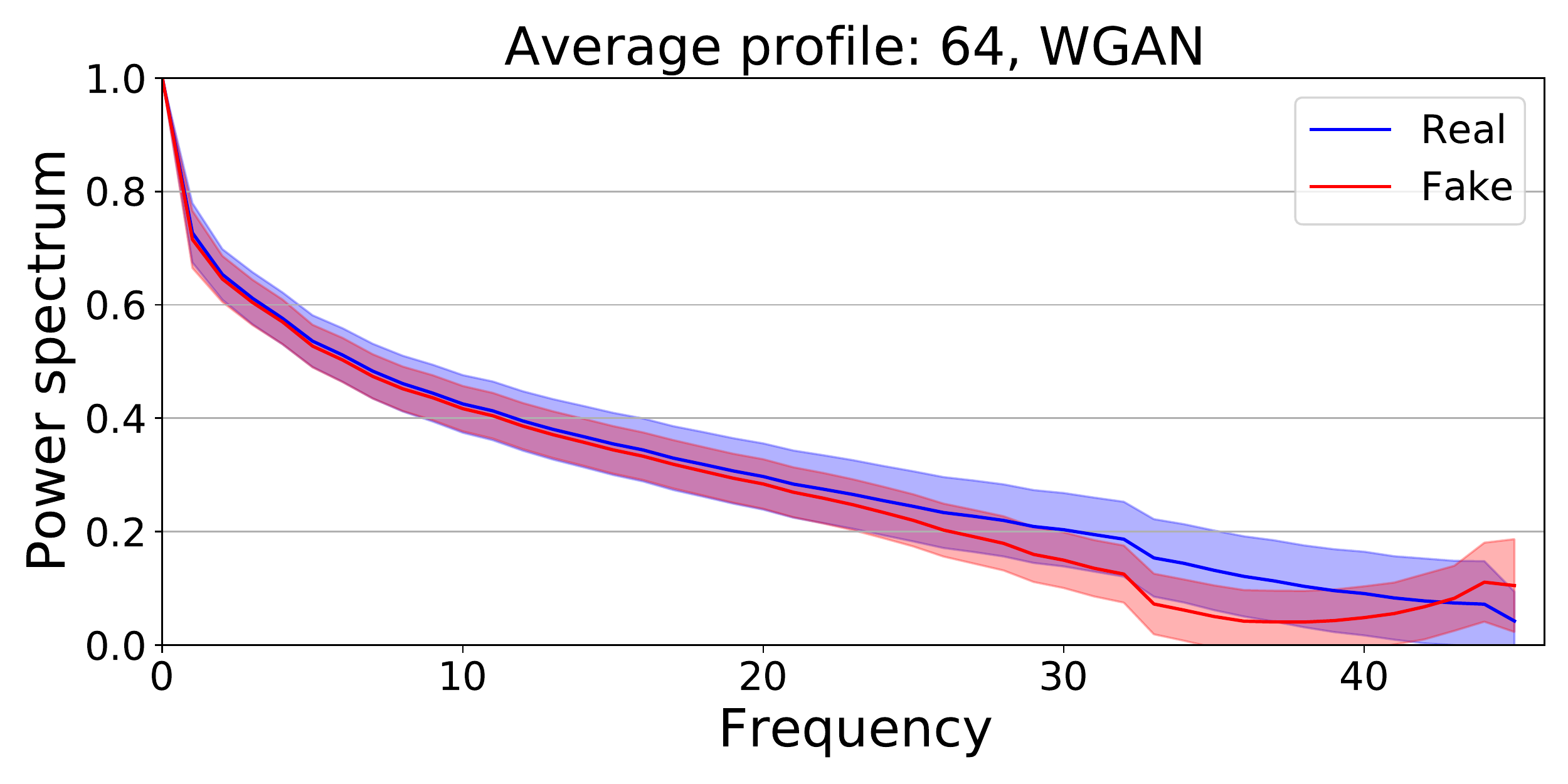}&
        \includegraphics[width=0.49\linewidth]{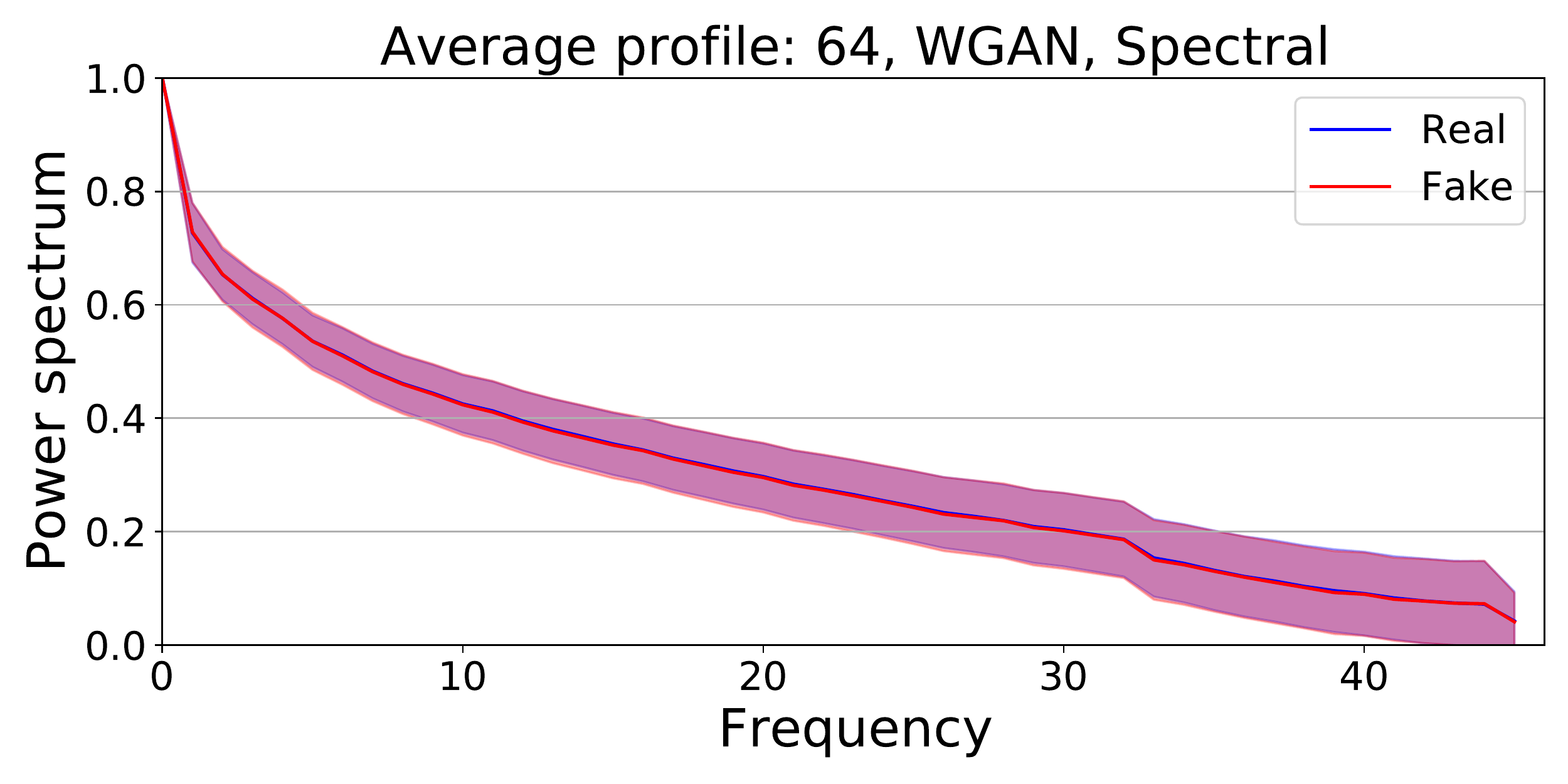}\\
        WGAN& WGAN Spectral\\
        \includegraphics[width=0.49\linewidth]{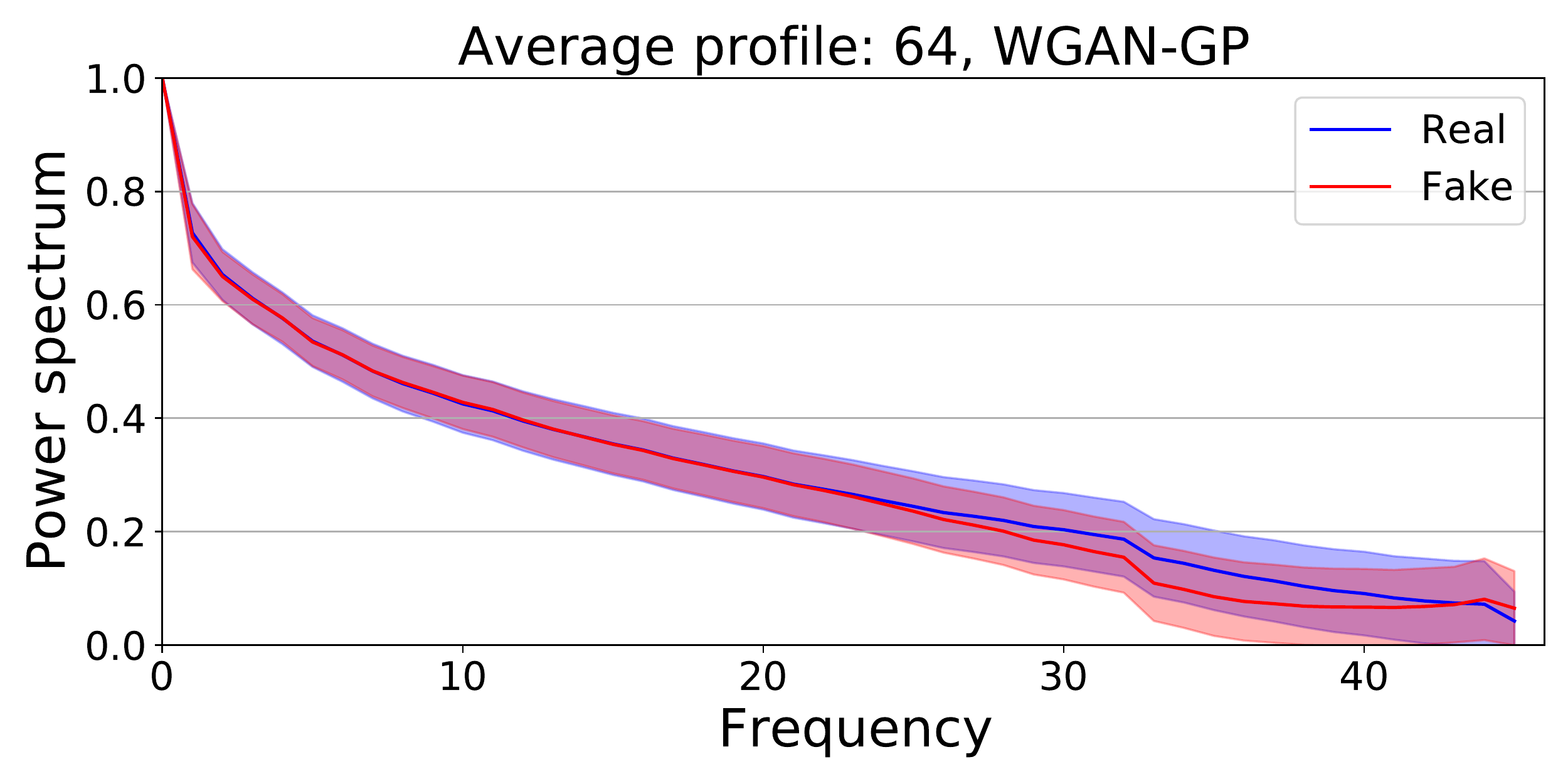}&
        \includegraphics[width=0.49\linewidth]{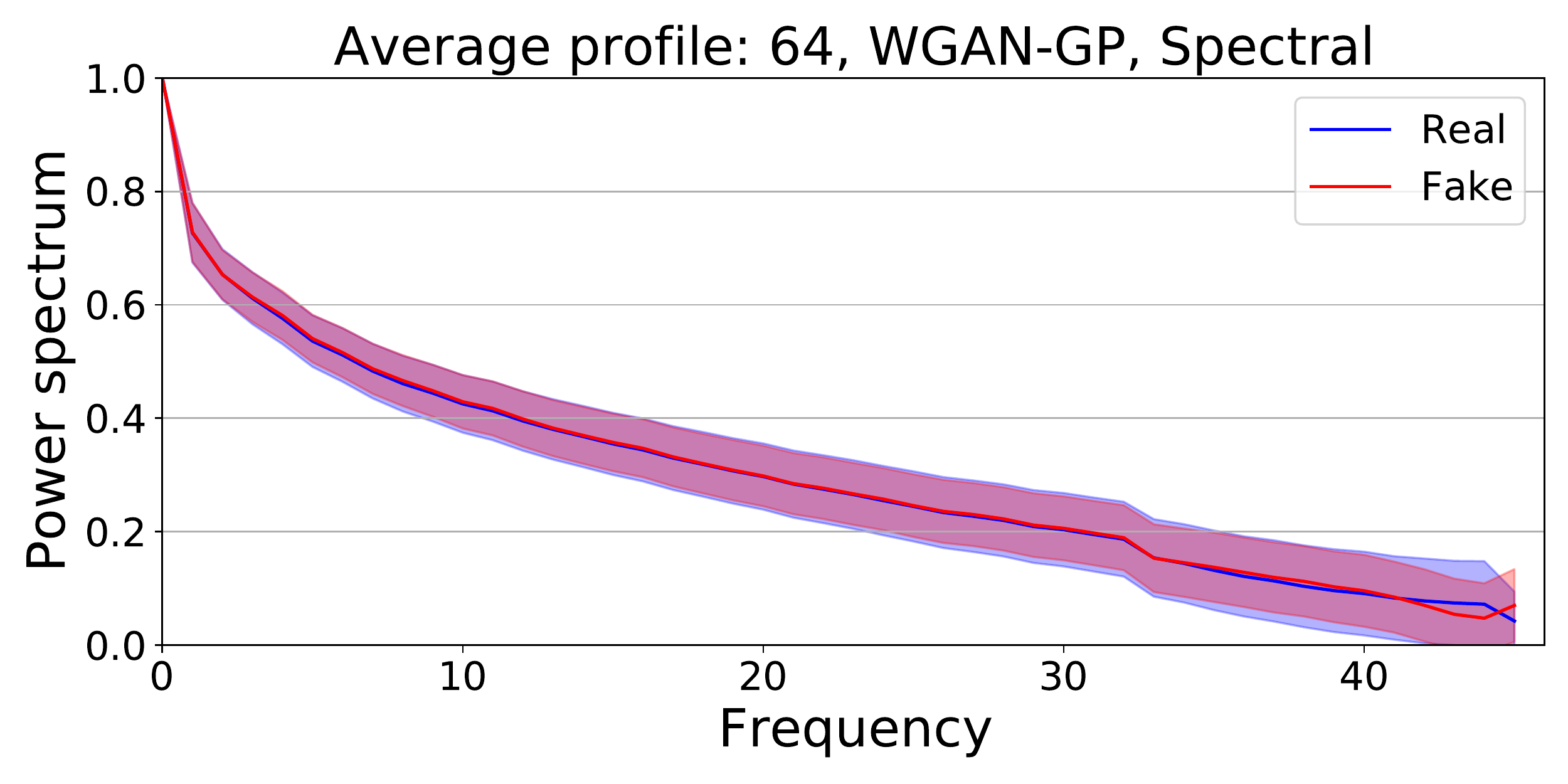}\\
        WGAN-GP& WGAN-GP Spectral\\
        \includegraphics[width=0.49\linewidth]{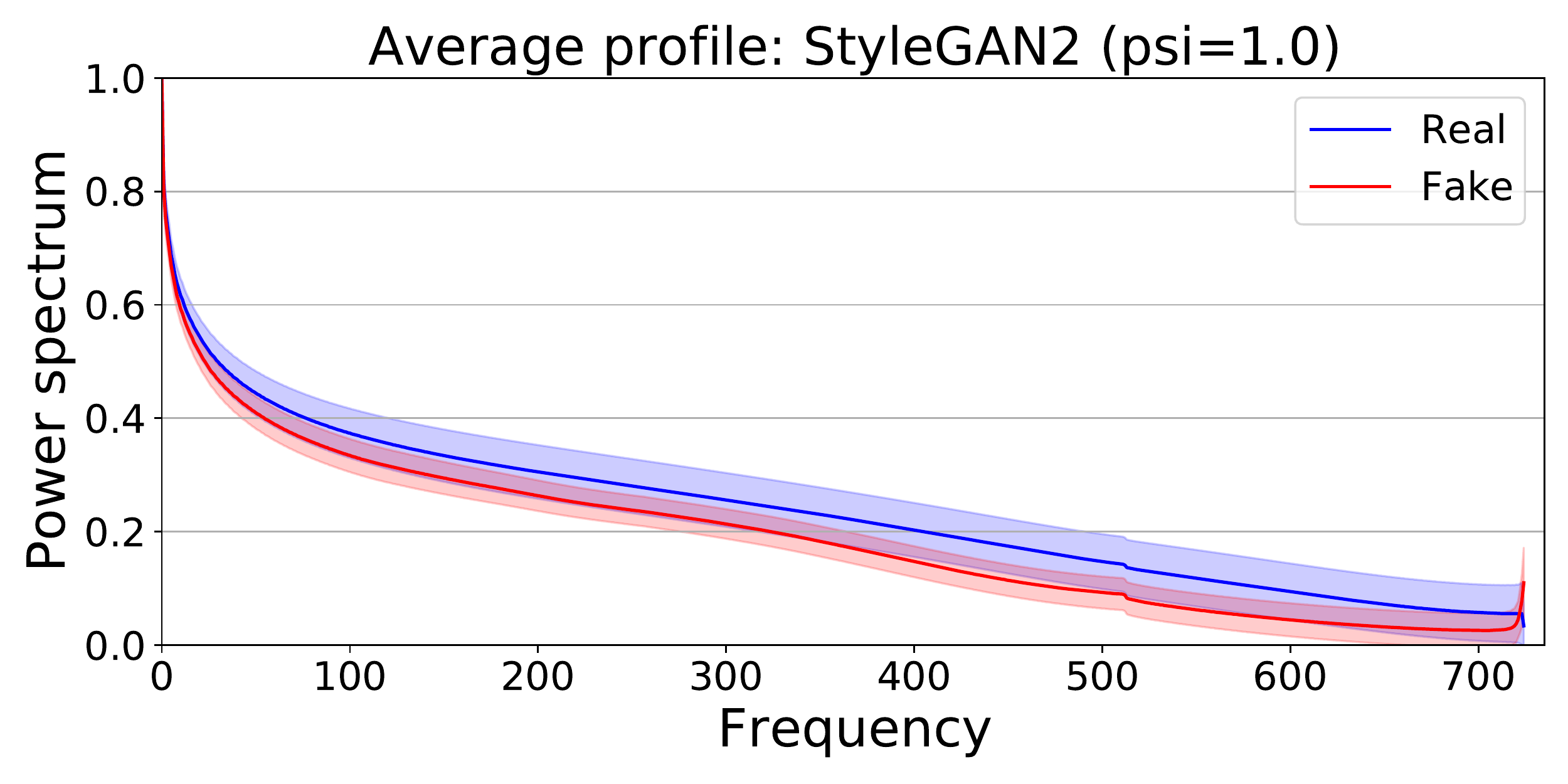}&
        \includegraphics[width=0.49\linewidth]{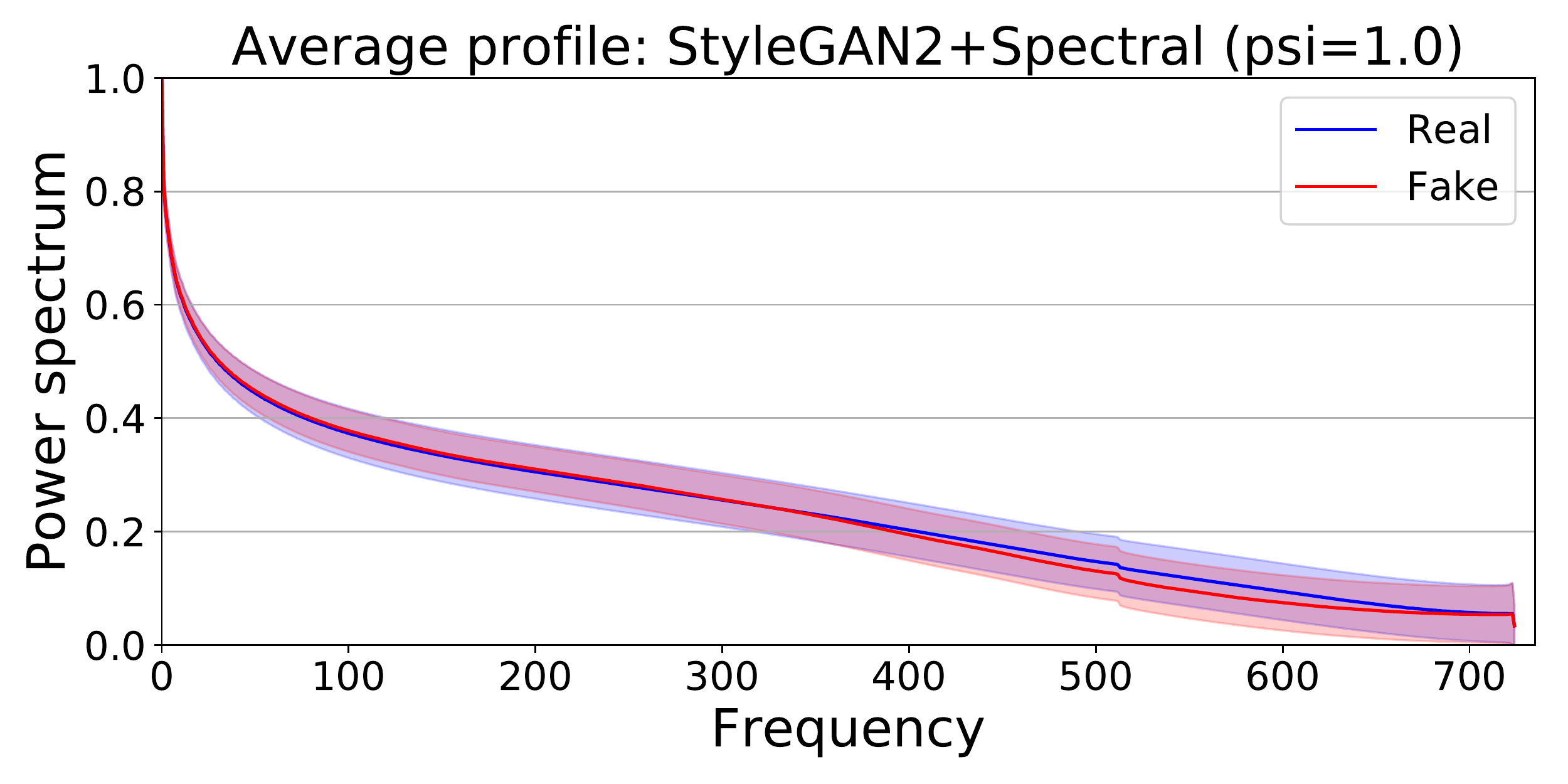}\\
        StyleGAN2 & StyleGAN2 Spectral\\

     %

    \end{tabular}
    \caption{
        Resulting spectral profiles of experiments with our models trained on FFHQ64 and StyleGAN2.
        \emph{Spectral} indicates that $D_F$ was applied. 
        With the spectral discriminator, the mean and standard deviation of the spectral profiles fit almost perfectly.
    }
    \vspace{1em}
    \label{fig:cloaking2}
\end{figure}

The resulting FIDs over the training epochs as well as the spectral differences are displayed in Fig.~\ref{fig:profiles} and Fig.~\ref{fig:detection}, respectively. It can be seen that the proposed discriminator has a relatively small impact on the FID during training while the spectral difference is significantly reduced by our method. The effect on the spectrum can be assessed in Tab.~\ref{tab1} (left block) and in Tab.~\ref{tab1-stylegan}. For all models and resolutions, the FID is on par with the plain version without additional discriminator while the spectral difference is considerably decreased and the cloaking score increased when $D_F$ is added. The spectral regularization proposed by \citet{margret2020upconvolution} yields a significantly higher FID and lower cloaking score than the proposed approach.  

In Tab.~\ref{tab1} (right block), we further evaluate the effect of the proposed discriminator on the detectability of the generated images using recent methods from \citet{margret2020upconvolution} and \citet{Wang_2020_CVPR}. 
While \citet{margret2020upconvolution} leverage the 1D frequency spectra for the detection in a simple support vector machine (SVM) or logistic regression (LR) classifier and require retraining in every setting, \citet{Wang_2020_CVPR} train a CNN on the generated images using various data augmentations as a "universal" detector. We evaluate the method from \citet{margret2020upconvolution} in two scenarios: in the left column in Tab.~\ref{tab1} (right block) indicated with the asterix, we train on the data without spectral discriminator and evaluate in the transfer setting; in the right column, we train separately for every setting.
As can be seen, the detection is almost random in the transfer setting for the regularization approach by \citet{margret2020upconvolution} as well as for our approach in all tested GAN settings and resolutions while the generated images from \cite{margret2020upconvolution} can still be detected with about 80\% accuracy when a model is trained specifically on this data.
With respect to the universal detector by \citet{Wang_2020_CVPR}, both method decrease the detection accuracy significantly. This confirms that the removed frequency artifacts can in fact be perceived in the image domain and their removal is to be desired.

Figure \ref{fig:cloaking2} gives a visualization of the resulting average frequency profiles with our approach as an overlay on the average frequency of real images. With the proposed discriminator, the average frequencies fit almost perfectly. For the respective higher resolution experiments, as well as for example images generated by our approach see the supplementary material. Figure \ref{fig:spectra} shows the average differences of real and generated 2D power spectra in log-space (compare the visualization in \cite{Wang_2020_CVPR}). It can be seen that all tested GANs show specific differences. When the proposed discriminator, acting on the 1D projection of the spectra, is employed, the differences become more diffuse and have lower magnitudes. This indicates that, on average, the generated images have less perceivable sampling artifacts.  

\section{Conclusion}
\label{sec:conclusion}
In this paper, we proposed an adversarial image generation approach that enables to generate images with a significantly reduced amount of spectral artifacts. The proposed method employs a simple discriminator on the 1D projections of frequency spectra of real and generated images. Thus, the generator aims to match the real data distribution not only in the image domain but also in the frequency domain. Our approach is very lightweight and allows for a stable training, as we experimentally show for different loss functions and resolutions. It generates images that can not easily be distinguished from real ones by frequency artifacts and improves, in this respect, over the recent method by~\cite{margret2020upconvolution}.


%
%
%
 \bibliography{manuscrip}

\end{document}


%
%
%
%
%
%
\maketitle              

\begin{figure}
\centering
\begin{tabular}{@{}c@{}c@{}c@{}}
     \imagetop{\includegraphics[width=0.32\linewidth]{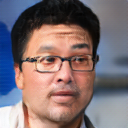}} &
   \imagetop{ \includegraphics[width=0.32\linewidth]{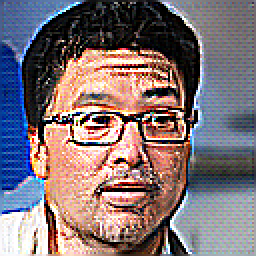} }&
     \imagetop{   \includegraphics[width=0.32\linewidth]{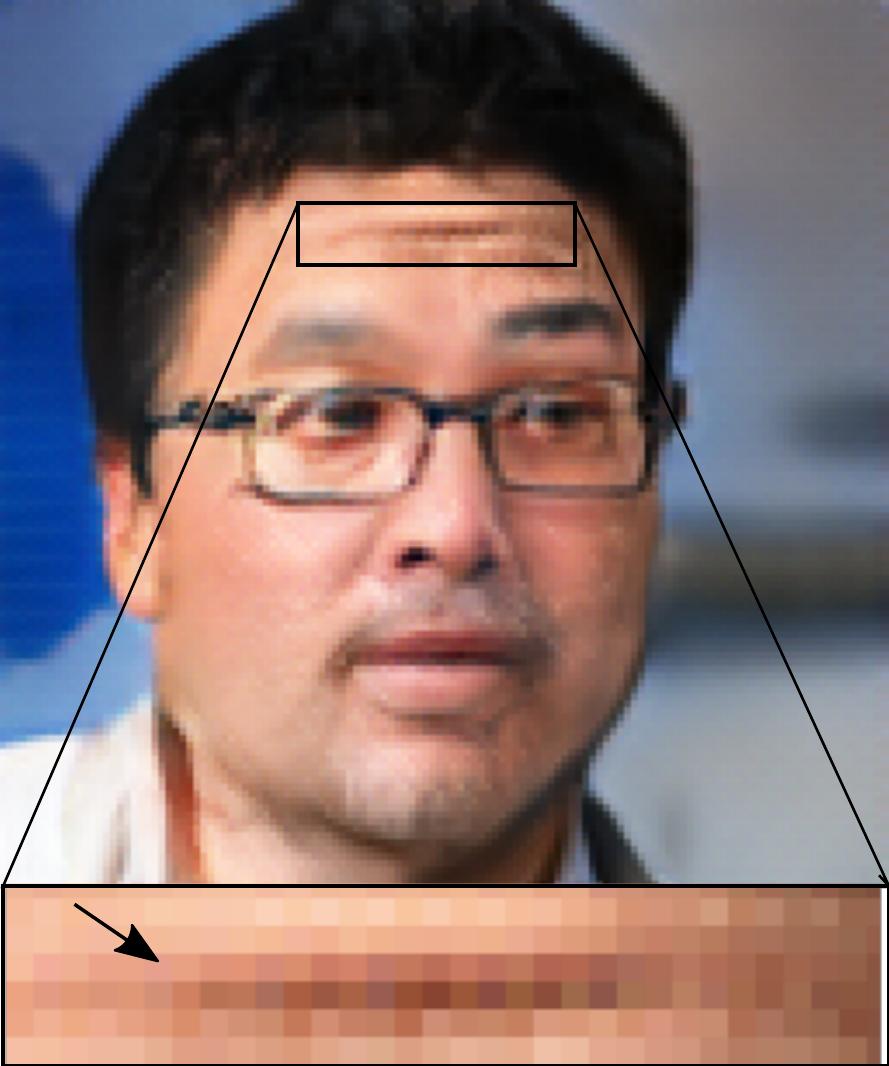}} \\
        (a) DCGAN&(b) sharpened&(c) zoomed in

\end{tabular}

\caption{Sample from DCGAN, $128^2$. Peaks in the high frequencies which we measure in the power spectrum correspond to grid artifacts in the images. After applying an image sharpening operation, they become obvious (b) but at a close look, they can also be perceived in the original generated images (c). }
\label{fig:sharpened}
\end{figure}
\section{High Frequency Artifacts}
In the proposed paper, we show that frequency artifacts in generated images can be significantly removed be using a simple spectral discriminator. Here, we want to emphasize again why a peak in the high frequency regime of the generated images' power spectra are undesired. We provide an example in Figure \ref{fig:sharpened}, showing a face in front of what appears to be a mostly homogeneous background. Yet after applying a sharpening operation (Fig.\ref{fig:sharpened}(b)) , grid artifacts become obvious even in these supposedly flat regions. At a close look, they can also be seen in the plain images without sharpening (Fig.\ref{fig:sharpened}(c)).

\section{Training Stability of Durall et al. (CVPR 2020)}
\begin{figure}
    \centering
    \begin{subfigure}[b]{\linewidth}
        \centering
        \includegraphics[width=\linewidth]{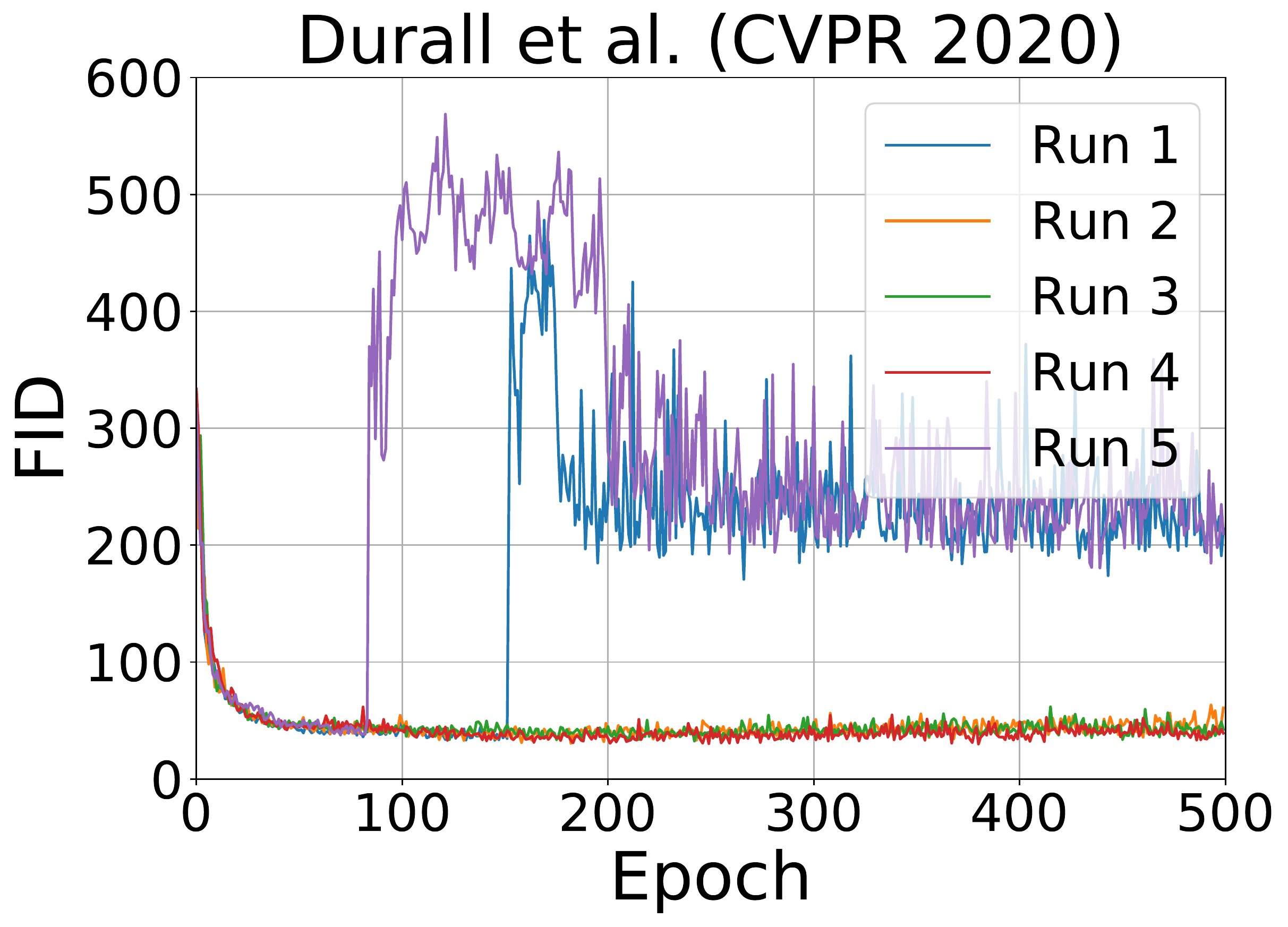}
        \caption{$2$ out of $5$ runs are unstable using the implementation provided by Durall et al.}
        \label{fig:y equals x}
    \end{subfigure}
    \hfill
    \begin{subfigure}[b]{\linewidth}
        \centering
        \includegraphics[width=\linewidth]{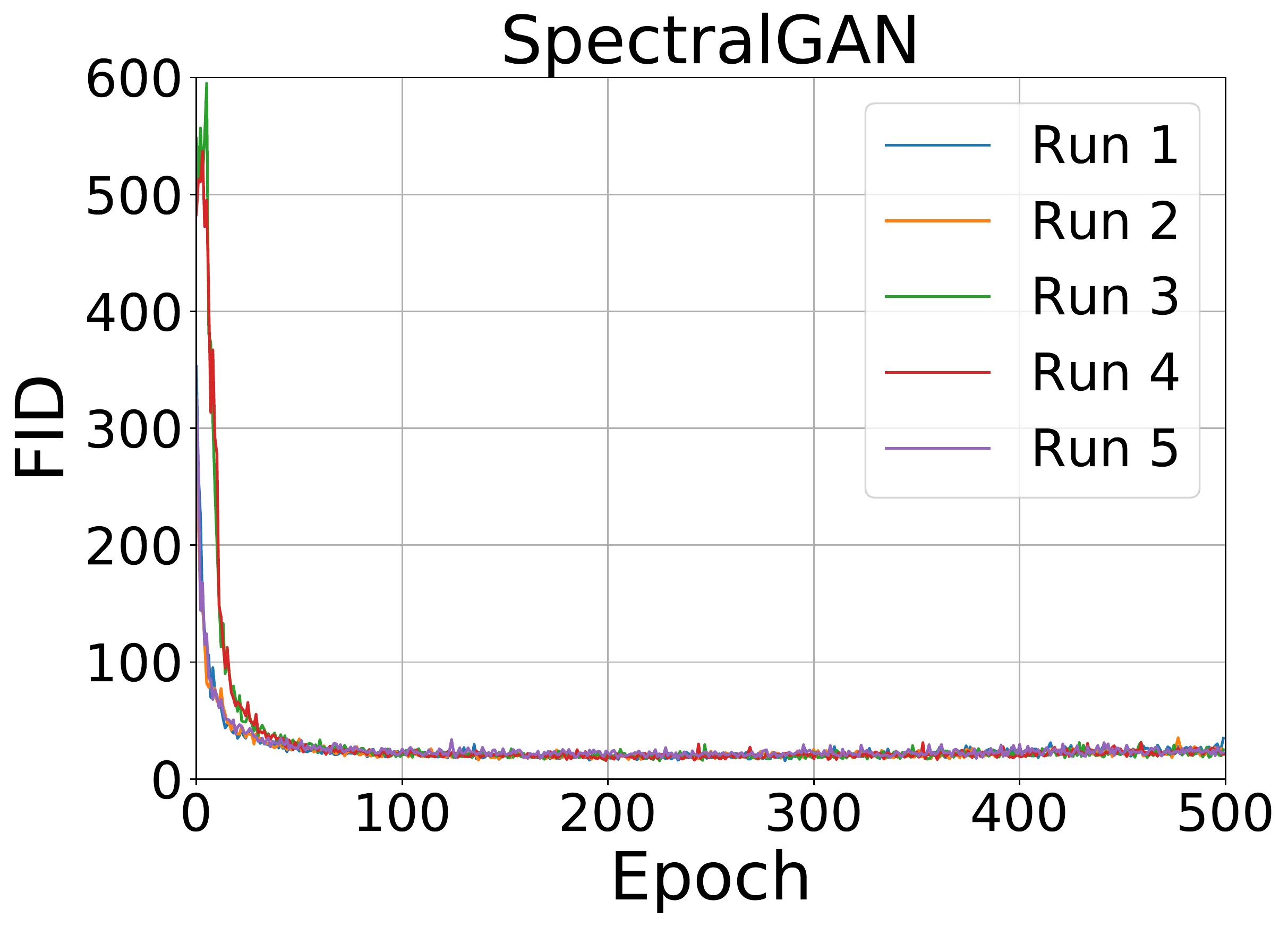}
        \caption{All of our $5$ runs are stable.}
        \label{fig:three sin x}
    \end{subfigure}
    \caption{Comparison of the training stability of Durall et al. and our proposed method. The training was run for $500$ epochs on $64^2$ resolutions using the DCGAN loss.}
    \label{fig:three graphs}
\end{figure}

Next, we compare the training stability of the proposed method with the one from \cite{margret2020upconvolution}. We show in Figure \ref{fig:y equals x} the FID over 500 training epochs for five runs of the model from \cite{margret2020upconvolution}. Two out of five of these runs collapsed before reaching epoch 200. Below in Figure \ref{fig:three sin x}, we show the same experiment for the proposed model, where not only the training runs stably but also the resulting FID is lower.

\section{Additional Evaluation of Generated Power Spectra}
\begin{figure}
\scriptsize
\begin{tabular}{@{\hspace{0cm}}c@{\hspace{0cm}}c@{}c@{}c@{}c@{}c@{}c@{}c@{}}
\imagetop{\includegraphics[width=0.33\linewidth]{img/img_2/Spectrum_64_DCGAN.pdf}}&
\imagetop{\includegraphics[width=0.33\linewidth]{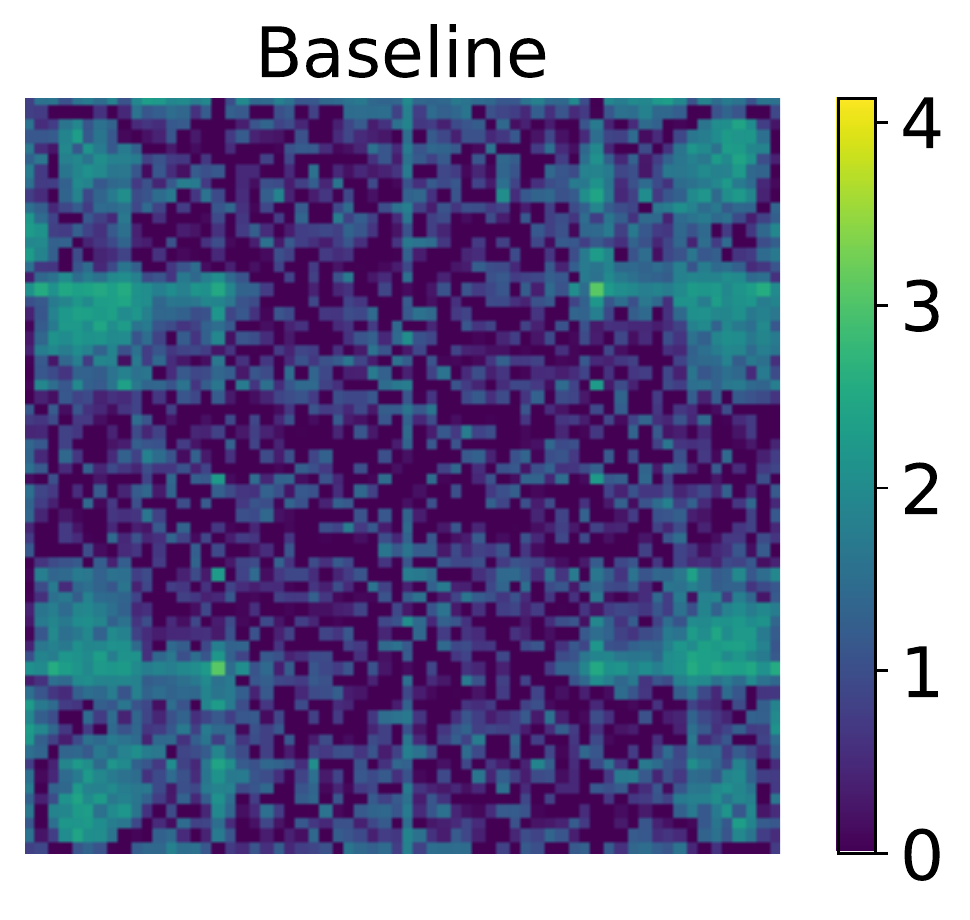}}&
\imagetop{\includegraphics[width=0.33\linewidth]{img/img_2/Spectrum_64_DCGAN_Spectral.pdf}}\\
(a) DCGAN $64\times64$& (b) Durall et al., CVPR'20& (c) DCGAN Spectral
\end{tabular}
\caption{Average magnitude differences of the 2D FFT between real and generated images. We compare (a) DCGAN without additional loss or regularization, (b) DCGAN with the regularization proposed in \cite{margret2020upconvolution} and (c) the proposed model with spectral discriminator.}
\label{fig:spec_baseline}
\end{figure}

\begin{figure}[!ht]
\scriptsize
\begin{tabular}{@{\hspace{0cm}}c@{\hspace{0cm}}c@{\hspace{0cm}}}
        \includegraphics[width=0.49\linewidth]{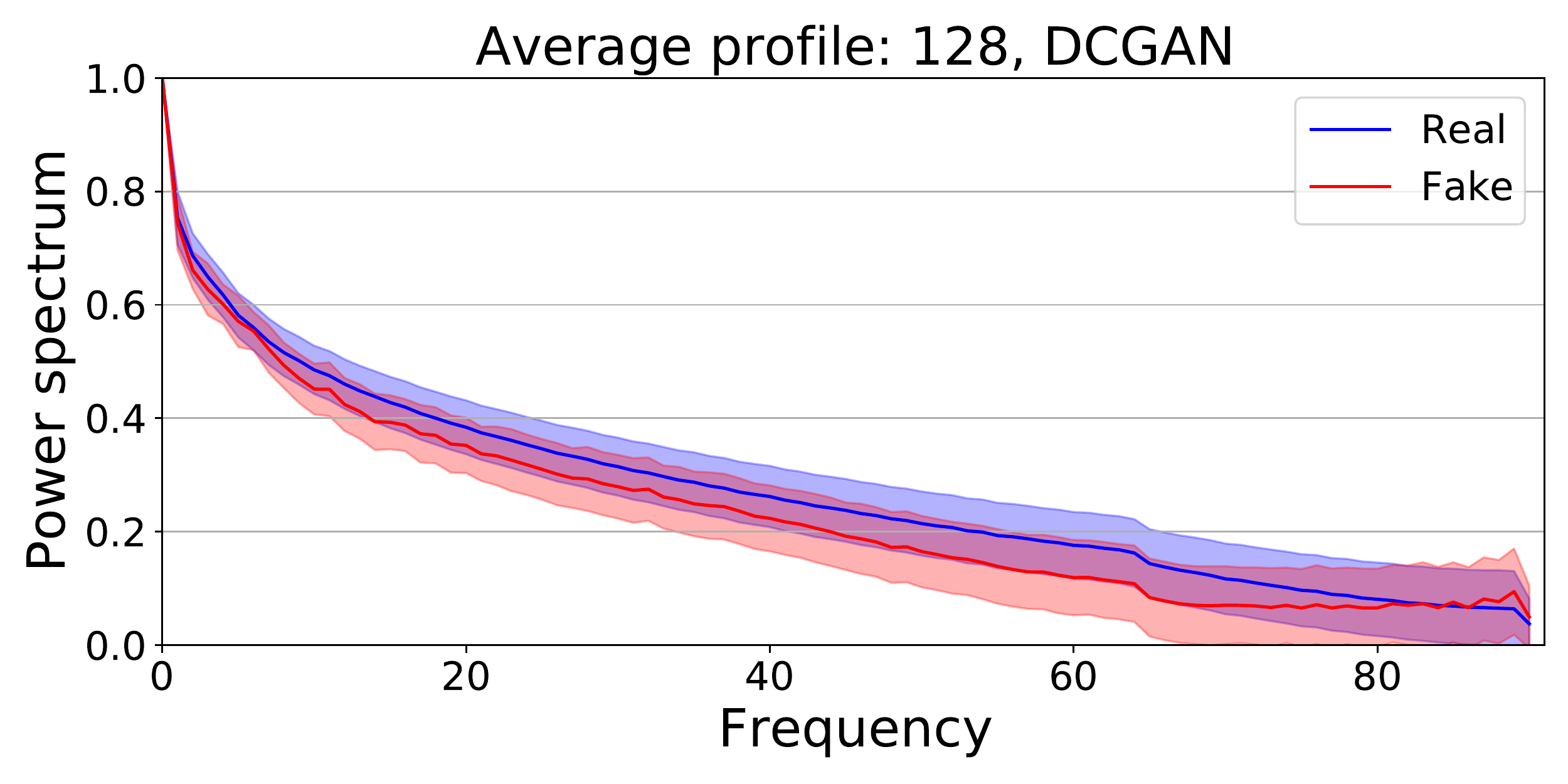}&
        \includegraphics[width=0.49\linewidth]{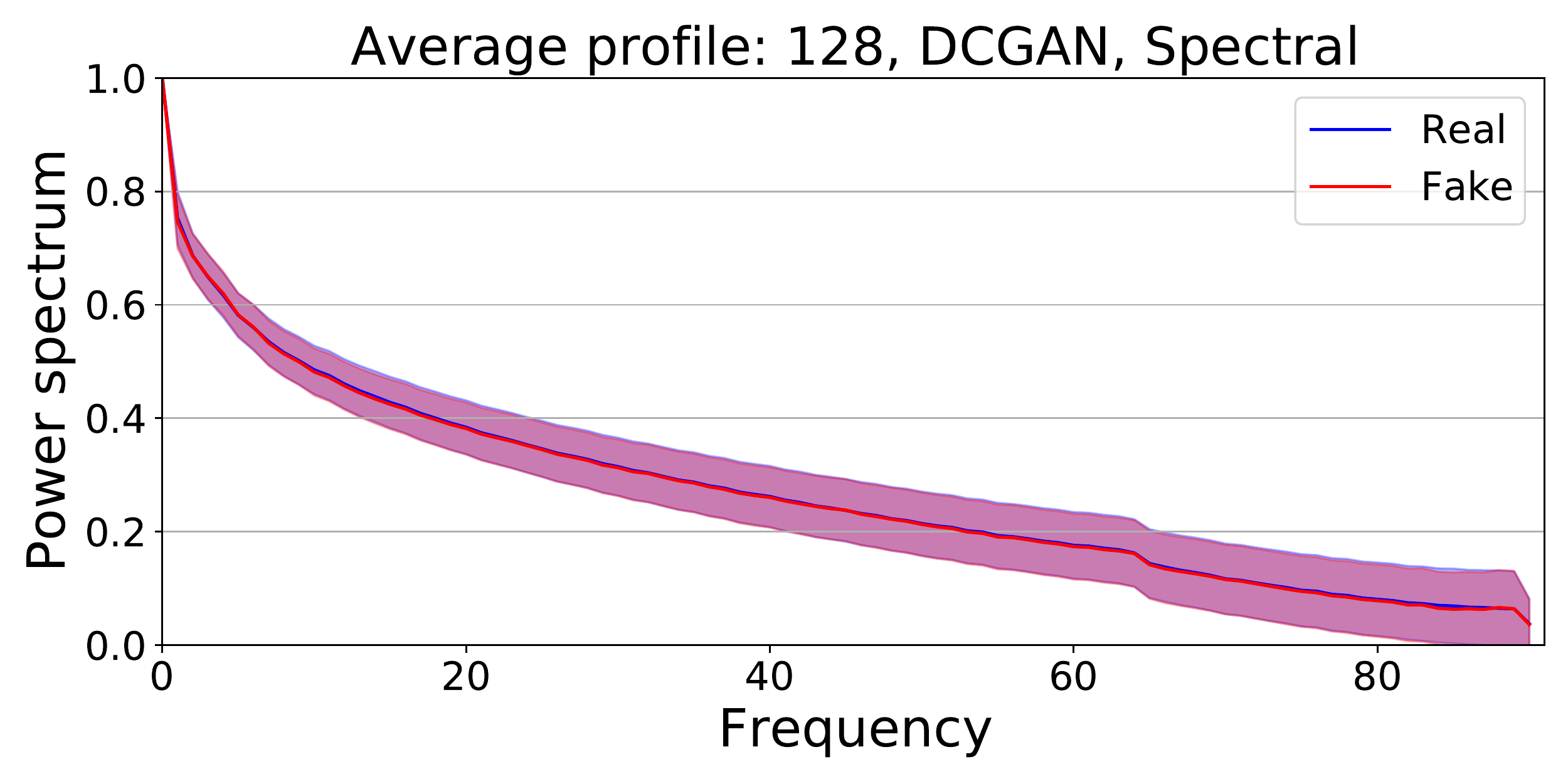}\\
        $128\times 128 $ DCGAN& DCGAN Spectral\\
        \includegraphics[width=0.49\linewidth]{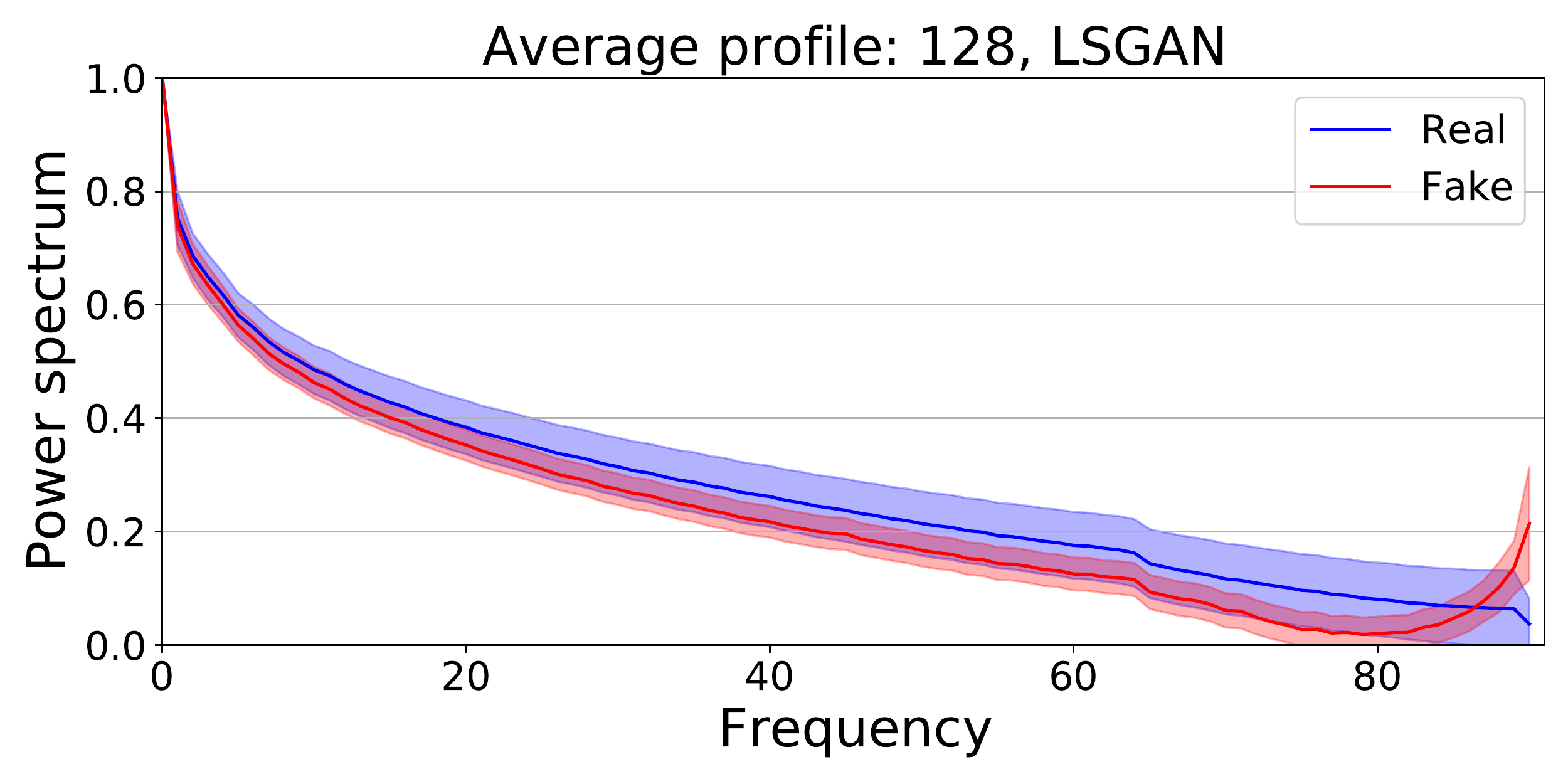}&
        \includegraphics[width=0.49\linewidth]{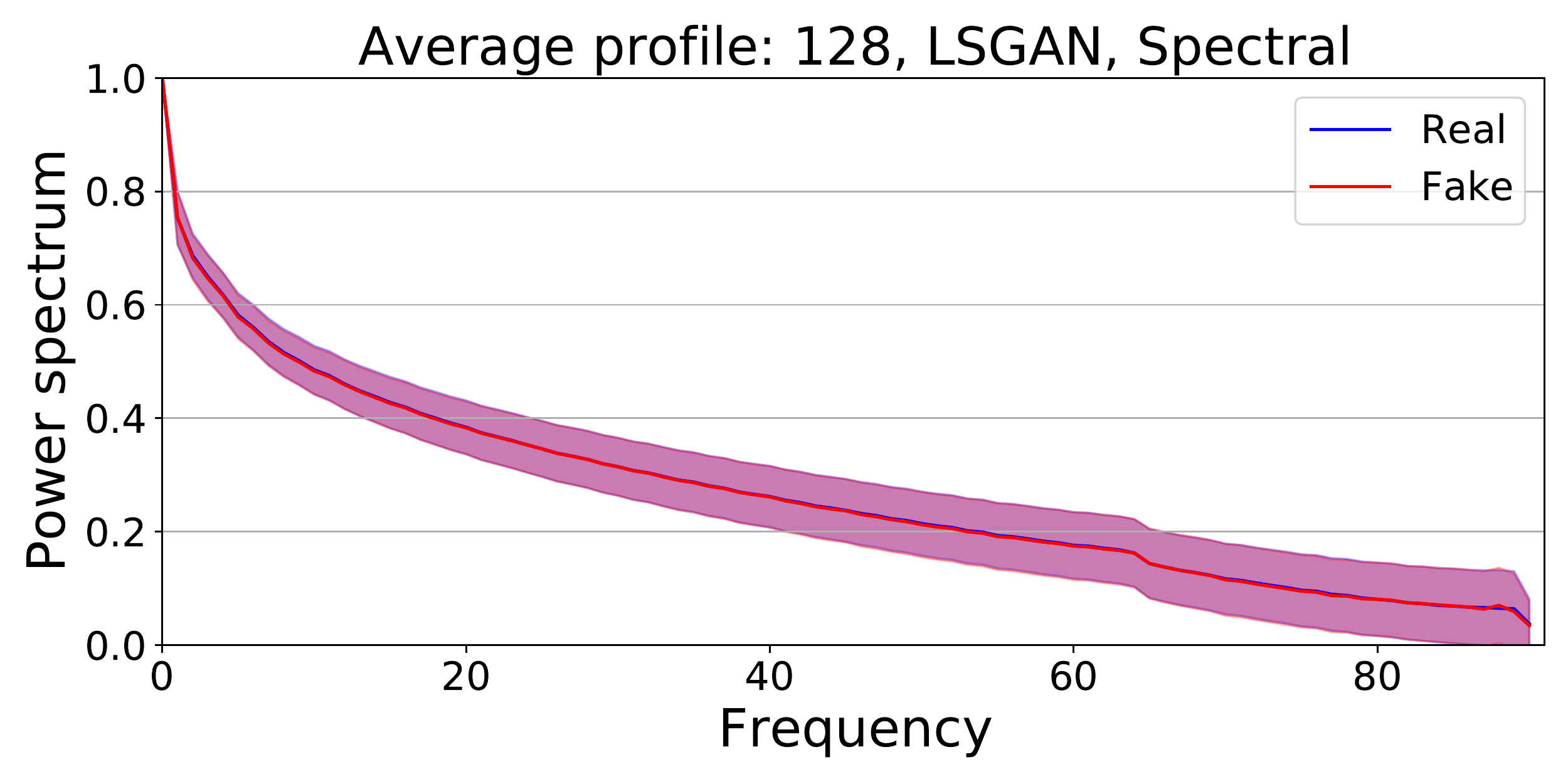}\\
        $128\times 128 $ LSGAN& LSGAN Spectral\\
                \includegraphics[width=0.49\linewidth]{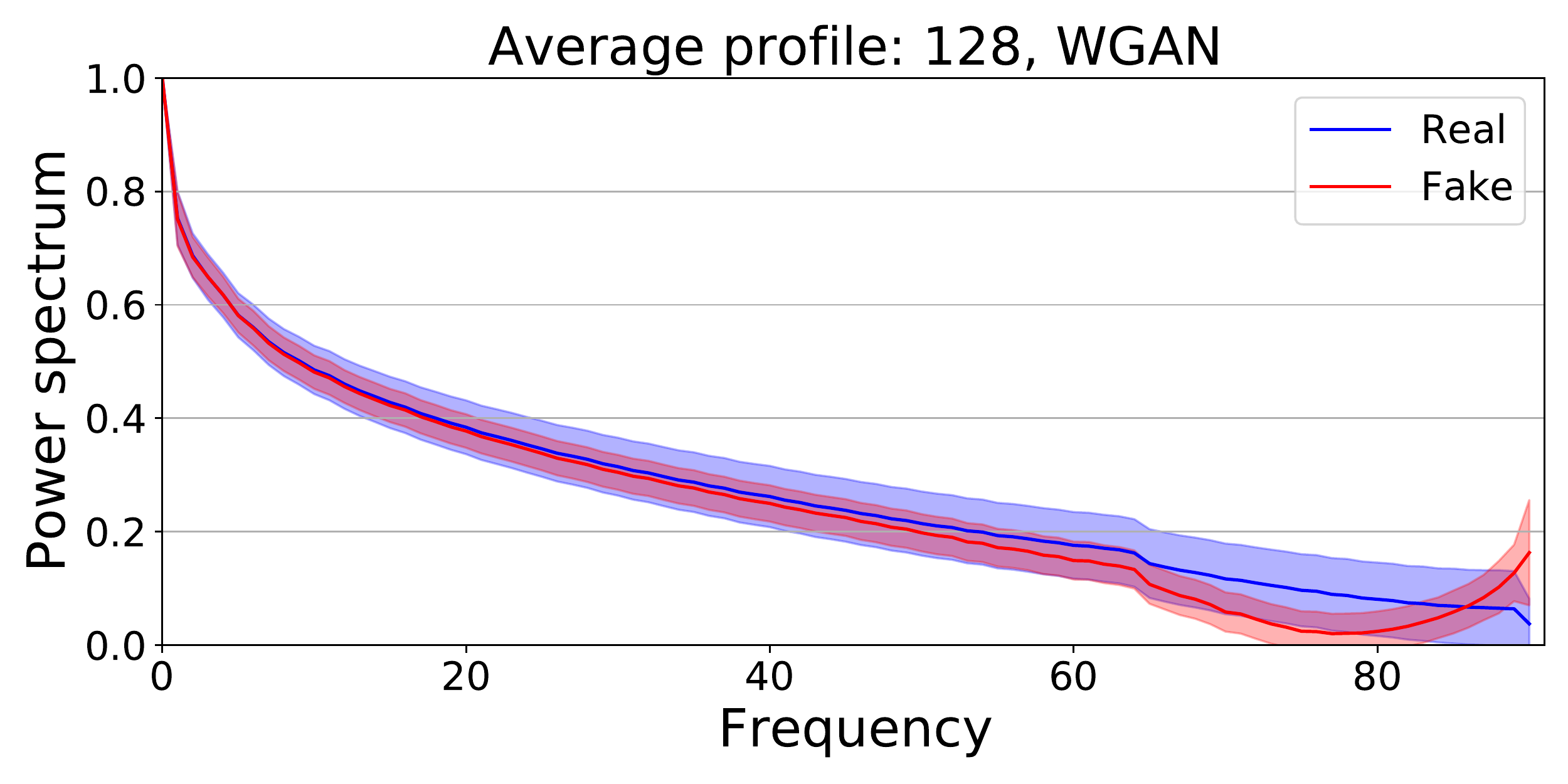}&
        \includegraphics[width=0.49\linewidth]{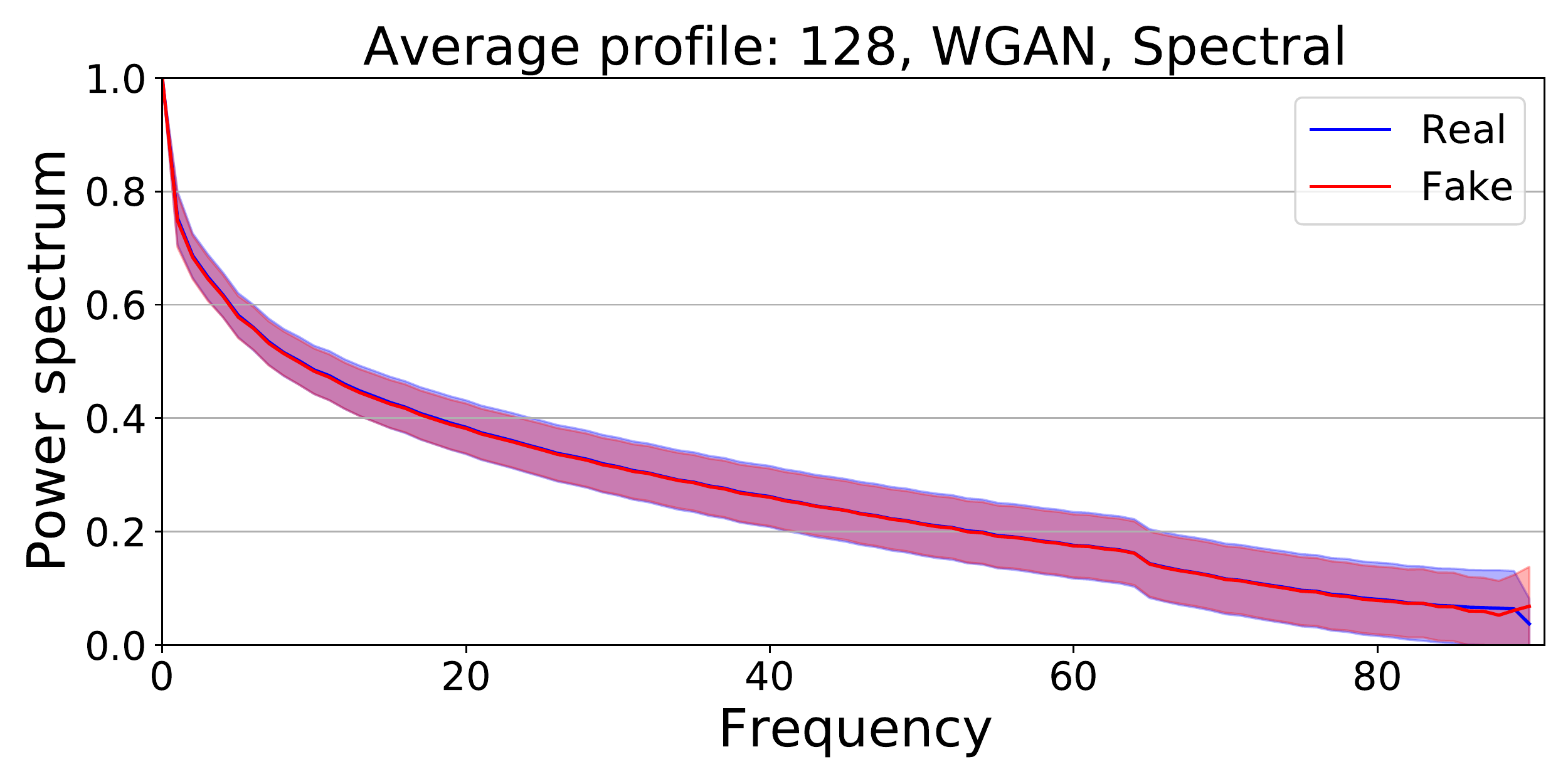}\\
        $128\times 128 $ WGAN& WGAN Spectral\\
        \includegraphics[width=0.49\linewidth]{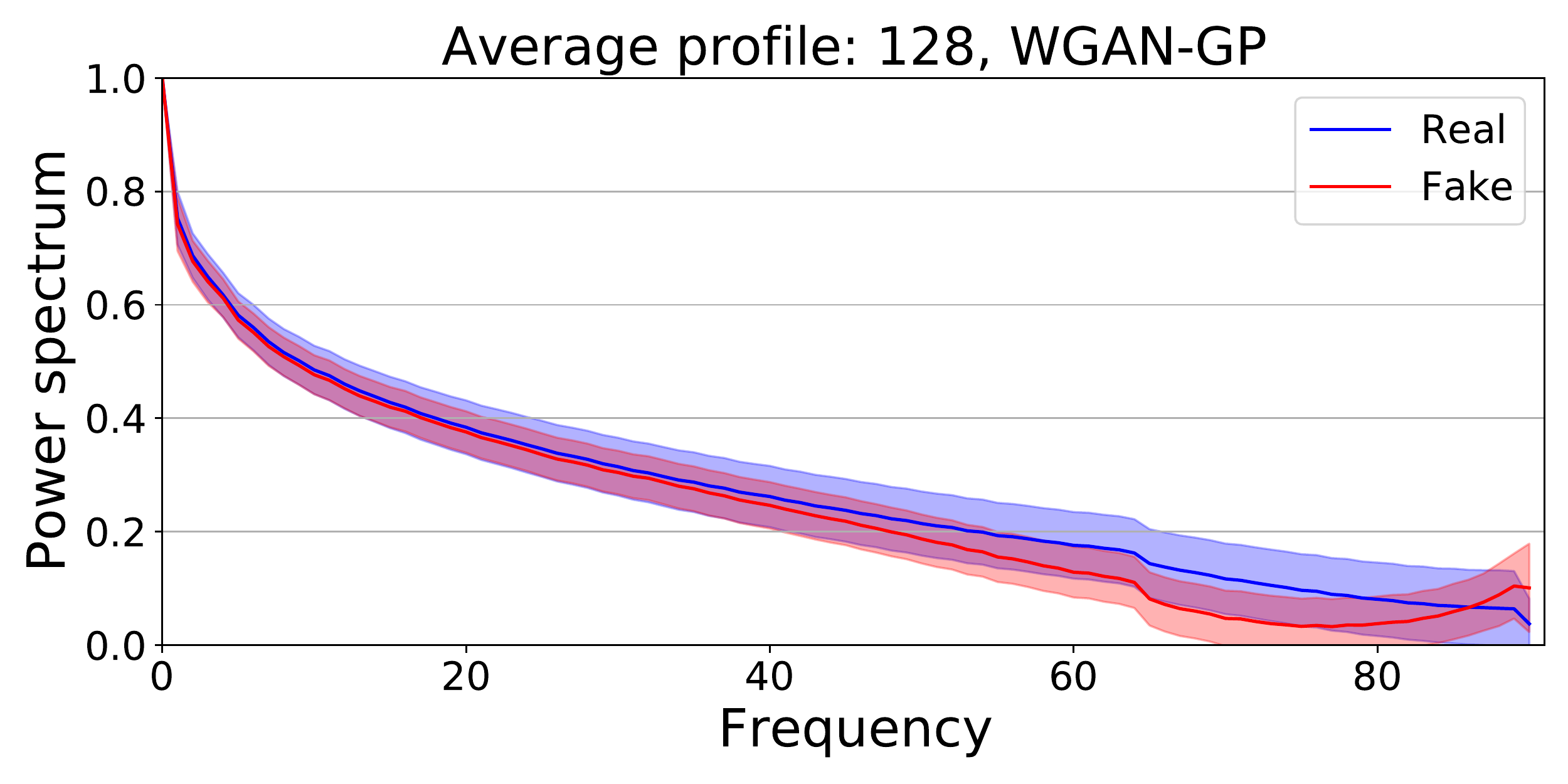}&
        \includegraphics[width=0.49\linewidth]{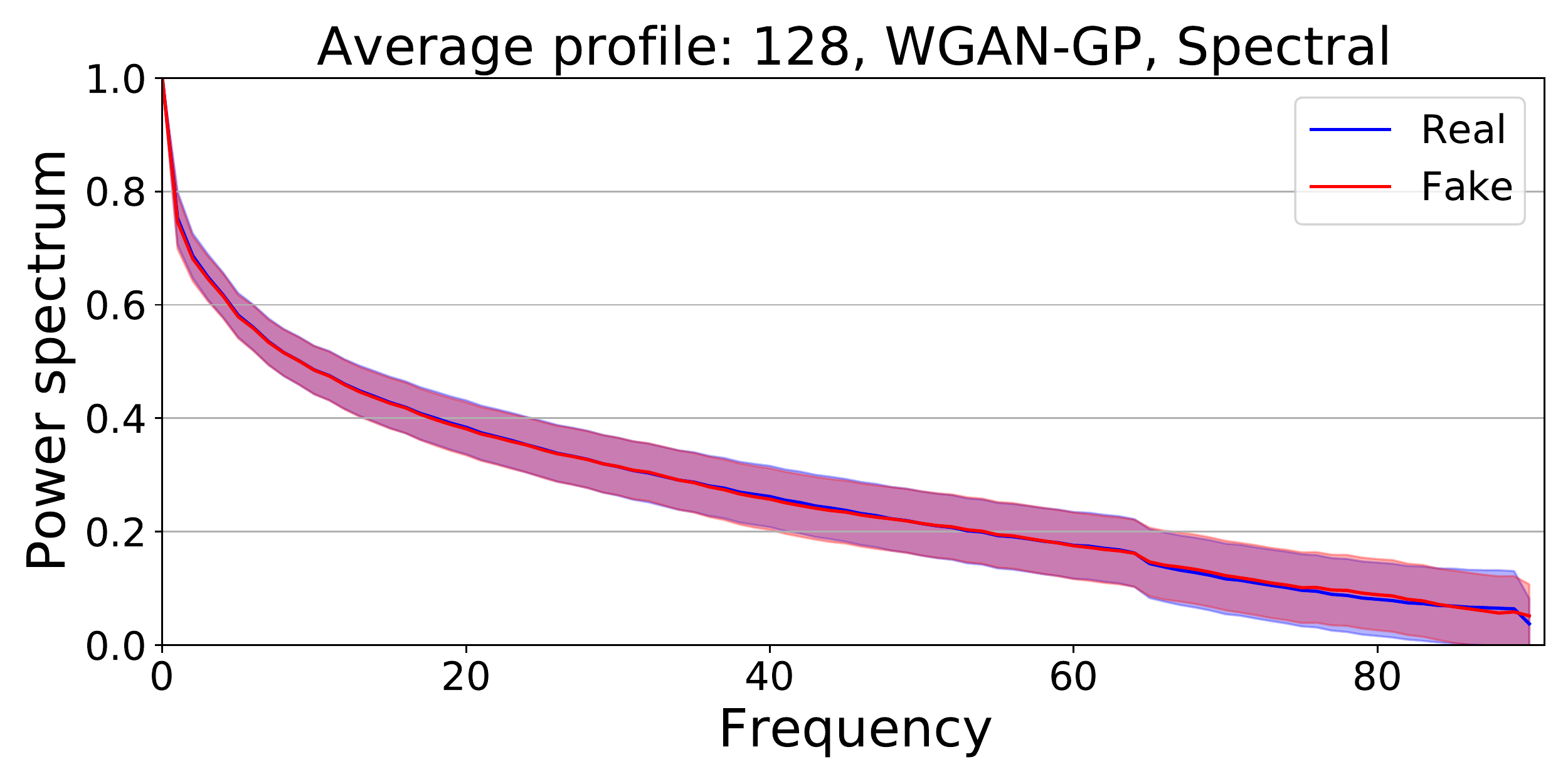}\\
        $128\times 128 $ WGAN-GP& WGAN-GP Spectral\\
        \includegraphics[width=0.49\linewidth]{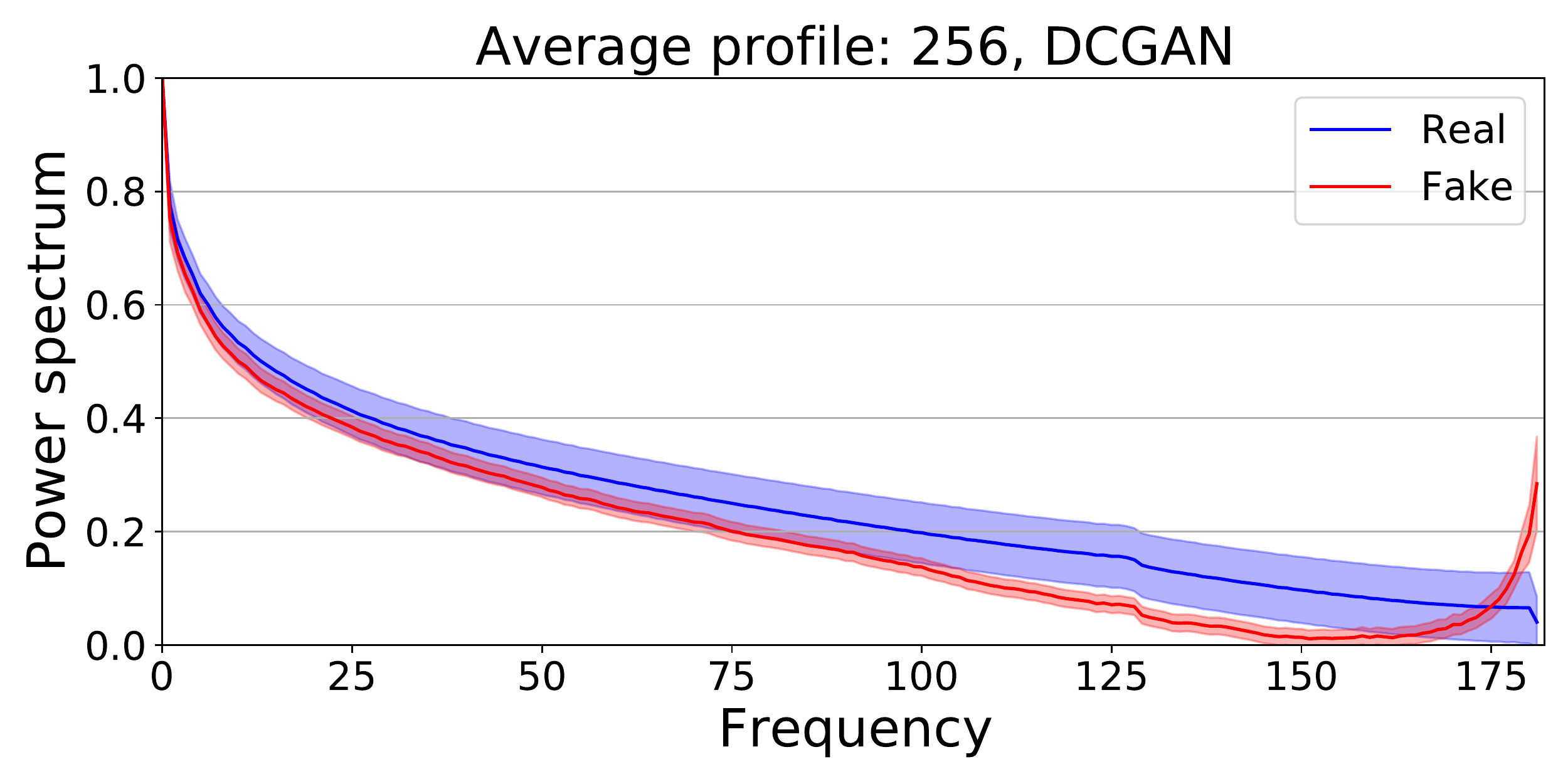}&
        \includegraphics[width=0.49\linewidth]{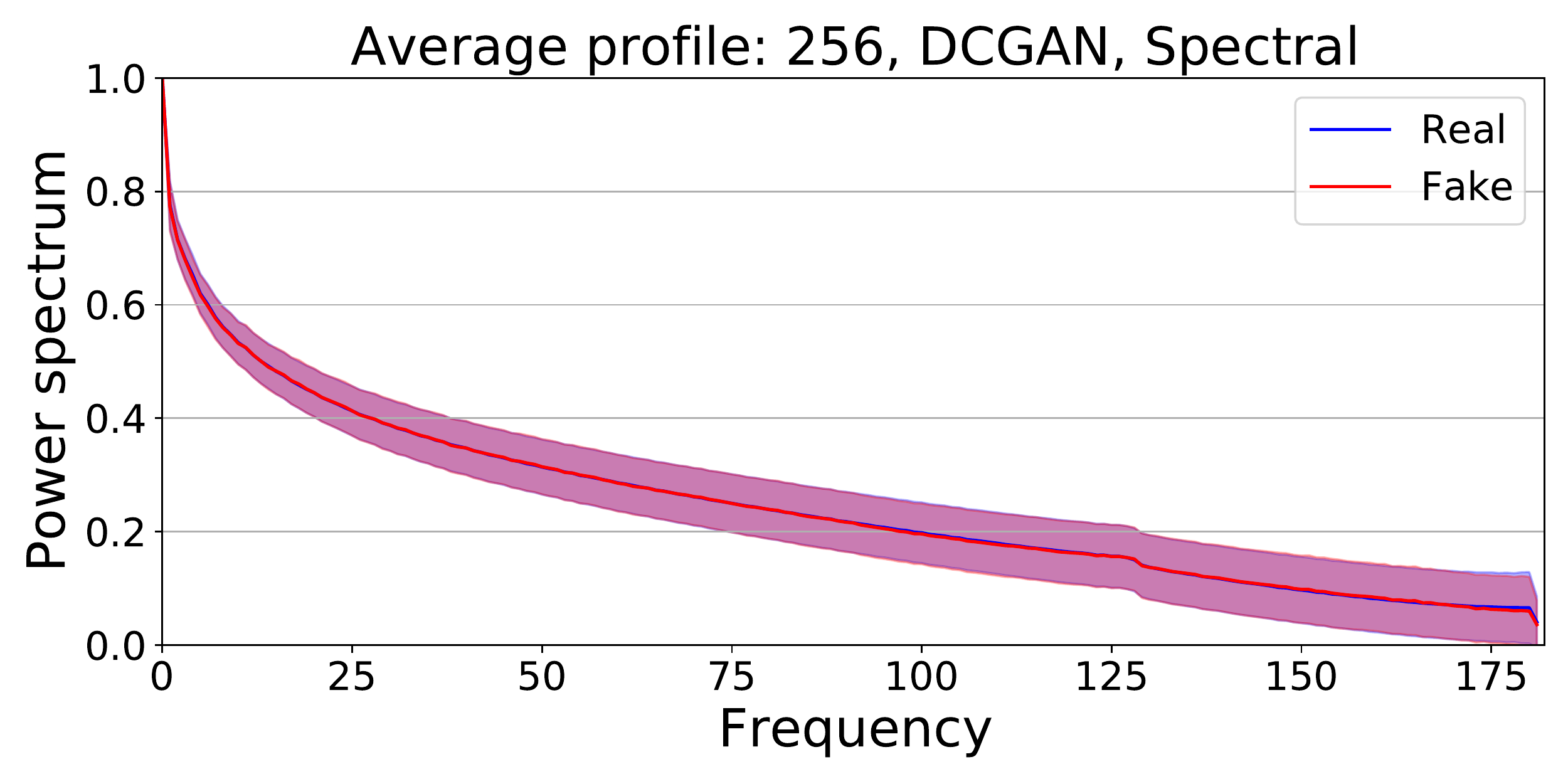}\\
        $256\times 256 $ DCGAN& DCGAN Spectral\\
        \includegraphics[width=0.49\linewidth]{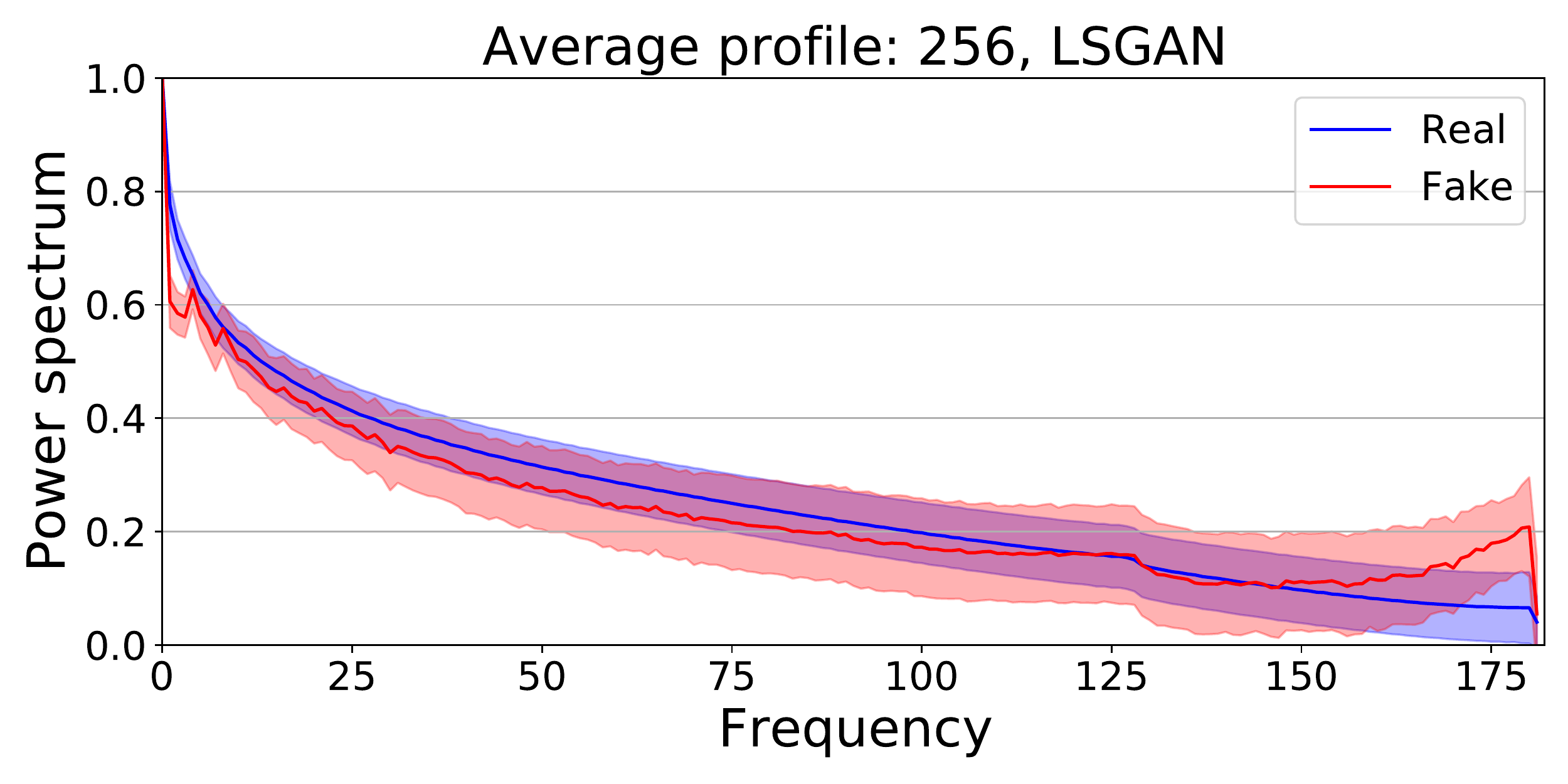}&
        \includegraphics[width=0.49\linewidth]{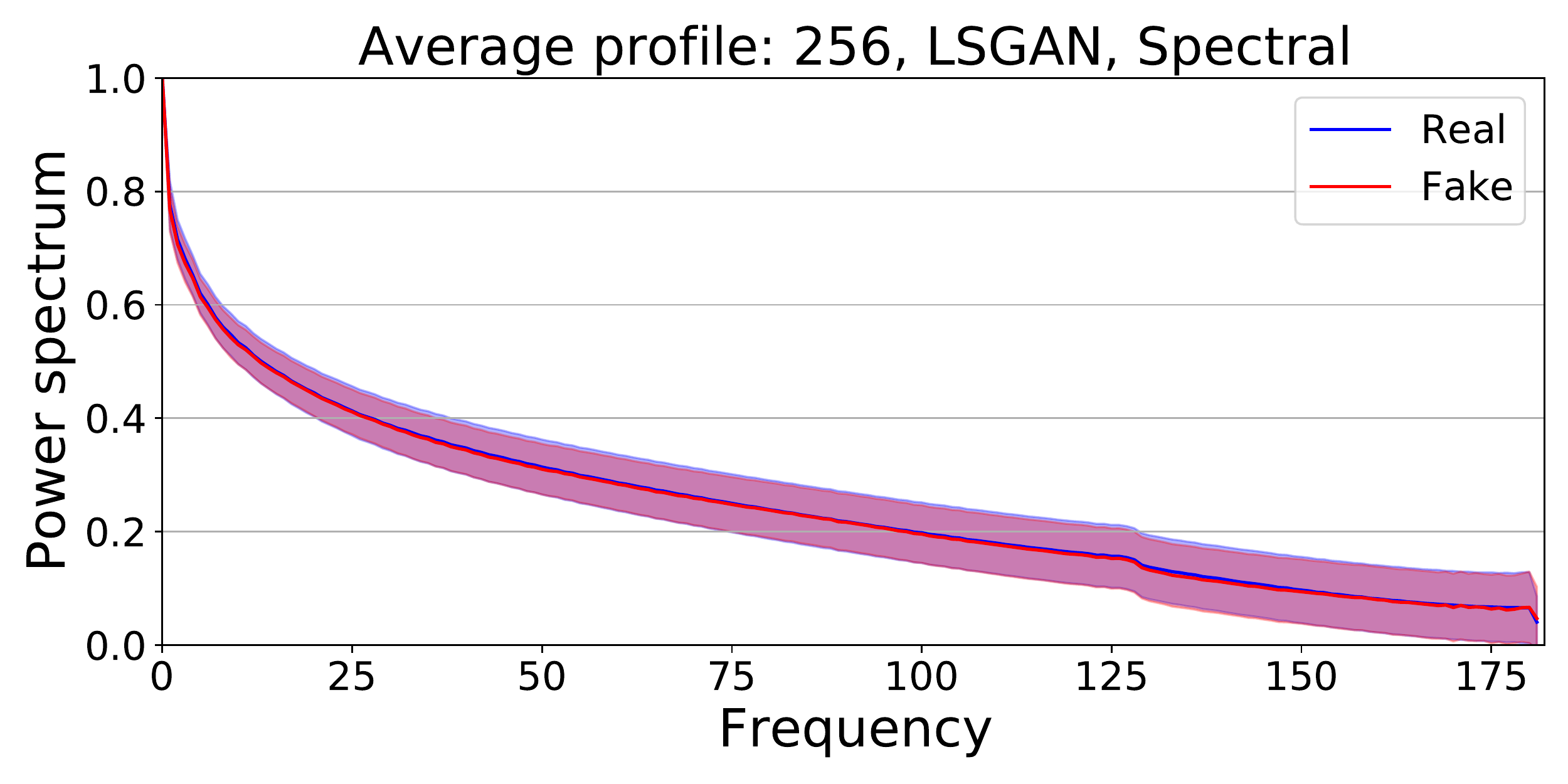}\\
        $256\times 256 $ LSGAN& LSGAN Spectral\\

       
     %

    \end{tabular}
    \caption{
        Experiments with our models trained on FFHQ128 and FFHQ256.
        \emph{Spectral} indicates that $D_F$ was applied. 
         Without the spectral discriminator, the spectral profiles of real and generated images are significantly different in their distribution.
        With the spectral discriminator, the mean and standard deviation of the spectral profiles fit almost perfectly. 
    }
    \label{fig:cloaking2}
    
\end{figure}

Figure \ref{fig:spec_baseline} shows the mean absolute differences of the 2D power spectra of real and generated images as in Fig. 5 of the main paper. Here, we compare the spectra resulting from DCGAN (Figure \ref{fig:spec_baseline} (a)), from DCGAn with the spectral regularization from \citet{margret2020upconvolution} (Figure \ref{fig:spec_baseline} (b)) and our proposed model (Figure \ref{fig:spec_baseline} (c)). With the proposed model, the mean differences are significantly smaller even in the 2D power spectral, although the discriminator only operates on 1D projections of the spectra. The 2D spectra of the reference method \cite{margret2020upconvolution} still exhibit strong deviations.  
In Figure \ref{fig:cloaking2} we report the mean and standard deviation of the spectral profiles (azimuthal integrals) of the proposed model for higher resolution images. As shown in the maim paper for images of resolution $64\times 64$, the distributions fit almost perfectly when our proposed discriminator is use. Figure \ref{fig:cloaking2} shows the corresponding plots for resolutions $128\times 128$ and $256\times256$. In both cases, the 1D projections of the power spectra are very similarly distributed when our model is used while they show obvious differences otherwise.

\section{Training of the Logistic Regression for the Cloaking Score}
\begin{figure}[t]
        \centering
        \includegraphics[width=0.6\linewidth]{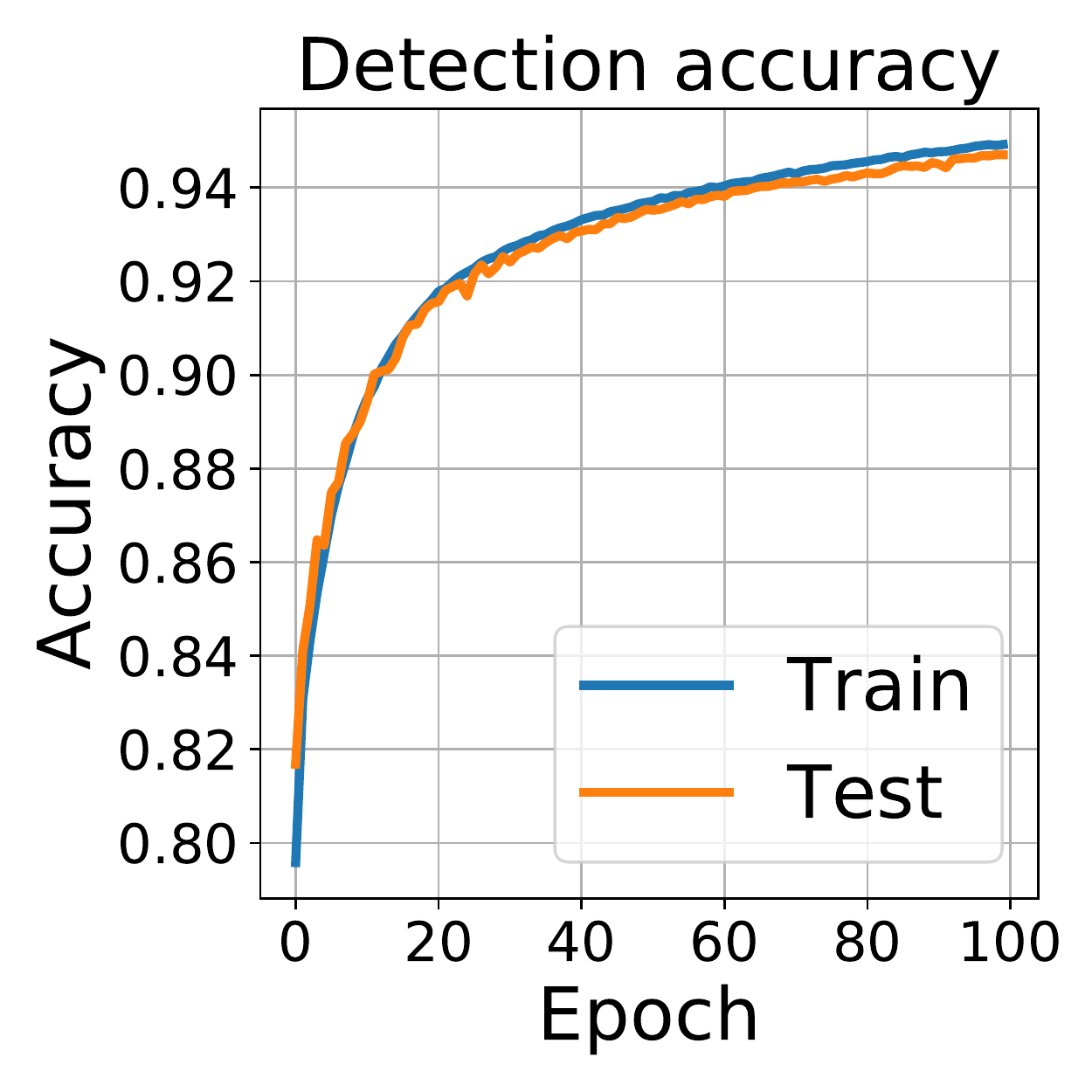}
        \caption{
            Training and test accuracy of a Logistic Regression (LR) trained on $120k$ training images, $60k$ taken from FFHQ and $60k$ generated by StyleGAN2.
            $20k$ images are used for testing. The real and generated images can to a large degree be distinguished using LR.
        }
        \label{fig:cdetection}
    \end{figure}

Figure \ref{fig:cdetection} shows the training progress of our regression model used in the cloaking score computation in the section \emph{GAN Evaluation in the Frequency Domain} of the main paper.
Training and testing accuracy are almost identical.
Training accuracy increases progressively to $0.949$ after $100$ epochs, $0.979$ after $1\,000$ epochs, $0.989$ after $10\,000$ epochs, and $0.991$ after $40\,000$ epochs.
We stopped training at this point.
    \pagebreak
    \pagebreak
\section{Sample images generated from the Proposed Model and the Baselines}
In Figures 6 to 13, we show examples of images generated with DCGAN and LSGAN without and with the proposed spectral discriminator for different resolutions. In all cases, the generated images are visually appealing and diverse.
\newpage

\begingroup
\setlength\tabcolsep{0pt} 
\renewcommand{\arraystretch}{0}
\centering
\begin{figure}[t]
\begin{tabular}{ccccc}
     \includegraphics[width=0.18\linewidth]{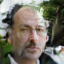} &
     \includegraphics[width=0.18\linewidth]{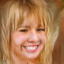} &
     \includegraphics[width=0.18\linewidth]{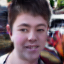} &
     \includegraphics[width=0.18\linewidth]{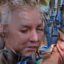} &
     \includegraphics[width=0.18\linewidth]{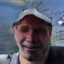} \\
     \includegraphics[width=0.18\linewidth]{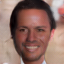} &
     \includegraphics[width=0.18\linewidth]{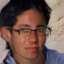} &
     \includegraphics[width=0.18\linewidth]{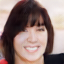} &
     \includegraphics[width=0.18\linewidth]{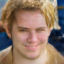} &
     \includegraphics[width=0.18\linewidth]{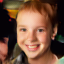} \\
     \includegraphics[width=0.18\linewidth]{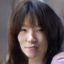} &
     \includegraphics[width=0.18\linewidth]{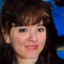} &
     \includegraphics[width=0.18\linewidth]{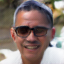} &
     \includegraphics[width=0.18\linewidth]{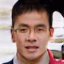} &
     \includegraphics[width=0.18\linewidth]{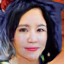} \\
     \includegraphics[width=0.18\linewidth]{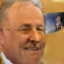} &
     \includegraphics[width=0.18\linewidth]{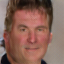} &
     \includegraphics[width=0.18\linewidth]{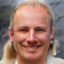} &
     \includegraphics[width=0.18\linewidth]{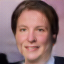} &
     \includegraphics[width=0.18\linewidth]{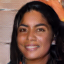}
\end{tabular}
\caption{Generated images, DCGAN, $64^2$.}
\end{figure}
\endgroup

\begingroup
\setlength\tabcolsep{0pt} 
\renewcommand{\arraystretch}{0}
\begin{figure}[t]
\begin{tabular}{ccccc}
     \includegraphics[width=0.18\linewidth]{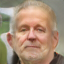} &
     \includegraphics[width=0.18\linewidth]{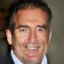} &
     \includegraphics[width=0.18\linewidth]{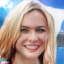} &
     \includegraphics[width=0.18\linewidth]{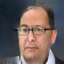} &
     \includegraphics[width=0.18\linewidth]{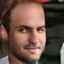} \\
     \includegraphics[width=0.18\linewidth]{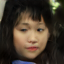} &
     \includegraphics[width=0.18\linewidth]{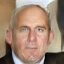} &
     \includegraphics[width=0.18\linewidth]{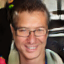} &
     \includegraphics[width=0.18\linewidth]{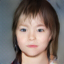} &
     \includegraphics[width=0.18\linewidth]{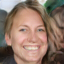} \\
     \includegraphics[width=0.18\linewidth]{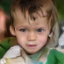} &
     \includegraphics[width=0.18\linewidth]{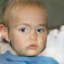} &
     \includegraphics[width=0.18\linewidth]{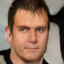} &
     \includegraphics[width=0.18\linewidth]{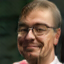} &
     \includegraphics[width=0.18\linewidth]{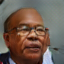} \\
     \includegraphics[width=0.18\linewidth]{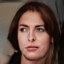} &
     \includegraphics[width=0.18\linewidth]{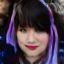} &
     \includegraphics[width=0.18\linewidth]{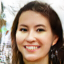} &
     \includegraphics[width=0.18\linewidth]{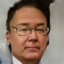} &
     \includegraphics[width=0.18\linewidth]{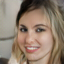}
\end{tabular}
\caption{Generated images, DCGAN, spectral, $64^2$.}
\end{figure}
\endgroup

\begingroup
\setlength\tabcolsep{0pt} 
\renewcommand{\arraystretch}{0}
\centering
\begin{figure}[t]
\begin{tabular}{ccccc}
     \includegraphics[width=0.18\linewidth]{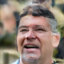} &
     \includegraphics[width=0.18\linewidth]{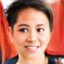} &
     \includegraphics[width=0.18\linewidth]{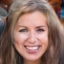} &
     \includegraphics[width=0.18\linewidth]{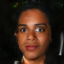} &
     \includegraphics[width=0.18\linewidth]{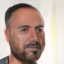} \\
     \includegraphics[width=0.18\linewidth]{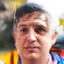} &
     \includegraphics[width=0.18\linewidth]{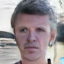} &
     \includegraphics[width=0.18\linewidth]{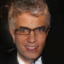} &
     \includegraphics[width=0.18\linewidth]{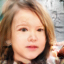} &
     \includegraphics[width=0.18\linewidth]{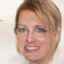} \\
     \includegraphics[width=0.18\linewidth]{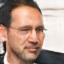} &
     \includegraphics[width=0.18\linewidth]{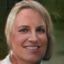} &
     \includegraphics[width=0.18\linewidth]{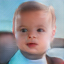} &
     \includegraphics[width=0.18\linewidth]{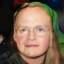} &
     \includegraphics[width=0.18\linewidth]{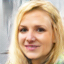} \\
     \includegraphics[width=0.18\linewidth]{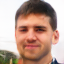} &
     \includegraphics[width=0.18\linewidth]{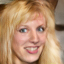} &
     \includegraphics[width=0.18\linewidth]{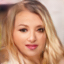} &
     \includegraphics[width=0.18\linewidth]{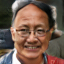} &
     \includegraphics[width=0.18\linewidth]{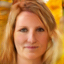}
\end{tabular}
\caption{Generated images, LSGAN, $64^2$.}
\end{figure}
\endgroup

\begingroup
\setlength\tabcolsep{0pt} 
\renewcommand{\arraystretch}{0}
\begin{figure}[t]
\begin{tabular}{ccccc}
     \includegraphics[width=0.18\linewidth]{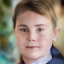} &
     \includegraphics[width=0.18\linewidth]{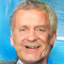} &
     \includegraphics[width=0.18\linewidth]{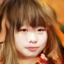} &
     \includegraphics[width=0.18\linewidth]{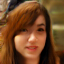} &
     \includegraphics[width=0.18\linewidth]{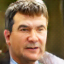} \\
     \includegraphics[width=0.18\linewidth]{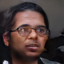} &
     \includegraphics[width=0.18\linewidth]{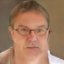} &
     \includegraphics[width=0.18\linewidth]{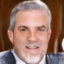} &
     \includegraphics[width=0.18\linewidth]{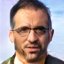} &
     \includegraphics[width=0.18\linewidth]{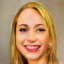} \\
     \includegraphics[width=0.18\linewidth]{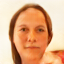} &
     \includegraphics[width=0.18\linewidth]{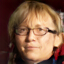} &
     \includegraphics[width=0.18\linewidth]{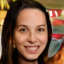} &
     \includegraphics[width=0.18\linewidth]{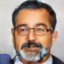} &
     \includegraphics[width=0.18\linewidth]{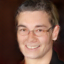} \\
     \includegraphics[width=0.18\linewidth]{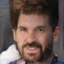} &
     \includegraphics[width=0.18\linewidth]{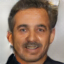} &
     \includegraphics[width=0.18\linewidth]{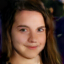} &
     \includegraphics[width=0.18\linewidth]{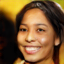} &
     \includegraphics[width=0.18\linewidth]{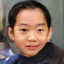}
\end{tabular}
\caption{Generated images, LSGAN, spectral, $64^2$.}
\end{figure}
\endgroup

\newpage

\begingroup
\setlength\tabcolsep{0pt} 
\renewcommand{\arraystretch}{0}
\begin{figure}[t]
\centering
\begin{tabular}{ccccc}
     \includegraphics[width=0.32\linewidth]{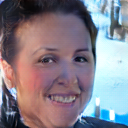} &
     \includegraphics[width=0.32\linewidth]{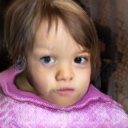} &
     \includegraphics[width=0.32\linewidth]{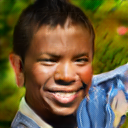} \\
     \includegraphics[width=0.32\linewidth]{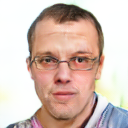} &
     \includegraphics[width=0.32\linewidth]{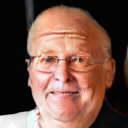} &
     \includegraphics[width=0.32\linewidth]{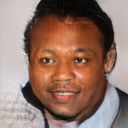} \\
     \includegraphics[width=0.32\linewidth]{img/generated_samples/DCGAN128/0294.png} &
     \includegraphics[width=0.32\linewidth]{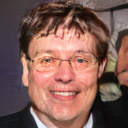} &
     \includegraphics[width=0.32\linewidth]{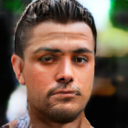}
\end{tabular}
\caption{Generated images, DCGAN, $128^2$.}
\end{figure}
\endgroup

\begingroup
\setlength\tabcolsep{0pt} 
\renewcommand{\arraystretch}{0}
\begin{figure}[t]
\centering
\begin{tabular}{ccccc}
     \includegraphics[width=0.32\linewidth]{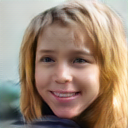} &
     \includegraphics[width=0.32\linewidth]{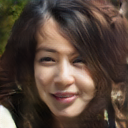} &
     \includegraphics[width=0.32\linewidth]{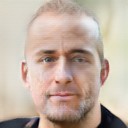} \\
     \includegraphics[width=0.32\linewidth]{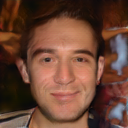} &
     \includegraphics[width=0.32\linewidth]{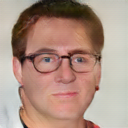} &
     \includegraphics[width=0.32\linewidth]{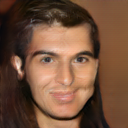} \\
     \includegraphics[width=0.32\linewidth]{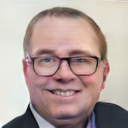} &
     \includegraphics[width=0.32\linewidth]{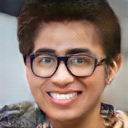} &
     \includegraphics[width=0.32\linewidth]{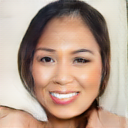}
\end{tabular}
\caption{Generated images, DCGAN, spectral, $128^2$.}
\end{figure}
\endgroup

\begingroup
\setlength\tabcolsep{0pt} 
\renewcommand{\arraystretch}{0}
\begin{figure}[t]
\centering
\begin{tabular}{ccccc}
     \includegraphics[width=0.32\linewidth]{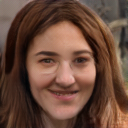} &
     \includegraphics[width=0.32\linewidth]{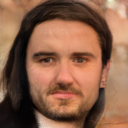} &
     \includegraphics[width=0.32\linewidth]{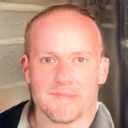} \\
     \includegraphics[width=0.32\linewidth]{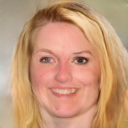} &
     \includegraphics[width=0.32\linewidth]{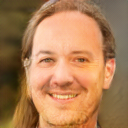} &
     \includegraphics[width=0.32\linewidth]{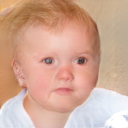} \\
     \includegraphics[width=0.32\linewidth]{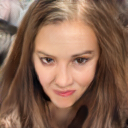} &
     \includegraphics[width=0.32\linewidth]{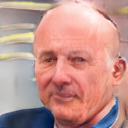} &
     \includegraphics[width=0.32\linewidth]{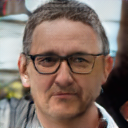}
\end{tabular}
\caption{Generated images, LSGAN, $128^2$.}
\end{figure}
\endgroup

\begingroup
\setlength\tabcolsep{0pt} 
\renewcommand{\arraystretch}{0}
\begin{figure}[t]
\centering
\begin{tabular}{ccccc}
     \includegraphics[width=0.32\linewidth]{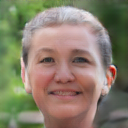} &
     \includegraphics[width=0.32\linewidth]{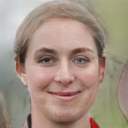} &
     \includegraphics[width=0.32\linewidth]{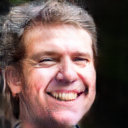} \\
     \includegraphics[width=0.32\linewidth]{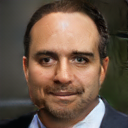} &
     \includegraphics[width=0.32\linewidth]{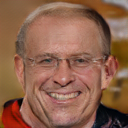} &
     \includegraphics[width=0.32\linewidth]{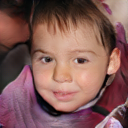} \\
     \includegraphics[width=0.32\linewidth]{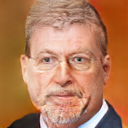} &
     \includegraphics[width=0.32\linewidth]{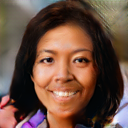} &
     \includegraphics[width=0.32\linewidth]{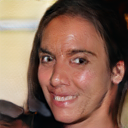}
\end{tabular}
\caption{Generated images, LSGAN, spectral, $128^2$.}
\end{figure}
\endgroup

\bibliography{manuscrip}